\documentclass{article}
\usepackage{ijcai26}

\usepackage{times}
\usepackage{soul}
\usepackage{url}
\usepackage[hidelinks]{hyperref}
\usepackage[utf8]{inputenc}
\usepackage[T1]{fontenc}
\usepackage[small]{caption}
\usepackage{graphicx}
\usepackage{amsmath}
\usepackage{amsthm}
\usepackage{booktabs}
\usepackage{algorithm}
\usepackage{algorithmic}
\usepackage[switch]{lineno}
\usepackage{amssymb}
\usepackage{bm}
\usepackage[dvipsnames]{xcolor}
\usepackage[most]{tcolorbox}
\usepackage{enumitem}
\usepackage{multirow}
\usepackage{makecell}
\usepackage{arydshln}
\usepackage{cancel}
\usepackage{wrapfig}
\usepackage{float}

\newtheorem{assumption}{Assumption}
\newtheorem{theorem}{Theorem}
\newtheorem{proposition}{Proposition}
\newtheorem{lemma}{Lemma}
\newcommand{\dd}{\mathrm{d}}
\newcommand{\eff}{\mathrm{eff}}
\DeclareMathOperator{\Tr}{Tr}
\DeclareMathOperator*{\argmin}{argmin}

\title{
Can Stationary Distributions of Scale-Invariant Neural Networks \\ Be Described by the Thermodynamics of an Ideal Gas?
}

\author{
Ildus Sadrtdinov$^1$\and
Ekaterina Lobacheva$^{2,3}$\and
Ivan Klimov$^{1}$\and
Mikhail	Burtsev$^{4}$\and\\
Mikhail I. Katsnelson$^{1,5}$\And
Dmitry Vetrov$^1$\\
\affiliations
$^1$Constructor University\\
$^2$Mila – Quebec AI Institute\\
$^3$Université de Montréal\\
$^4$London Institute for Mathematical Sciences\\
$^5$Institute for Molecules and Materials, Radboud University \\
\emails
isadrtdinov@constructor.university
}

\begin{document}

\maketitle

\begin{abstract}
    Understanding the training dynamics of deep neural networks remains a major open problem, with physics-inspired approaches offering promising insights. Building on this perspective, we develop a thermodynamic framework to describe the stationary distributions of stochastic gradient descent (SGD) with weight decay for scale-invariant neural networks, a setting that both reflects practical architectures with normalization layers and permits theoretical analysis. We establish analogies between training hyperparameters (e.g., learning rate, weight decay) and thermodynamic variables such as temperature, pressure, and volume. Starting with a simplified isotropic noise model, we uncover a close correspondence between SGD dynamics and ideal gas behavior, validated through theory and simulation. Extending to training of neural networks, we show that key predictions of the framework, including the behavior of stationary entropy, align closely with experimental observations. This framework provides a principled foundation for interpreting training dynamics and may guide future work on hyperparameter tuning and the design of learning rate schedulers.
\end{abstract}

\section{Introduction}

The optimization dynamics of deep neural networks remain poorly understood despite their empirical success.
A promising approach is to draw inspiration from physics, which studies systems with many degrees of freedom and complex interactions.
In particular, thermodynamics has motivated numerous connections to neural networks~\cite{jastrzkebski2017three,chen2024constructing,zhang2024temperature,kozyrev2025explaingrokking}.
In this work, we extend this line of research and propose a framework that relates neural network optimization to thermodynamic systems.
Specifically, we interpret stochastic gradient noise as thermal fluctuations and introduce macroscopic thermodynamic variables, like temperature, pressure, and volume, to describe stationary distributions of stochastic gradient descent (SGD).
We also connect these variables to training hyperparameters such as learning rate and weight decay.

We focus on \emph{scale-invariant} neural networks, which resemble common architectures with normalization layers (e.g., BatchNorm~\cite{ioffe2015batch}, LayerNorm~\cite{ba2016layer}), while remaining theoretically tractable.
Theories not accounting for scale invariance would fail to explain phenomena observed in practical training.
Unlike prior works that focused on energy, entropy, and temperature, we show that SGD with weight decay on scale-invariant networks naturally introduces pressure and volume, establishing a direct analogy with the ideal gas law.
Our \textbf{main contributions} are:
\begin{enumerate}
    \item We study scale-invariant neural networks under three training protocols: (1) fixed parameter norm, (2) fixed effective learning rate (ELR), and (3) fixed learning rate (LR). For each setting, we derive the corresponding stochastic differential equation (SDE) that governs the dynamics and characterize their stationary distributions.
    \item We introduce a thermodynamic framework for analyzing stationary distributions of SGD. Under a simplified \emph{isotropic noise model}, we establish the thermodynamic analogy rigorously and validate it empirically, revealing a close correspondence with the ideal gas model.
    \item We discuss how the ideal gas model generalizes to actual training of scale-invariant neural networks and show that our framework accurately predicts how the entropy of the stationary distribution varies with learning rate and weight decay.
\end{enumerate}

Overall, our framework reveals a deep link between optimization and thermodynamics, providing a physics-based foundation for interpreting neural network training. 
This insight can inform future work on training hyperparameters, including learning rate scheduling.

\section{Background}

In this section, we outline the main concepts from thermodynamics and the optimization of scale-invariant neural networks that form the basis of our analogy.
For convenience, a complete list of notations is provided in Appendix~A. 
A broader overview of related work is given in Appendix~B.

\paragraph{Thermodynamics.}
Below we summarize the key principles of thermodynamics relevant to our discussion.
A more detailed introduction is given in Appendix~C.
\begin{enumerate}[label=\textbf{T\arabic*}]
    \item \label{T1} A \emph{thermodynamic system} is described by \emph{state variables}, including \emph{internal energy} $U$ (the total energy of all particles in the system), \emph{entropy} $S$ (a measure of disorder), \emph{temperature} $T$, \emph{pressure} $p$, and \emph{volume} $V$.
    \item \label{T2} The \emph{First Law of Thermodynamics} expresses energy conservation: $\dd U = \delta Q - p\,\dd V$, where $\delta Q$ is the infinitesimal heat supplied to the system. The \emph{Second Law of Thermodynamics} requires $\dd S \ge~\delta Q/T$, implying that systems evolve toward \emph{equilibrium}, where an appropriate \emph{thermodynamic potential} is minimized (\ref{T4}, \ref{T5}). A \emph{reversible process} satisfies $\dd S = \delta Q/T$.
    \item \label{T3} The \emph{Gibbs distribution} connects thermodynamics with statistical mechanics. At equilibrium, the probability of microstate $i$ with energy $E_i$ is $p_i \propto \exp(-E_i/T)$, with the temperature $T$ controlling the distribution's spread. Internal energy and entropy can be expressed as $U=\mathbb{E}_{p_i} [E_i]$ and $S=\mathbb{E}_{p_i} [-\log p_i]$.
    \item \label{T4} At fixed $T$ and $V$, the \emph{Helmholtz energy} $F = U - TS$ is minimized at equilibrium. The \emph{Maxwell relation} links derivatives of state variables and helps to determine changes in entropy, which is otherwise difficult to measure. In this case, it is given by $\big(\frac{\partial S}{\partial V}\big)_T = \big(\frac{\partial p}{\partial T}\big)_V$~\footnote{$\big(\frac{\partial f}{\partial x}\big)_y$ denotes the partial derivative of $f$ w.r.t. $x$ under fixed $y$.}.
    \item \label{T5} At fixed $T$ and $p$, the \emph{Gibbs energy} $G = U - TS + pV$ is minimized at equilibrium, with Maxwell relation $-\big(\frac{\partial S}{\partial p}\big)_T = \big(\frac{\partial V}{\partial T}\big)_p$.
    \item \label{T6} An \emph{ideal gas} is a simplified model neglecting intermolecular interactions. At equilibrium, its state variables satisfy the \emph{ideal gas law} $pV = RT$, where $R$ is the gas constant.
    \item \label{T7} The \emph{isochoric} and \emph{isobaric heat capacities} are $C_V = \big(\frac{\delta Q}{\dd T}\big)_V$ and $C_p = \big(\frac{\delta Q}{\dd T}\big)_p$. For an ideal gas, both are constants with $C_p - C_V = R$. In an \emph{adiabatic process}, defined by $\delta Q=0$, the ideal gas follows $pV^\gamma = \text{const}$ with $\gamma = C_p/C_V$.
\end{enumerate}

\paragraph{Stationary behavior of SGD.}
In classical optimization theory, SGD differs from full-batch gradient descent: instead of converging to a single stationary point of the loss function, SGD converges to a \emph{stationary distribution} centered around it.
We denote this distribution by $\rho_{\bm{w}}(\bm{w})$, where $\bm{w}\in \mathbb{R}^d$ are the model parameters.
This behavior is driven by the finite learning rate and stochastic gradient estimates, both of which inject noise into the dynamics, analogous to thermal fluctuations and naturally motivating a thermodynamic perspective.

\paragraph{Scale-invariant networks.}
Modern architectures often include normalization layers that make preceding parameters \emph{scale-invariant}, i.e., the output of the normalization layer is independent of the parameter norm.
We focus on fully scale-invariant networks, where the whole output of the network does not depend on the parameter norm (i.e., only on the unit direction vector of the weights).
The dynamics of the model on the unit sphere (i.e., the loss evolution) is governed by the \emph{effective learning rate} (ELR), defined as $\eta_{\eff} = \eta / \|\bm{w}\|^2$, where $\eta$ is the usual learning rate (LR) used in SGD iterates.

\section{Stationary Distributions of SDE}
\label{sec:sde}

In this section, we start from a stochastic differential equation (SDE) introduced in prior work~\cite{li2020reconciling,wang22three} that captures the training dynamics of scale-invariant neural networks.
We adapt this SDE to three training protocols of increasing complexity and derive their stationary distributions under an \emph{isotropic noise model}, introduced below.
Although these stationary distributions have been studied previously, we present them in a unified notation and systematize the results across the three protocols.

We begin by introducing several assumptions that enable a theoretical analysis of SGD.
First, we approximate the stochastic gradient noise by a Gaussian random vector, a standard assumption in prior work~\cite{jastrzkebski2017three,Mandt2017} naturally leading to an SDE formulation.
\begin{assumption}[Gaussian noise]
    \label{asm:1}
    Let $L(\bm{w})$ and $L_{\mathcal{B}}(\bm{w})$ denote the (full-batch) training loss and the loss on a mini-batch $\mathcal{B}$, respectively. Then,
    \begin{equation}
        \label{eqt:1}
        \nabla L_{\mathcal{B}} (\bm{w}) \approx \nabla L (\bm{w}) + \big(\bm{\Sigma} _{\bm{w}}\big)^{1/2} \bm{\varepsilon},
    \end{equation}
    where $\bm{\varepsilon} \sim \mathcal{N}(0, \bm{I}_d)$ and $\bm{\Sigma}_{\bm{w}}$ is a spatially dependent covariance matrix of stochastic gradients\footnote{We treat $\bm{\Sigma}_{\bm{w}}$ as implicitly dependent on the batch size and therefore do not model the batch size explicitly.}.
\end{assumption}
\noindent
Second, we focus on scale-invariant networks, which induce scale invariance of the loss function.
\begin{assumption}[Scale invariance]
    \label{asm:2}
    For any mini-batch $\mathcal{B}$ (including the full dataset), the loss satisfies $L_{\mathcal{B}}(\alpha \bm{w}) = L_{\mathcal{B}}(\bm{w})$ for all $\alpha > 0$.
\end{assumption}
Let $r=\|\bm{w}\|_2$ and $\overline{\bm{w}} = \bm{w} / r$. Then $L(\bm{w}) = L(\overline{\bm{w}})$.
As shown in~\cite{li2020reconciling} (Lemmas 3.1 and B.1), the full-batch gradient and the covariance matrix of a scale-invariant loss satisfy
\begin{gather}
    \label{eqt:2}
    \nabla L(\bm{w}) = \nabla L(\overline{\bm{w}})/r\qquad \bm{w}^T \nabla L(\bm{w}) = 0 \\
    \label{eqt:3} \bm{\Sigma}_{\bm{w}}=\bm{\Sigma}_{\overline{\bm{w}}}/r^2 \qquad \big(\bm{\Sigma}_{\bm{w}}\big)^{1/2}\bm{w} = 0
\end{gather}
\noindent
Next, we assume that the discrete-time dynamics of SGD can be approximated by a continuous-time SDE.
\begin{assumption}[SDE approximation]
    \label{asm:3}
    For sufficiently small (effective) learning rate and weight decay coefficient, the SGD dynamics in the weight space are well approximated by the corresponding continuous SDE.
\end{assumption}
\noindent
Finally, we introduce a simplified \emph{isotropic noise model}, which yields explicit stationary distributions of the resulting SDE.
A related assumption has been considered in~\cite{jastrzkebski2017three,wang22three}.
\begin{assumption}[Isotropic noise model]
    \label{asm:4}
    The covariance matrix at a unit vector $\overline{\bm{w}}$ takes a form $\bm{\Sigma}_{\overline{\bm{w}}} = \bm{P}_{\overline{\bm{w}}} \bm{\Sigma} \bm{P}_{\overline{\bm{w}}}$, where $\bm{P}_{\overline{\bm{w}}} = \bm{I}_d - \overline{\bm{w}}\,\overline{\bm{w}}^T$ projects onto the tangent subspace orthogonal to $\overline{\bm{w}}$, ensuring the constraints of Eq.~\ref{eqt:3}, and $\bm{\Sigma} = \sigma^2 \bm{I}_d$ for some $\sigma > 0$. Hence, $\bm{\Sigma}_{\overline{\bm{w}}}=\sigma^2 \bm{P}_{\overline{\bm{w}}}$ and $\bm{\Sigma}_{\overline{\bm{w}}}^{1/2}=\sigma \bm{P}_{\overline{\bm{w}}}$.
\end{assumption}

With these assumptions in place, we analyze three training protocols and present their corresponding discrete dynamics, SDEs, and stationary distributions.
In the main text, we report only the expressions characterizing the stationary distributions; detailed derivations are deferred to Appendix~D.

\subsubsection{SGD on a Fixed Sphere}
A natural way to handle scale-invariant parameters is to optimize them directly on their intrinsic domain, a sphere of fixed radius.
This can be implemented by projecting the weights onto the sphere after each update~\cite{kodryan2022training,loshchilov2025ngpt} or by using Riemannian optimization methods~\cite{riemannian_bn}.
Here, we consider projected SGD on the sphere $\mathbb{S}^{d-1}(r)$ with discrete dynamics
\begin{equation}
    \label{eqt:4}
    \bm{w}_{k+1} = \text{proj}_{\mathbb{S}^{d-1}(r)} \Big(\bm{w}_k - \eta \nabla L_{\mathcal{B}_k}(\bm{w}_k)\Big),
\end{equation}
where $\eta$ is the learning rate and $\mathcal{B}_k$ is the mini-batch sampled at iteration $k$.
Taking the continuous-time limit of Eq.~\ref{eqt:4} yields the SDE governing the direction vector $\overline{\bm{W}}_t$:
\begin{equation}
    \label{eqt:5}
    \resizebox{.91\linewidth}{!}{$
        \dd \overline{\bm{W}}_t = -\bigg[\eta_{\eff}\nabla L(\overline{\bm{W}_t}) + \frac{\eta_{\eff}^2}{2} \Tr \big(\bm{\Sigma}_{\overline{\bm{W}}_t}\big) \overline{\bm{W}}_t\bigg]\dd t + \eta_{\eff} \big(\bm{\Sigma}_{\bm{\overline{W}}_t}\big)^{\frac{1}{2}}\dd\bm{B}_t,
    $}
\end{equation}
where $\bm{B}_t$ denotes standard Brownian motion in $\mathbb{R}^{d}$ and $\eta_{\eff} = \eta/r^2$ is the effective learning rate (ELR).
While~\cite{wang22three} analyze the Riemannian version of this SDE, we remain with the Euclidean formulation.

\begin{proposition}[Fixed sphere]
    Under Assumptions~\ref{asm:1}-\ref{asm:4}, the stationary distribution on the unit sphere $\rho^*_{\overline{\bm{w}}}$ of Eq.~\ref{eqt:5} is
    \begin{equation}
        \label{eqt:6}
        \rho^*_{\overline{\bm{w}}}(\overline{\bm{w}}) \propto \exp\Big(-\frac{L(\overline{\bm{w}})}{\tau_{\eff}}\Big), \; \mathrm{with} \; \tau_{\eff} = \frac{\eta_{\eff}\sigma^2}{2}
    \end{equation}
\end{proposition}

\subsubsection{SGD With Fixed ELR}
We next allow the radius to evolve and explicitly include weight decay into the dynamics.
We consider training with a fixed ELR, which makes the loss dynamics independent of the parameter norm.
This is achieved by scaling the learning rate with the squared norm of the parameters, $\eta = \eta_{\eff}\|\bm{w}_k\|^2$.
Although less common, such ELR control has been used in practice~\cite{elr_rl,weight_dynamics_normalized}, for example to enhance plasticity in reinforcement learning.
The discrete update rule is
\begin{equation}
    \label{eqt:7}
    \bm{w}_{k+1} = \bm{w}_k - \eta_{\eff} \|\bm{w}_k\|^2 \Big(\nabla L_{\mathcal{B}_k} (\bm{w}_k) + \lambda \bm{w}_k \Big),
\end{equation}
where $\lambda$ is the weight decay (WD) coefficient. The corresponding SDE is
\begin{equation}
    \label{eqt:8}
    \resizebox{.91\linewidth}{!}{$
        \dd\bm{W}_t = -\eta_{\eff} \|\bm{W}_t\|^2 \Big(\nabla L(\bm{W}_t) + \lambda \bm{W}_t\Big)\dd t + \eta_{\eff} \|\bm{W}_t\|^2 \big(\bm{\Sigma}_{\bm{W}_t}\big)^{\frac{1}{2}}\dd\bm{B}_t
    $}
\end{equation}

Applying Ito's formula separates the dynamics of the direction vector $\overline{\bm{W}}_t$ and the radius $r_t$:
\begin{gather}
    \label{eqt:9}
    \resizebox{.91\linewidth}{!}{$
        \dd \overline{\bm{W}}_t = -\bigg[\eta_{\eff}\nabla L(\overline{\bm{W}_t}) + \frac{\eta_{\eff}^2}{2} \Tr \bm{\Sigma}_{\overline{\bm{W}}_t} \cdot \overline{\bm{W}}_t\bigg]\dd t + \eta_{\eff} \big(\bm{\Sigma}_{\bm{\overline{W}}_t}\big)^{\frac{1}{2}}\dd\bm{B}_t 
    $} \\
    \label{eqt:10}
    \resizebox{.55\linewidth}{!}{$
        \frac{\dd r_t}{\dd t} = -\eta_{\eff}\lambda r_t^3 + \frac{\eta_{\eff}^2}{2}r_t \Tr \bm{\Sigma}_{\overline{\bm{W}}_t}
    $}
\end{gather}
\noindent
Importantly, the radius follows an ordinary differential equation and therefore evolves deterministically.
Under the isotropic noise model, $\Tr \bm{\Sigma}_{\overline{\bm{W}}_t} = \sigma^2(d-1)$ is constant.
Consequently, the system converges to a stationary distribution for $\overline{\bm{w}}$ on the unit sphere and a fixed stationary radius $r^*$.
\begin{proposition}[Fixed ELR]
    Under Assumptions~\ref{asm:1}-\ref{asm:4}, the stationary distribution $\rho^*_{\overline{\bm{w}}}$ on the unit sphere of Eq.~\ref{eqt:9} and stationary radius $r^*$ of Eq.~\ref{eqt:10} are
    \begin{align}
        \label{eqt:11}
        \rho^*_{\overline{\bm{w}}}(\overline{\bm{w}}) &\propto \exp\Big(-\frac{L(\overline{\bm{w}})}{\tau_{\eff}}\Big), \; \mathrm{with} \; \tau_{\eff} = \frac{\eta_{\eff}\sigma^2}{2} \\
        \label{eqt:12}
        r^* &= \sqrt{\frac{\tau_{\eff}(d-1)}{\lambda}} = \sqrt{\frac{\eta_{\eff}\sigma^2(d-1)}{2\lambda}}
    \end{align}
\end{proposition}

\subsubsection{SGD With Fixed LR}
Finally, we consider standard training with a fixed LR.
In this case, the dynamics on the unit sphere depends explicitly on the parameter norm.
The SGD update is
\begin{equation}
    \label{eqt:13}
    \bm{w}_{k+1} = \bm{w}_k - \eta \Big(\nabla L_{\mathcal{B}_k} (\bm{w}_k) + \lambda \bm{w}_k \Big),
\end{equation}
which corresponds to the SDE
\begin{equation}
    \label{eqt:14}
    \dd\bm{W}_t = -\eta \Big(\nabla L(\bm{W}_t) + \lambda \bm{W}_t\Big)\dd t + \eta \big(\bm{\Sigma}_{\bm{W}_t}\big)^{\frac{1}{2}}\dd\bm{B}_t
\end{equation}
The induced direction $\overline{\bm{W}}_t$ and radius $r_t$ dynamics are
\begin{gather}
    \label{eqt:15}
    \resizebox{.95\linewidth}{!}{$
        \dd \overline{\bm{W}}_t = -\bigg[\frac{\eta}{r_t^2}\nabla L(\overline{\bm{W}_t}) + \frac{\eta^2}{2 r_t^4} \Tr \bm{\Sigma}_{\overline{\bm{W}}_t} \cdot \overline{\bm{W}}_t\bigg]\dd t + \frac{\eta}{r_t^2} \big(\bm{\Sigma}_{\bm{\overline{W}}_t}\big)^{\frac{1}{2}}\dd\bm{B}_t
    $} \\
    \label{eqt:16}
    \resizebox{.55\linewidth}{!}{$
        \frac{\dd r_t}{\dd t} = -\eta\lambda r_t + \frac{\eta^2}{2r_t^3} \Tr \bm{\Sigma}_{\overline{\bm{W}}_t}
    $}
\end{gather}
As in the fixed-ELR case, the radius evolves deterministically and converges to a fixed stationary value $r^*$.
\begin{proposition}[Fixed LR]
    Under Assumptions~\ref{asm:1}-\ref{asm:4}, the stationary distribution $\rho^*_{\overline{\bm{w}}}$ of Eq.~\ref{eqt:15} and stationary radius $r^*$ of Eq.~\ref{eqt:16} are
    \begin{align}
        \label{eqt:17}
        \rho^*_{\overline{\bm{w}}}(\overline{\bm{w}}) &\propto \exp\Big(-\frac{L(\overline{\bm{w}})}{\tau/(r^*)^2}\Big), \; \mathrm{with} \; \tau = \frac{\eta\sigma^2}{2} \\
        \label{eqt:18}
        r^* &= \sqrt[4]{\frac{\tau (d-1)}{\lambda}} = \sqrt[4]{\frac{\eta \sigma^2 (d-1)}{2 \lambda}}
    \end{align}
\end{proposition}

\section{Thermodynamic Framework}

In this section, we introduce optimization quantities that correspond to thermodynamic state variables in~\hyperref[T1]{\bf{T1}}.
Our framework is based on the stationary distribution of the weights, $\rho^*_{\bm{w}}$, which we relate to the thermodynamic Gibbs distribution~(\ref{T3}).
This correspondence naturally identifies thermodynamic microstates $i$ with weight vectors $\bm{w}$.

We then define the analogues of temperature $T$, pressure $p$, and volume $V$ so that the appropriate thermodynamic potential, Helmholtz energy $F$~(\ref{T4}) or Gibbs energy $G$~(\ref{T5}), is minimized at stationary.
Concretely, we associate $T$ with the noise level through the ELR and gradient variance, $V$ with the radius, and $p$ with the L2 regularization strength.
This correspondence reflects an intuition that higher pressure leads to smaller volume, just as stronger weight decay leads to smaller parameter norms.
These interpretations are confirmed by explicit minimization of the corresponding potentials.

\subsubsection{Microstates and Weights}
Under Assumptions~\ref{asm:1}-\ref{asm:4}, the stationary distribution $\rho^*_{\bm{w}}$ is supported on a sphere $\mathbb{S}^{d-1}(r^*)$. 
The radius $r^*$ is either pre-defined (fixed-sphere protocol) or determined by the training hyperparameters (fixed-ELR and fixed-LR protocols). 
In this setting, the density on $\mathbb{S}^{d-1}(r^*)$ can be expressed in terms of a density on the unit sphere.
\begin{lemma}
\label{lemma:1}
Let $\rho_{\bm{w}}$ be a density on the sphere $\mathbb{S}^{d-1}(r)$. Then,
\begin{equation}
    \rho_{\bm{w}} (\bm{w}) = \rho_{\overline{\bm{w}}} (\overline{\bm{w}})/r^{d-1},
\end{equation}
where $\rho_{\overline{\bm{w}}}$ is the corresponding density on the unit sphere.
The associated spherical entropy $S(\rho_{\bm{w}}) = \mathbb{E}_{\rho_{\bm{w}}}[-\log \rho_{\bm{w}}(\bm{w})]$ is
\begin{equation}
    S(\rho_{\bm{w}}) = S(\rho_{\overline{\bm{w}}}) + (d-1)\log r,
\end{equation}
with $S(\rho_{\overline{\bm{w}}})$ denoting the spherical entropy of $\rho_{\overline{\bm{w}}}$.
\end{lemma}

As a consequence, $\rho^*_{\bm{w}}$ has the same functional form as $\rho^*_{\overline{\bm{w}}}$ and takes the Gibbs form $\rho^*_{\bm{w}} \propto \exp(-L(\bm{w})/T)$ for all three training protocols (Eqs.~\ref{eqt:6}, \ref{eqt:11}, \ref{eqt:17}).
This mirrors the equilibrium distribution of a thermodynamic system, $p_i \propto \exp(-E_i/T)$ (\ref{T3}), motivating the identifications $i\leftrightarrow\bm{w}$, $E_i\leftrightarrow L(\bm{w})$, $U \leftrightarrow U(\rho_{\bm{w}}) = \mathbb{E}_{\rho_{\bm{w}}}[L(\bm{w})]$, and $S \leftrightarrow S(\rho_{\bm{w}})$.

\subsubsection{Thermodynamic Potentials}
We now define $T$, $p$, and $V$ through the requirement that the corresponding potential is minimized at stationary.

\begin{theorem}[Helmholtz energy minimization]
\label{thm:1}
Under Assumptions~\ref{asm:1}-\ref{asm:4}, consider SGD on a fixed sphere with radius $r$ and ELR $\eta_{\eff}$. Defining $T=\frac{\eta_{\eff}\sigma^2}{2}$, the Helmholtz energy $F=U-TS$ is minimized at stationary distribution $\rho_{\overline{\bm{w}}}$ among all distributions on the unit sphere $\mathbb{S}^{d-1}$.
\end{theorem}

\begin{theorem}[Gibbs energy minimization]
\label{thm:2}
Under Assumptions~\ref{asm:1}-\ref{asm:4}, consider SGD with a fixed ELR $\eta_{\eff}$ (or a fixed LR $\eta$) and WD $\lambda$. Let $T=\frac{\eta_{\eff}\sigma^2}{2}$ (or $T=\sqrt{\frac{\eta\lambda\sigma^2}{2(d-1)}}$), $p=\lambda$ and $V=\frac{r^2}{2}$. Then, the Gibbs energy $G=U-TS+pV$ is minimized at the stationary distribution $\rho_{\bm{w}}$ among all distributions supported on a sphere $\mathbb{S}^{d-1}(r)$ with deterministic radius $r>0$.
\end{theorem}

The proofs and a detailed discussion of the choice of $p$ and $V$ are provided in Appendix~D. 
Importantly, the choice of potential follows thermodynamics: $F$ applies when the volume (radius) is fixed, while $G$ applies when the pressure (weight decay coefficient) is fixed.
In both cases, the temperature $T$ is set by constant hyperparameters and thus remains fixed.
Therefore, our framework offers an alternative thermodynamic description of SGD stationary distributions\footnote{Instead of Assumption~\ref{asm:3}, one may postulate (1) minimization of the potentials at stationary and (2) deterministic radius (while preserving Assumptions~\ref{asm:1}, \ref{asm:2}, \ref{asm:4}), and obtain the same stationary distributions without appealing to the SDE.}.

\subsubsection{Ideal Gas Law}
Remarkably, inserting the definitions of $T$, $p$, and $V$ into the stationary radius $r^*$ (Eq.~\ref{eqt:12}, \ref{eqt:18}) yields exactly the ideal gas law $pV=RT$ (\ref{T6}) with $R=\frac{d-1}{2}$ both for fixed-ELR (Eq.~\ref{eqt:21}) and fixed-LR (Eq.~\ref{eqt:22}) training protocols.
A similar expression also holds in the fixed-sphere case if we introduce ``effective'' weight decay as pressure (see Appendix~D.4).
\begin{align}
    V &= \frac{(r^*)^2}{2} = \frac{\eta_{\eff}\sigma^2}{2} \cdot \frac{d-1}{2\lambda} = \frac{RT}{p} \label{eqt:21} \\
    V &= \frac{(r^*)^2}{2} = \frac12 \sqrt{\frac{\eta \lambda \sigma^2}{2 (d-1)}} \frac{d-1}{2\lambda} = \frac{RT}{p} \label{eqt:22}
\end{align}

This correspondence is particularly striking given the very different microscopic assumptions.
In statistical mechanics, the ideal gas law assumes non-interacting particles, so the energy of a microstate $i$ decomposes as $E_i=\sum_j e_{ij}$, where $e_{ij}$ is the energy of particle $j$.
By contrast, the loss $L(\bm{w})$ typically exhibits strong parameter coupling and does not admit a decomposition of the form $L(\bm{w})=\sum_j l_j(w_j)$.
Nevertheless, under scale invariance, the stationary optimization dynamics obey the same equation of state.
While such an assumption would be unnatural in physical systems, requiring interactions to depend on angular coordinates rather than distances, it is natural for neural networks and yields the ideal gas law despite strong parameter interactions.

\section{Empirical Validation}
To support our theoretical framework, we design experiments inspired by the thermodynamic analogy.
We focus on tests that do not follow directly from the SDE formulation, providing strong evidence for the thermodynamic perspective.

\paragraph{(V1) Stationary radius.}\label{V1}
The SDE predicts that $r^*$ scales as $\sqrt[4]{\eta/\lambda}$ when training with a fixed LR, and as $\sqrt{\eta_{\eff}/\lambda}$ when training with a fixed ELR.
Since this prediction follows directly from the SDE (without invoking thermodynamics), it primarily serves as a sanity check.
At the same time, it also verifies the ideal gas law (\ref{T6}) in our analogy.

\paragraph{(V2) Minimizing thermodynamic potentials.}\label{V2}
Our analogy predicts that stationary distribution of SGD minimizes the appropriate thermodynamic potential.
Numerical experiments cannot test minimization over the entire space of distributions $\rho_{\overline{\bm{w}}}$ and radii $r^*$.
Instead, we check whether the observed stationary states minimize $F$ or $G$ at least among the distributions induced by different hyperparameter settings.

For example, in fixed LR training, consider a set of hyperparameter configurations $\mathcal{H} = \{(\eta_i, \lambda_i)\}_{i=1}^{h}$.
For each $(\eta_i, \lambda_i)$, we train the model and measure the corresponding internal energy $U_i$, entropy $S_i$, and volume $V_i$.
We then verify that for each $(\eta^*, \lambda^*) \in \mathcal{H}$:
\begin{equation}
    (\eta^*, \lambda^*) = \argmin_{(\eta_i, \lambda_i) \in \mathcal{H}} \Big\{U_i - T^* S_i + p^* V_i\Big\},
\end{equation}
where $T^* = \sqrt{\frac{\eta^* \lambda^* \sigma^2}{2(d-1)}}$ and $p^* = \lambda^*$.
A similar procedure applies for the settings with a fixed ELR and on a fixed sphere.

\paragraph{(V3) Maxwell relations.}\label{V3}
Maxwell relations describe how entropy $S$ varies when thermodynamic variables change (\ref{T4}, \ref{T5}).
In our setting, training with fixed LR corresponds to fixing $p$ and $T$, so the relevant Maxwell relation is $\big(\frac{\partial S}{\partial p}\big)_T = \big(\frac{\partial V}{\partial T}\big)_p$.
In Appendix~D.4, 
we show this is equivalent to
\begin{equation}
    \label{eq:25}\Big(\frac{\partial S}{\partial \log \eta}\Big)_{\lambda} - \Big(\frac{\partial S}{\partial \log \lambda}\Big)_{\eta} = \frac{d-1}{2}
\end{equation}
This elegant equation quantifies, how the two hyperparameters influence the stationary entropy.
Similar relations for the fixed sphere and fixed ELR protocols are also derived there.

\paragraph{(V4) Adiabatic process. }\label{V4}
Previously, we interpreted convergence to a stationary distribution as a non-reversible thermodynamic process.
Here, we instead view stationary distributions corresponding to different hyperparameters as states of a \emph{reversible} process.
This perspective enables a thermodynamic analysis of how hyperparameters shape the stationary distribution.
We focus on the adiabatic process, defined by $\delta Q = 0$ (\ref{T7}). 
Although heat $Q$ does not appear explicitly in our analogy, it can be inferred from the First Law (\ref{T2}) as $\delta Q = \dd U + p \dd V$.
This allows us to define the isochoric and isobaric heat capacities, $C_V$ and $C_p$, as in \ref{T7}.
In Appendix~D.5, 
we show that $C_p - C_V = \tfrac{d-1}{2} = R$, matching the ideal gas relation.
To simulate an adiabatic process, we vary $(\eta, \lambda)$ jointly such that $p V^{\gamma} = \text{const}$, where $\gamma = C_p / C_V$.
For a reversible adiabatic process, $\dd S = \delta Q / T = 0$ (\ref{T2}), implying constant entropy.
Empirically, this predicts that varying hyperparameters along an adiabatic process gives stationary distributions of different shape supported on spheres of different radii but with identical entropy, i.e., overall ``disorder.''

\begin{figure*}[t]
    \centering
    \includegraphics[width=0.93\textwidth]{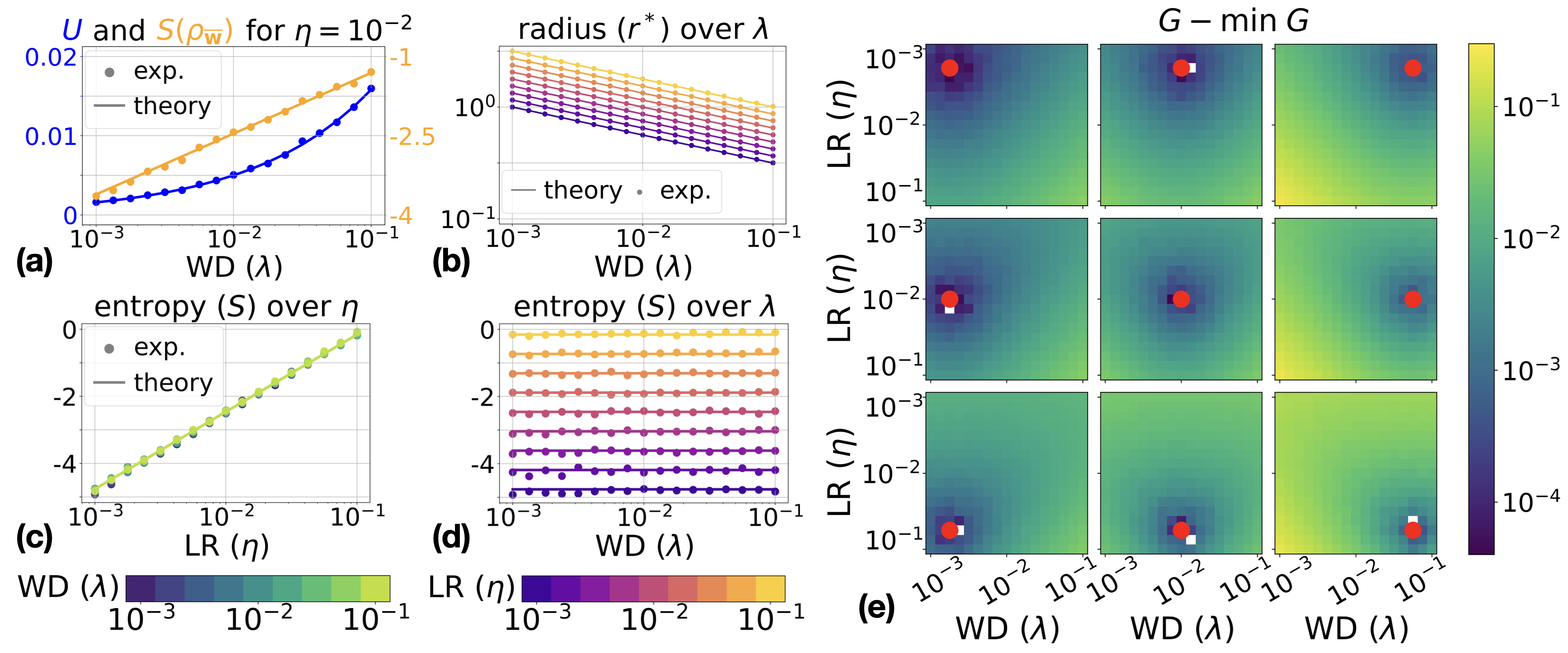}
    \caption{
    VMF isotropic noise model with fixed LR~$\eta$ and WD~$\lambda$. Subfigures a--d: points are numerical measurements, solid lines are theoretical predictions: $U=\frac{d-1}{2}T$, $S(\rho_{\overline{\bm{w}}})=\frac{d-1}{2}\log(2\pi eT)$, $S = S(\rho_{\overline{\bm{w}}}) + (d-1)\log r^*$, and $r^*=\sqrt{\frac{T(d-1)}{p}}$, with $T=\sqrt{\frac{\eta\lambda\sigma^2}{2(d-1)}}$ and $p=\lambda$. Subfigure e: Gibbs energy minimization (\hyperref[V2]{\bf{V2}}). Each subplot corresponds to a fixed pair $(\eta^*, \lambda^*)$, denoted with red circle. The colormap shows the difference between $G$ and its minimum across stationary distributions, with the minimizer marked by a white square. Ideally, red circles coincide with white squares; in practice, they either match or lie very close.}
    \label{fig:isotropic_vmf_lr}
\end{figure*}

\section{Isotropic Noise Model Experiments}
\label{sec:toy}

\paragraph{Experimental setup.} 
We start by validating the isotropic noise model through numerical simulation.
Here, we show results for fixed LR, while other training protocols are presented in Appendix~E.
We consider a scale-invariant function $L(\bm{w}) = 1 + \frac{\bm{\mu}^T \bm{w}}{\|\bm{w}\|}$ with some fixed $\bm{\mu} \in \mathbb{R}^d, \|\bm{\mu}\|=1$ and train it using noisy gradient descent:
\begin{equation}
    \label{eq:24}
    \bm{w}_{k+1} = \bm{w}_k - \eta \Big(\nabla L (\bm{w}_k) + \frac{1}{\|\bm{w}_k\|} \bm{P}_{\overline{\bm{w}}_k} \sigma \bm{\varepsilon} + \lambda \bm{w}_k \Big),
\end{equation}
with $\bm{\varepsilon} \sim \mathcal{N}(0, \bm{I}_d)$.
We select this function because the SDE theory predicts that, at stationarity, the angular distribution follows a Von Mises-Fisher (VMF) form: $\rho_{\overline{\bm{w}}}(\overline{\bm{w}}) \propto \exp(-\bm{\mu}^T \overline{\bm{w}}/T)$.
For this distribution, the analytical values of $U$ and $S$ are known (see Appendix~E), 
allowing direct comparison with simulation results.
In this setting, we set $d=3$.

\paragraph{Entropy estimation.}
While calculating mean training loss $U$ is straightforward, estimating entropy $S$ is a more complicated task.
We rely on the nearest neighbor entropy estimator~\cite{KozLeo87}, which requires a sample from the stationary distribution.
In Appendix~D.6, 
we present a detailed entropy estimation protocol. 
In the experiments, we wait until the loss and radius stabilize during the training run.
Then, we maintain a queue of $1000$ weight vectors sampled every $50$ iterations along the training trajectory to reduce correlation between consecutive samples.

\paragraph{Results.}
In Figure~\ref{fig:isotropic_vmf_lr}, we present the results of numerical simulation.
Subfigure~\ref{fig:isotropic_vmf_lr}a shows that both the average loss $U$ and the entropy on the unit sphere $S(\rho_{\overline{\bm{w}}})$ match the analytical VMF predictions, providing a strong evidence, that this distribution is indeed stationary.
Subfigures~\ref{fig:isotropic_vmf_lr}b and~\ref{fig:isotropic_vmf_lr}e validate \hyperref[V1]{\bf{V1}} and \hyperref[V2]{\bf{V2}}, respectively.
To verify \hyperref[V3]{\bf{V3}}, we compute the partial derivatives of the total entropy $S$:
\begin{equation}
    \Big(\frac{\partial S}{\partial \log \eta}\Big)_{\lambda} = \frac{d-1}{2}, \qquad \Big(\frac{\partial S}{\partial \log \lambda}\Big)_{\eta} = 0
\end{equation}
Subfigures~\ref{fig:isotropic_vmf_lr}c,d show that empirical measurements closely follow these theoretic predictions.
For \hyperref[V4]{\bf{V4}}, we calculate the isochoric heat capacity
\begin{equation}
    C_V = \Big(\frac{\delta Q}{\dd T}\Big)_V = \Big(\frac{\partial U}{\partial T}\Big)_V = \frac{d-1}{2}
\end{equation}
Combining it with $C_p - C_V = \frac{d-1}{2}$, we get $\gamma=\frac{C_p}{C_V}=2$.
In the adiabatic process, we maintain $p V^\gamma = \text{const}$, which reduces to
\begin{equation}
    pV^\gamma \propto p^{1-\gamma}T^{\gamma} \propto \eta^{\frac{\gamma}{2}} \lambda^{1-\frac{\gamma}{2}} \Big|_{\gamma=2}\propto \eta
\end{equation}
Therefore, keeping $\eta$ fixed and varying $\lambda$ yields an adiabatic process with $\dd S = 0$.
This is indeed observed in the simulation (Subfigures~\ref{fig:isotropic_vmf_lr}c,d): the entropy depends only on $\eta$ and remains constant as $\lambda$ changes.
Note that this property is specific to the VMF distribution.
For other distributions $\rho_{\overline{\bm{w}}}$, the values of $C_V$, $C_p$, and $\gamma$ may differ, requiring coordinated changes of $\eta$ and $\lambda$ to achieve an adiabatic process.

\section{Neural Network Experiments}
\label{sec:neural_nets}

\subsection{Generalizing Isotropic Noise Model}
\label{sec:neural_nets:theory}

For neural networks, the isotropic noise assumption breaks down: the covariance matrix $\bm{\Sigma}_{\overline{\bm{w}}}$ is generally anisotropic and spatially dependent.
\cite{chaudhari2018stochastic} show that for general networks (i.e., not necessarily scale-invariant) trained with fixed LR $\eta$, the stationary distribution takes the form $\rho^*_{\bm{w}} \propto \exp(-\Phi(\bm{w})/T_0)$ with $T_0 \propto \eta$,
where $\Phi(\bm{w})$ is an implicit \emph{potential} determined by the loss $L(\bm{w})$ and the covariance matrix $\bm{\Sigma}_{\bm{w}}$, but independent of $\eta$.
In this setting, $\Phi$ replaces the training loss as the effective energy.
An explicit expression for $\Phi$ is known only in special cases, such as linear regression~\cite{kunin2023limiting}.
We extend this framework to scale-invariant models by restricting to the unit sphere and defining $\rho^*_{\overline{\bm{w}}}\propto \exp(-\Phi(\overline{\bm{w}})/T_0)$ with $T_0$ determined by training hyperparameters.
This distribution arises in all three training protocols introduced in Section~\ref{sec:sde}.

To extend the ideal gas analogy to neural networks, a well-defined stationary radius is required.
Following~\cite{li2020reconciling,wan2021spherical}, we introduce the constant covariance trace assumption, which ensures that Eq.~\ref{eqt:10} and~\ref{eqt:16} admit a well-defined limit $r^*$.
\begin{assumption}[Constant covariance trace]
    \label{asm:5}
    For all $\overline{\bm{w}}$, $\Tr \bm{\Sigma}_{\overline{\bm{w}}} = \mathrm{const}$ (while the covariance matrix $\bm{\Sigma}_{\overline{\bm{w}}}$ itself might be different). We then define $\sigma^2 = \frac{1}{d-1} \Tr \bm{\Sigma}_{\overline{\bm{w}}}$.
\end{assumption}
\noindent
Defining $T=\sigma^2 T_0$, minimization of $\mathbb{E}_{\rho_{\overline{\bm{w}}}} [\Phi(\overline{\bm{w}})] - T_0S(\rho_{\overline{\bm{w}}})$, achieved at $\rho^*_{\overline{\bm{w}}}$, is equivalent to minimizing
\begin{equation}
    U(\rho_{\overline{\bm{w}}}) - TS(\rho_{\overline{\bm{w}}}), \quad \text{with } U = \mathbb{E}_{\rho_{\overline{\bm{w}}}} [\sigma^2 \Phi(\overline{\bm{w}})],
\end{equation}
We therefore interpret this quantity as the Helmholtz energy $F$, minimized in the fixed-sphere case. 
Consequently, we define $G=U(\rho_{\overline{\bm{w}}}) - TS(\rho_{\bm{w}}) + pV(r)$, which corresponds to fixed-ELR and fixed-LR cases. 

To summarize, the thermodynamic analogy for neural networks differs from the isotropic noise model in two points:
\begin{enumerate}
    \item The covariance matrix $\bm{\Sigma}_{\overline{\bm{w}}}$ is anisotropic, so the energy function is not the training loss $L(\overline{\bm{w}})$, but a potential $\sigma^2 \Phi(\overline{\bm{w}})$ with no explicit form.
    \item The formulae for temperature $T$ and stationary radius $r^*$ remain the same as in Section~\ref{sec:sde}, but $\sigma^2 = \frac{1}{d-1} \Tr \bm{\Sigma}_{\overline{\bm{w}}}$ replaces the isotropic variance.
\end{enumerate}
Thus, under Assumption~\ref{asm:5}, the ideal gas laws still hold, and we expect the tests \hyperref[V1]{\bf{V1}}--\hyperref[V4]{\bf{V4}} to hold.
However, since $\Phi$ is unknown, we can only directly verify \hyperref[V1]{\bf{V1}} and \hyperref[V3]{\bf{V3}}.

\begin{figure*}[t]
    \centering
    \includegraphics[width=0.93\textwidth]{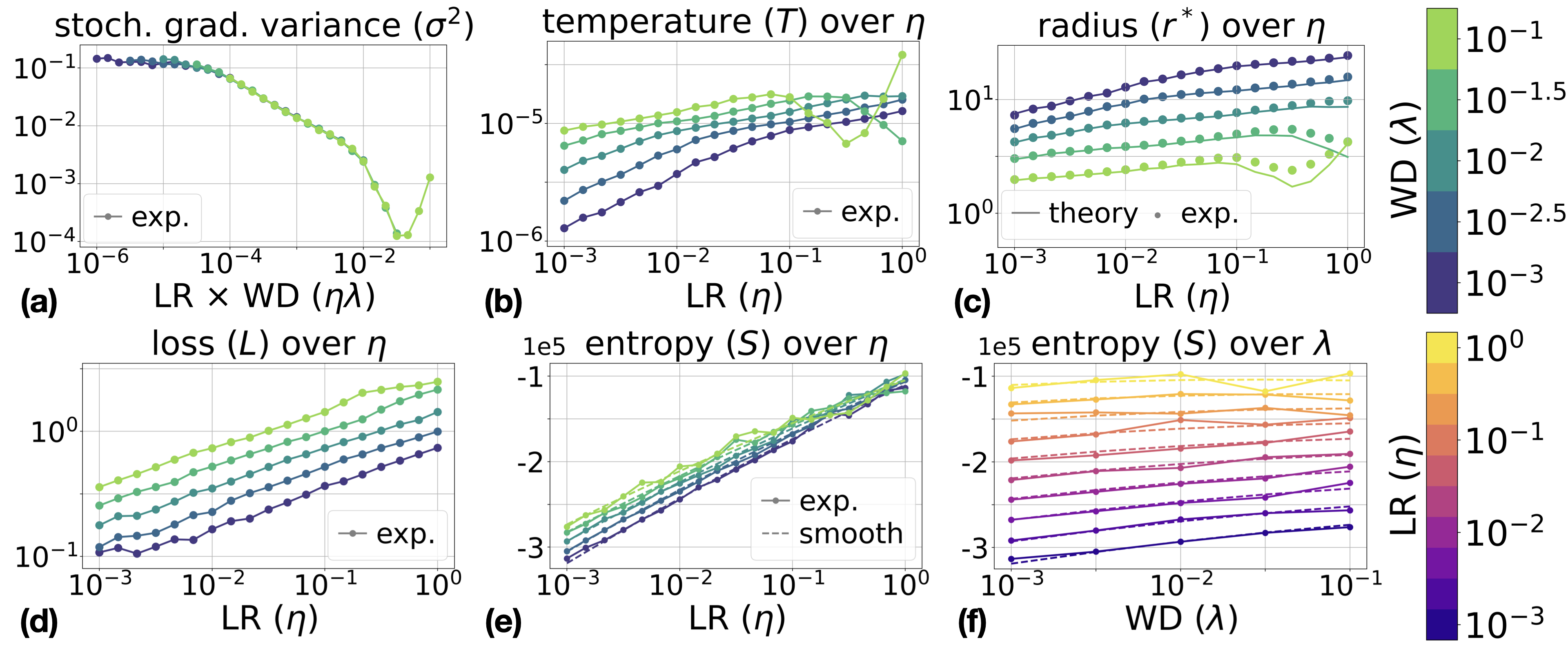}
    \caption{
    ResNet-18 on CIFAR-10 with fixed LR~$\eta$ and WD~$\lambda$. Subfigures a, b, d: empirically measured $\sigma^2$, mean loss $L$, and temperature $T=\sqrt{\frac{\eta\lambda\sigma^2}{2(d-1)}}$, respectively. Subfigure c: stationary radius $r^*=\sqrt{\frac{T(d-1)}{p}}$ (solid lines, theory) vs. experimental values (points). Subfigures e and f: entropy $S$ as a function of $\eta$ and $\lambda$; solid lines with markers show experimental estimates, dashed lines their smoothed versions.}
    \label{fig:resnet_cifar10_lr}
\end{figure*}

\subsection{Experimental Setup}
\label{sec:neural_nets:exp_setup}
We train ResNet-18~\cite{deep_resnet} on CIFAR-10~\cite{cifar10} with varying LR $\eta$ and WD $\lambda$, fixed batch size $B=128$, for $t=10^6$ iterations.
The entropy estimation queue stores $1000$ vectors with new weights added every $25$ iterations.
All models are made fully scale-invariant following~\cite{li2020exponential}.
We also use a ``thin'' model with the width multiplier $k=4$ (while the default value is $k=64$), the resulting number of trainable parameters is $d=43692$.
This choice is motivated by (1) memory constraints for storing the entropy estimation queue, and (2) avoiding \emph{overparameterization}, which imposes a specific behavior of stochastic gradients for small LRs later in training (which advocates convergence to loss minimum, as discussed in Appendix~I).
In the main text, we show the results only for the fixed LR case, two other training protocols are presented in the Appendix~F, 
supported by the results for the ConvNet architecture from~\cite{kodryan2022training} and the CIFAR-100 dataset~\cite{cifar100}.
In Appendix~G, 
we also experiment with partially scale-invariant models. 

Estimating entropy in high dimensions is challenging: the nearest-neighbor estimator is unbiased only asymptotically, with a bias of order $\mathcal{O}(N^{-2/d})$~\cite{DELATTRE2017}.
We assume the bias is roughly constant across different stationary distributions, allowing accurate reconstruction of entropy derivatives for Maxwell relations.
Empirically, \hyperref[V3]{\bf{V3}} holds with high precision, supporting this assumption.

\subsection{Results}
We present the results in Figure~\ref{fig:resnet_cifar10_lr}.
Subfigure~\ref{fig:resnet_cifar10_lr}a shows that $\sigma^2$ varies across stationary distributions.
While theory assumes a fixed $\sigma^2$, in practice it depends on the hyperparameters, primarily the product $\eta\lambda$.
We therefore define temperature as $T=\sqrt{\frac{\eta\lambda\sigma^2}{2(d-1)}}$ using the measured $\sigma^2$, with values shown in Subfigure~\ref{fig:resnet_cifar10_lr}b.
$T$ generally increases with $\eta$ and $\lambda$, but for large $\lambda$ it can be non-monotonic, e.g., at $\lambda = 10^{-1}$, $T$ decreases for $10^{-1} \le \eta \le 10^{-0.5}$ and rises again for $\eta \ge 10^{-0.5}$, reflecting deviations in $\sigma^2$ from the trend seen for $\eta\lambda < 10^{-2}$.
These results suggest that, for large $\eta \lambda$, stationary distributions explore different regions of the weight space, consistent with prior observations by \cite{sadrtdinov2024where}, who analyze the loss landscape under large LRs.

Subfigure~\ref{fig:resnet_cifar10_lr}c shows that the stationary radius $r^*$ closely follows theoretical predictions, confirming \hyperref[V1]{\bf{V1}}.
Deviations appear only at large $\eta$ and $\lambda$, due to \emph{discretization error} in the SDE: the continuous model neglects the norm of the full-batch gradient, which acts like a centrifugal force and enlarges $r^*$.
In Appendix~H, 
we derive a correction that predicts $r^*$ more accurately for larger values of $\eta$ and $\lambda$.

Finally, to verify \hyperref[V3]{\bf{V3}}, we represent $S$ as a function of $\log \eta$ and $\log \lambda$, as shown in Subfigures~\ref{fig:resnet_cifar10_lr}e and f.
The empirical estimates are noisy, so we smooth them using polynomial regression with quadratic terms (see details in Appendix F). 
This approximation achieves a coefficient of determination of $R^2 \approx 0.99$.
We then substitute the partial derivatives of the approximation into Eq.~\ref{eq:25} and obtain a maximal absolute error of $\approx 3\%$ between the left- and right-hand sides, indicating that the Maxwell relation holds with high precision.

\section{Discussion}

\subsubsection{Extending Ideal Gas Analogy}
Answering the question posed in the title, when the assumption $\Tr \bm{\Sigma}_{\overline{\bm{w}}}=\text{const}$ holds, the stationary behavior of SGD is fully consistent with the ideal gas model, as in the isotropic noise setting.
In practice, however, we observe that the value of $\sigma^2$ varies with training hyperparameters.
Nevertheless, our experiments show that \hyperref[V1]{\bf{V1}} and \hyperref[V3]{\bf{V3}} remain valid.

Developing a more general thermodynamic description that accounts for variable $\Tr \bm{\Sigma}_{\overline{\bm{w}}}$ is an interesting direction for future work.
One possibility is to draw an analogy with a \emph{real gas}, described by the state equation $V = Z(p,T)\frac{RT}{p}$, where $Z$ is the compressibility factor capturing deviations from the ideal gas.
This mirrors the relation $\frac{(r^*)^2}{2} = \Tr \bm{\Sigma}_{\overline{\bm{w}}} \frac{\eta_{\eff}}{4\lambda}$ (from Eq.~\ref{eqt:10}), suggesting that extending the analogy to real gases may be as natural for scale-invariant networks as the ideal gas correspondence established here.
Another promising direction is extending the framework beyond fully scale-invariant networks and to more complex optimizers, such as SGD with momentum or Adam~\cite{adam}, which could yield insights directly relevant to practical training.

\subsubsection{Practical Implications}
Our work adopts a thermodynamic perspective to clarify how training hyperparameters, particularly learning rate and weight decay, shape the stationary distribution of SGD.
Since neural networks are rarely trained with a fixed learning rate, these insights may inform the design of more effective LR schedulers~\cite{liu2025neural}.
A key outcome of our framework is the Maxwell relations, which suggest a possible approach to explicitly control the rate of entropy decay.
Properly regulating entropy decay may help reduce training loss while avoiding premature convergence to sharp minima often associated with poor generalization~\cite{kodryan2022training}.

Moreover, hyperparameters determine the effective temperature, governing the trade-off between internal energy and entropy.
This balance is crucial for weight averaging~\cite{izmailov2018averaging}, which improves performance by combining multiple low-loss checkpoints.
While low loss is necessary for accuracy, high entropy promotes diversity of solutions and thus effective averaging, making hyperparameter selection non-trivial.
Indeed, \cite{sadrtdinov2024where} show that LRs optimal for weight averaging often exceed the convergence threshold.
Hence, understanding how entropy dynamics affect weight averaging is a promising direction for future work, with Maxwell relations providing a quantitative tool for both theoretical and empirical analysis.

\section{Conclusion}

In this work, we introduced and empirically validated a thermodynamic framework for describing the stationary distributions of SGD in scale-invariant neural networks. 
We defined macroscopic thermodynamic variables and related them to learning rate, weight decay, and parameter norm. 
We rigorously showed that, under the simplified isotropic noise model, stationary distributions follow ideal gas laws. 
Importantly, the key predictions of this framework also hold in experiments with neural networks.
Future work may extend this analogy to settings with variable noise, non-scale-invariant architectures, or more complex optimizers, and may further support principled approaches to hyperparameter tuning for learning rate scheduling and weight averaging.  

\section*{Acknowledgments}
We would like to thank Alexander Shabalin for insightful conversations and personal support.
Ekaterina Lobacheva was supported by IVADO and the Canada First Research Excellence Fund.
The authors gratefully acknowledge the computing time made available to them on the high-performance computer at the NHR Center of TU Dresden.
This center is jointly supported by the Federal Ministry of Research, Technology and Space of Germany and the state governments participating in the NHR (\url{www.nhr-verein.de/unsere-partner}).
The empirical results were also enabled by compute resources and technical support provided by Mila - Quebec AI Institute (mila.quebec).

\bibliographystyle{named}
\bibliography{ref}

\newpage
\appendix
\onecolumn

\begin{center}
    \LARGE \bf
    Can Stationary Distributions of Scale-Invariant Neural Networks \\ Be Described by the Thermodynamics of an Ideal Gas? \\
    Technical Appendix
\end{center}

\section{Table of notations}
\label{app:table_of_notations}

In this section, we present Table~\ref{tab:math} with general mathematical notations used in the paper and Table~\ref{tab:analogy}, which relates optimizational and thermodynamic quantities. 

\begin{table}[h]
\renewcommand{\arraystretch}{1.1}
\caption{General mathematical notations used throughout the paper.} \label{tab:math}
\begin{center}
\begin{tabular}{cc}
\toprule
\textbf{Object} & \textbf{Notation} \\
\toprule
\makecell{dimensionality (number of trainable parameters)} & $d$\\
identity matrix & $\bm{I}_d$ \\
\makecell{real Euclidian space} & $\mathbb{R}^d$ \\
\makecell{unit sphere embedded in $\mathbb{R}^d$} & $\mathbb{S}^{d-1}$ \\
\makecell{sphere of radius $r$ embedded in $\mathbb{R}^d$} & $\mathbb{S}^{d-1}(r)$ \\
orthogonal projection matrix & $\bm{P}_{\bm{x}} = \bm{I}_d - \frac{\bm{x}\bm{x}^T}{\|\bm{x}\|^2}$ \\
standard Gaussian random vector & $\bm{\varepsilon} \sim \mathcal{N}(0, \bm{I}_d)$ \\
standard Brownian motion in $\mathbb{R}^d$ & $\bm{B}_t$ \\
proportional to & $\propto$ \\
matrix trace & $\Tr$ \\
angular vector (spherical coordinates) & $\bm{\theta} = \{\theta_1, \dots, \theta_{d-2}, \phi\}$ \\
spherical coordinates transformation Jacobian & $J(\bm{\theta})$ 
\end{tabular}
\end{center}
\end{table}

\begin{table}[]
\renewcommand{\arraystretch}{1.1}
\caption{Notations used throughout the paper. Left column shows quantities from optimization, right column presents analogous variables from thermodynamics (if applicable).} \label{tab:analogy}
\begin{center}
\begin{tabular}{cc}
\textsc{Optimization} & \textsc{Thermodynamics} \\
\toprule
\multicolumn{2}{c}{\textbf{Weight vector and microstates}} \\
\toprule
weight vector $\bm{w}$ & microstate $i$ \\
unit direction vector $\overline{\bm{w}} = \bm{w}/\|\bm{w}\|$ & --- \\ 
\midrule
isotropic noise model: scale-invariant loss function $L(\bm{w}) = L(\overline{\bm{w}})$ & \multirow{2}{*}{microstate energy $E_i$} \\
anisotropic noise model: potential $\Phi(\overline{\bm{w}})$ & \\
\midrule
radius $r = \|\bm{w}\|$ & --- \\
stationary radius $r^*$ & --- \\
\toprule
\multicolumn{2}{c}{\textbf{Stationary distribution, internal energy and entropy}} \\
\toprule
stationary distribution (on $\mathbb{S}^{d-1}$) $\rho_{\overline{\bm{w}}} (\overline{\bm{w}}) \propto \exp\bigl(-\frac{L(\overline{\bm{w}})}{T}\bigr)$ & --- \\
stationary distribution (on $\mathbb{S}^{d-1}(r^*)$) $\rho_{\bm{w}} (\bm{w}) = \rho_{\overline{\bm{w}}} (\overline{\bm{w}})\delta(r -r^*) $ &  Gibbs distribution $p_i \propto \exp \bigl(-\frac{E_i}{T}\bigr)$ \\
internal energy $U = \mathbb{E}_{\rho_{\overline{\bm{w}}}} [L(\overline{\bm{w}})]$ & internal energy $U = \mathbb{E}_{p_i} [E_i]$ \\
entropy (on $\mathbb{S}^{d-1}$) $S(\rho_{\overline{\bm{w}}}) = \mathbb{E}_{\rho_{\overline{\bm{w}}}} [-\log \rho_{\overline{\bm{w}}}]$ & --- \\
entropy (on $\mathbb{S}^{d-1}(r^*)$) $S(\rho_{\bm{w}}) = S(\rho_{\overline{\bm{w}}}) + (d-1) \log r^*$ & entropy $S = \mathbb{E}_{p_i} [-\log p_i]$ \\
\toprule
\multicolumn{2}{c}{\textbf{Variance of stochastic gradients}} \\
\toprule
covariance matrix of stochastic gradients $\bm{\Sigma}_{\bm{w}}, \bm{\Sigma}_{\overline{\bm{w}}}$ & --- \\
scalar variance of stochastic gradients $\sigma^2$ & --- \\
minibatch of data $\mathcal{B}$ & --- \\
\toprule
\multicolumn{2}{c}{\textbf{Training hyperparameters and macroscopic state variables}} \\
\toprule
weight decay coefficient (WD) $\lambda$ & pressure $p$ \\
squared radius $\frac{r^2}{2} $ & volume $V$ \\
learning rate (LR) $\eta$ & --- \\
effective learning rate (ELR) $\eta_{\eff} = \eta/r^2$ &  --- \\
\midrule
fixed sphere/fixed ELR: $T = \tfrac{\eta_{\eff}\sigma^2}{2} $ & \multirow{2}{*}{temperature $T$} \\
fixed LR: $T = \sqrt{\frac{\eta\lambda\sigma^2}{2(d-1)}}$ & \\
\toprule
\multicolumn{2}{c}{\textbf{Thermodynamic potentials}} \\
\toprule
\multicolumn{2}{c}{fixed $T$, $V$: Helmholtz (free) energy $F = U - TS$} \\
\multicolumn{2}{c}{fixed $T$, $V$: Maxwell relation $\bigl(\frac{\partial S}{\partial V}\bigr)_T = \bigl(\frac{\partial p}{\partial T}\bigr)_V$} \\
\midrule
\multicolumn{2}{c}{fixed $T$, $p$: Gibbs (free) energy $G = U - TS + pV$} \\
\multicolumn{2}{c}{fixed $T$, $p$: Maxwell relation $\bigl(\frac{\partial S}{\partial p}\bigr)_T = \bigl(\frac{\partial V}{\partial T}\bigr)_p$} \\
\toprule
\multicolumn{2}{c}{\textbf{Ideal gas law}} \\
\toprule
dimensionality constant $\tfrac{d-1}{2}$ & gas constant $R$ \\
\midrule
fixed ELR $r^* = \sqrt{\tfrac{\eta_{\eff} \sigma^2 (d-1)}{2\lambda}} $ & \multirow{2}{*}{ideal gas law $V = \frac{RT}{p}$} \\
fixed LR $r^* = \sqrt[4]{\tfrac{\eta \sigma^2 (d-1)}{2\lambda}} $ & \\
\toprule
\multicolumn{2}{c}{\textbf{Adiabatic process}} \\
\toprule
 \multicolumn{2}{c}{heat $Q$} \\ 
 \multicolumn{2}{c}{First Law of Thermodynamics $\dd U = \delta Q - p\dd V$} \\ 
 \multicolumn{2}{c}{isochoric heap capacity $C_V = \bigl(\frac{\delta Q}{\dd T}\bigr)_V = \bigl(\frac{\partial U}{\partial T}\bigr)_V$} \\
\multicolumn{2}{c}{isobaric heap capacity $C_p = \bigl(\frac{\delta Q}{\dd T}\bigr)_p$} \\
\multicolumn{2}{c}{adiabatic constant $\gamma = C_p/C_V$}
\end{tabular}
\end{center}
\end{table}

\section{Related work}
\label{app:related_work}

\subsubsection{Thermodynamic perspective}
Several studies interpret the stochasticity of SGD through a thermodynamic lens by introducing a temperature parameter $T \propto \eta / B$, where $\eta$ is the learning rate and $B$ is the batch size.
Grounded in SDE dynamics~\cite{jastrzkebski2017three,Mandt2017}, this temperature governs the noise magnitude in parameter updates, shaping different training regimes~\cite{sgd_regimes_thermo} (i.e., switching between stochastic and full-batch optimization) and convergence toward minima of varying sharpness~\cite{bayes_sgd_thermo}.
\cite{chaudhari2018stochastic} analyze the stationary Gibbs distribution $\rho_{\bm{w}}(\bm{w}) \propto \exp(-\Phi(\bm{w}) / T)$ and show that the potential $\Phi(\bm{w})$ equals the training loss $L(\bm{w})$ if and only if the stochastic gradient noise is isotropic. The stationary distribution $\rho_{\bm{w}}$ also minimizes the (Helmholtz) free energy $\mathbb{E}_{\rho_{\bm{w}}}[\Phi(\bm{w})] - T S(\rho_{\bm{w}})$. \cite{kunin2023limiting} further derive the explicit form of $\Phi(\bm{w})$ for linear regression trained using SGD with momentum and weight decay.
We follow the same principle and treat temperature as a function of training hyperparameters, in our case, the learning rate $\eta$ and weight decay $\lambda$. 
The thermodynamic interpretation of SGD offers valuable insights into practical training dynamics, including those observed in large language models~\cite{liu2025neural}, thereby helping to justify the use of specific learning rate schedulers. 

An alternative approach estimates temperature empirically as the mean kinetic energy of individual parameters. 
\cite{fioresi2021thermo} define a separate temperature for each network layer and show that its dependence on hyperparameters is more complex than the simple relation $T \propto \eta / B$.
This idea supports pruning techniques that remove the ``hottest'' parameter groups: filters in convolutional networks~\cite{lapenna2023thermo} or input features in graph neural networks~\cite{lapenna2025temperature}.
Another approach is to derive temperature directly from free energy minimization.  
\cite{sadrtdinov2025sgdfreeenergy} treat the training loss as the internal energy and define temperature as a non-linear, monotonically increasing function of the learning rate, which empirically minimizes the free energy functional.  

Beyond SGD dynamics, thermodynamic perspectives have also been applied to generalization.
Some works define temperature via the parameter-to-data ratio~\cite{zhang2018energy}, while others treat it as an explicit hyperparameter encouraging convergence to flatter, more generalizable minima~\cite{chaudhari2017entropysgd}.
Similar analogies appear in representation learning~\cite{alemi2019therml,gao2020,ziyin2025neural}, particularly through the Information Bottleneck principle~\cite{tishby2015learning}, which formalizes the trade-off between compression and predictive power of learned features.

\subsubsection{Scale-invariant neural networks}
Normalization layers, such as BatchNorm~\cite{ioffe2015batch} and LayerNorm~\cite{ba2016layer}, are indispensable in modern neural architectures.
They smooth the loss landscape~\cite{santurkar2018how} and make training faster~\cite{kohler2019exponential} and more stable~\cite{bjorck2018understanding}.
Beyond these benefits, normalization layers induce scale invariance in network parameters, fundamentally altering their optimization dynamics.
\cite{arora2019theoretical} show that BatchNorm implicitly tunes the learning rate, while \cite{hoffer2018norm} demonstrate that the weight direction evolves according to an effective learning rate $\eta / \|\bm{w}\|^2$.
\cite{van2017l2} argue that, in scale-invariant networks, weight decay does not serve as a regularizer but instead controls the learning rate through the parameter norm, a phenomenon also confirmed by \cite{zhang2019three} and \cite{li2020exponential}. 

Another line of research examines the equilibrium behavior of scale-invariant networks. \cite{wan2021spherical} establish conditions under which the weight norm converges.
Using the SDE framework, \cite{li2020reconciling,wang22three,wan2023how} describe the dynamics of such networks and show that the stationary distribution depends on the intrinsic learning rate $\eta \lambda$. \cite{lobacheva2021periodic} reveal that scale-invariant networks can exhibit periodic instabilities, while \cite{kodryan2022training} study the networks constrained to a fixed sphere and report different regimes: convergence, chaotic equilibrium, or divergence, which depend on the effective learning rate.
Finally, \cite{kosson2024rotational} extend these findings to modern optimizers such as AdamW~\cite{loshchilov2018decoupled} and Lion~\cite{chen2023symbolic}.  

\subsubsection{Statistical physics of learning}
The application of statistical physics to deep learning leverages tools from disordered systems, high-dimensional landscapes, and dynamics to interpret neural network training dynamics and generalization.
\cite{bahri2020statistical} review how concepts like energy landscapes, phase transitions, and random matrix spectra illuminate core phenomena such as loss landscape geometry and layerwise representation learning.
This work links deep learning to spin glass and thermodynamic theory, underpinning a growing effort to explain why overparameterized models are trainable and generalizable despite non-convexity.
Building on these conceptual foundations, a suite of recent studies has developed quantitative statistical physics frameworks capable of predicting and analyzing behavior of deep learning models.
These include generalization capabilities~\cite{canatar2021spectral,li2021statistical,pacelli2023statistical}, networks expressivity~\cite{poole2016exponential,li2018exploring}, feature learning~\cite{wakhloo2023linear,seroussi2023separation}, phase transitions~\cite{baldassi2022learning,veiga2022phase}, and loss landscape flatness~\cite{feng2021inverse}. 

\subsubsection{Our framework}
In this work, we extend the previously established analogies between SGD dynamics and thermodynamics.
Whereas prior studies primarily focused on quantities such as energy, entropy, and temperature, we demonstrate that the optimization of scale-invariant networks naturally gives rise to a richer thermodynamic framework, one that also admits well-defined notions of pressure and volume, and consequently, a direct analogy to the ideal gas law.

\section{Introduction to thermodynamics}
\label{app:intro_thermodyhamics}

\subsubsection{Thermodynamic state variables}
Thermodynamics focuses on systems made up of vast numbers of microscopic particles that move and interact with one another.
Although the motion of individual particles is highly chaotic, their collective behavior exhibits statistical regularities, which permit a description in terms of macroscopic \emph{state variables}.
One key quantity is the \emph{internal energy} $U$, the total kinetic and potential energy of all the particles in the system.
Other fundamental variables include the \emph{entropy} $S$, which measures the system’s disorder, as well as \emph{temperature} $T$, \emph{pressure} $p$, and \emph{volume} $V$. These variables naturally form two pairs of conjugates: $(S, T)$ and $(p, V)$.
To fully specify the state of a thermodynamic system, one typically fixes one variable from each pair; these are referred to as the system’s \emph{natural variables}.

\subsubsection{First and Second Law of Thermodynamics}
The \emph{First Law of Thermodynamics} establishes the principle of energy conservation in thermodynamic processes.
Its differential form is $\dd U = \delta Q - p \dd V$, where $\delta Q$~\footnote{$Q$ is not a state variable, like $U$ or $S$, thus we write $\delta Q$ instead of $\dd Q$.} denotes the infinitesimal heat supplied to the system.
The \emph{Second Law of Thermodynamics} introduces the concept of irreversibility and establishes the direction of spontaneous processes.
For any isolated system (i.e., $\delta Q=0$), the entropy satisfies $\dd S \ge 0$, with equality holding only at equilibrium.
This expresses the fact that thermodynamic systems spontaneously evolve toward \emph{equilibrium states}, where entropy reaches a maximum under the given constraints.
For non-isolated systems exchanging heat $\delta Q$ with the environment, the second law generalizes to $\dd S \ge \delta Q/T$, which accounts for entropy changes due to heat transfer and ensures that the total entropy, including that of the environment, does not decrease.
Here, equality $\dd S = \delta Q/T$ holds only for \emph{reversible processes}, while for irreversible processes the inequality is strict.
For a reversible process, the First Law can also be written as $\dd U = T\dd S - p\dd V$.

\subsubsection{Thermodynamic potentials}
The equation $\dd U = T\dd S - p\dd V$ enables the definition of \emph{thermodynamic potentials}, which describe equilibrium under different sets of natural variables.
At constant temperature $T$ and volume $V$, the \emph{Helmholtz (free) energy} $F = U - TS$ is minimized at equilibrium.
At constant temperature $T$ and pressure $p$, the \emph{Gibbs (free) energy} $G = U - TS + pV$ reaches its minimum.
This minimization principle directly follows from the Second Law, as $\dd S \ge 0$ implies that free energies decrease in spontaneous processes until equilibrium is attained.

The differential forms of thermodynamic potentials yield the \emph{Maxwell relations}, which link different thermodynamic derivatives.
The differentials $\dd F$ and $\dd G$ can be expressed as (see Appendix~\ref{app:proofs:td_laws} for proofs)
\begin{equation}
    \dd F = -S\dd T - p\dd V \qquad\qquad\qquad \dd G = -S\dd T + V\dd p
\end{equation}
From $\dd F = -S\dd T - p\dd V$, it follows that $\frac{\partial F}{\partial T} = -S, \frac{\partial F}{\partial V} =-p$, and one obtains
\begin{equation}
\left(\frac{\partial S}{\partial V}\right)_{T} = -\frac{\partial}{\partial V}\left(\frac{\partial F}{\partial T}\right) = -\frac{\partial}{\partial T}\left(\frac{\partial F}{\partial V}\right) = \left(\frac{\partial p}{\partial T}\right)_{V},
\end{equation} 
while from $\dd G = -S\dd T + V\dd p$, we have $\frac{\partial G}{\partial T} = -S, \frac{\partial G}{\partial p} = V$, and thus
\begin{equation}
-\left(\frac{\partial S}{\partial p}\right)_{T} = \frac{\partial}{\partial p}\left(\frac{\partial G}{\partial T}\right) = \frac{\partial}{\partial T}\left(\frac{\partial G}{\partial p}\right) = \left(\frac{\partial V}{\partial T}\right)_{p}
\end{equation}
Maxwell relations allow experimental determination of entropy changes, which are otherwise difficult to measure directly.

\subsubsection{Gibbs distribution}
On the microscopic scale, the \emph{Gibbs distribution} connects thermodynamics with statistical mechanics.
It assigns the probability $p_i$ for the system to occupy a microstate $i$ with energy $E_i$:
\begin{equation}
p_i = \frac{e^{-E_i/T}}{Z}, \qquad Z = \sum_i e^{-E_i/T}
\end{equation}
Temperature appears in the denominator of the exponential, controlling how energy is distributed among accessible states: lower temperatures concentrate probability on low-energy states, while higher temperatures produce a more uniform distribution.

\subsubsection{Ideal gas and adiabatic process}
An \emph{ideal gas} is a theoretical model of a gas in which intermolecular interactions are neglected and the internal energy depends only on temperature, with the dependence being linear, $U \propto T$.
Its macroscopic behavior is described by the \emph{ideal gas law}: $pV=RT$, where $R$ is the gas constant.
The \emph{heat capacity} quantifies the amount of heat required to change the system’s temperature.
Commonly used are the isochoric heat capacity $C_V = \big(\frac{\delta Q}{\dd T}\big)_V = \big(\frac{\partial U}{\partial T}\big)_V$ and the isobaric heat capacity $C_p = \big(\frac{\delta Q}{\dd T}\big)_p = \big(\frac{\partial U}{\partial T}\big)_p + p \big(\frac{\partial V}{\partial T}\big)_p$, which for an ideal gas are constants and satisfy $C_p - C_V=R$.
An important class of transformations is the \emph{adiabatic process}, in which no heat is exchanged with the surroundings ($\delta Q = 0$). For an adiabatic process, the First Law reduces to $dU = -p\dd V$, and the pressure and volume are related by $p V^\gamma=\text{const}$ with $\gamma=\frac{C_p}{C_V}$.
Moreover, if the process is reversible, we have $\dd S = \delta Q/T = 0$, meaning that we have the same entropy for different configurations of $p$ and $V$.

\section{Missing proofs}
\label{app:proofs}

\subsection{Derivation of SDE}
\label{app:proofs:sde}

\subsubsection{SGD on a fixed sphere / with fixed ELR}
We begin by considering two training protocols that both employ a fixed ELR.
First, we derive the SDE corresponding to the fixed ELR setting.
Then, we reuse the resulting equation for $\overline{\bm{W}}_t$ to describe training constrained to a fixed sphere, since the dynamics of the direction vector are independent of the projection onto the sphere.
The SGD updates in the fixed ELR case are given by
\begin{equation}
    \bm{w}_{k+1} = \bm{w}_k - \eta_{\eff} \|\bm{w}_k\|^2 \Bigl(\nabla L_{\mathcal{B}_k} (\bm{w}_k) + \lambda \bm{w}_k \Bigr) = \bm{w}_k - \eta_{\eff} \|\bm{w}_k\|^2 \Bigl(\nabla L (\bm{w}_k) + \lambda \bm{w}_k \Bigr) - \eta_{\eff} \|\bm{w}_k\|^2 \big(\bm{\Sigma} _{\bm{w}}\big)^{1/2} \bm{\varepsilon},
\end{equation}
where $\bm{\varepsilon} \sim \mathcal{N}(0, \bm{I}_d)$.
Similarly to~\cite{li2020reconciling}, this discrete dynamics leads to the following SDE:
\begin{equation}
    \dd\bm{W}_t = -\eta_{\eff} \|\bm{W}_t\|^2 \Big(\nabla L(\bm{W}_t) + \lambda \bm{W}_t\Big)\dd t  + \eta_{\eff} \|\bm{W}_t\|^2 \big(\bm{\Sigma}_{\bm{W}_t}\big)^{\frac{1}{2}}\dd\bm{B}_t
\end{equation}

To analyze the dynamics of the radius and the direction vector, we utilize the Ito's formula, which produces a new SDE describing the evolution of a twice-differentiable function $f(t, \bm{x})$ when substituting $\bm{x} = \bm{W}_t$.
Given an original SDE
\begin{equation}
    \dd \bm{W}_t = \bm{\mu}(t, \bm{W}_t) \dd t + \bm{G}(t, \bm{W}_t) \dd \bm{B}_t
\end{equation}
with $\bm{\mu}: \mathbb{R}\times \mathbb{R}^d \to \mathbb{R}^d$ and $\bm{G}: \mathbb{R}\times \mathbb{R}^d \to \mathbb{R}^{d\times d}$, we can write a new SDE for $f(t, \bm{W}_t)$ (we omit the dependence on $t$ and $\bm{W}_t$ for clarity)
\begin{equation}
\label{eq:44}
    \dd f_t = \frac{\partial f}{\partial t} \, \dd t + \sum_{i=1}^d \frac{\partial f}{\partial x_i} \, \mu_i \, \dd t + \sum_{i,j=1}^d \frac{\partial f}{\partial x_i} G_{ij} \, \dd \big(\bm{B}_t\big)_j + \frac{1}{2} \sum_{i,j=1}^d \frac{\partial^2 f}{\partial x_i \partial x_j} \big(\bm{G} \bm{G}^T\big)_{ij} \, \dd t,
\end{equation}
The last term is called the \emph{Ito's correction term} and expresses extra drift from the function’s curvature interacting with stochastic noise.
An equivalent of Eq.~\ref{eq:44} in the matrix notation is
\begin{equation}
    \dd f_t = \left( \frac{\partial f}{\partial t} + (\nabla_{\bm{x}} f)^T \bm{\mu} + \frac{1}{2} \Tr \Big[\bm{G} \bm{G}^T \, \nabla^2_{\bm{x}} f \Big] \right) \dd t + (\nabla_{\bm{x}} f)^T \bm{G} \, \dd \bm{B}_t
\end{equation}

We apply the Ito's formula for $r(t, \bm{x}) = \|\bm{x}\|$ and $\overline{\bm{x}}(t, \bm{x}) = \bm{x} / \|\bm{x}\|$. 
Strictly speaking, these functions are not smooth at the origin; therefore, the origin must be excluded from consideration.
In other words, the resulting SDEs are valid only when $\|\bm{W}_t\| \ge \epsilon$ for some small $\epsilon > 0$.
The derivatives of $r$ are
\begin{equation}
    \frac{\partial r}{\partial t} = 0, \quad \nabla_{\bm{x}} r = \frac{\bm{x}}{\|\bm{x}\|}, \quad \nabla^2_{\bm{x}} r = \frac{\dd}{\dd \bm{x}} \biggl(\frac{\bm{x}}{\|\bm{x}\|}\biggr) = \frac{1}{\|\bm{x}\|} \bm{I}_d - \frac{\bm{x}\bm{x}^T}{\|\bm{x}\|^3} = \frac{1}{\|\bm{x}\|} \biggl( \bm{I}_d - \frac{\bm{x}\bm{x}^T}{\|\bm{x}\|^2} \biggr) = \frac{1}{\|\bm{x}\|} \bm{P}_{\bm{x}}
\end{equation}
The Ito's correction term is
\begin{gather}
    \frac12 \sum_{ij} \frac{\partial^2 r}{\partial x_i \partial x_j} (\eta_{\eff}^2 \|\bm{x}\|^4 \bm{\Sigma}_{\bm{x}})_{ij} = \frac{\eta_{\eff}^2\|\bm{x}\|^3}{2} \sum_{ij} (\bm{P}_{\bm{x}})_{ij} (\bm{\Sigma}_{\bm{x}})_{ij} = \frac{\eta_{\eff}^2\|\bm{x}\|^3}{2} \Tr \Bigl(\bm{P}_{\bm{x}}^T \bm{\Sigma}_{\bm{x}} \Bigr) = \frac{\eta_{\eff}^2\|\bm{x}\|}{2} \Tr \Bigl(\bm{P}_{\bm{x}}^T \bm{\Sigma}_{\overline{\bm{x}}} \Bigr) = \nonumber \\
    = \frac{\eta_{\eff}^2\|\bm{x}\|}{2} \Tr \Bigl((\bm{I}_d - \overline{\bm{x}} \, \overline{\bm{x}}^T) \bm{\Sigma}_{\overline{\bm{x}}} \Bigr) = \frac{\eta_{\eff}^2\|\bm{x}\|}{2} \Tr \bm{\Sigma}_{\overline{\bm{x}}} - \frac{\eta_{\eff}^2\|\bm{x}\|}{2} \Tr \Bigl(\overline{\bm{x}}^T \bm{\Sigma}_{\overline{\bm{x}}} \overline{\bm{x}}\Bigr) = \frac{\eta_{\eff}^2\|\bm{x}\|}{2} \Tr \bm{\Sigma}_{\overline{\bm{x}}}
\end{gather}
The second trace is nullified due to Eq.~\ref{eqt:3}. Therefore, the SDE for radius is
\begin{gather}
    \dd r_t = \biggl(\cancelto{0}{\frac{\partial r_t}{\partial t}} - \frac{\eta_{\eff} \|\bm{W}_t\|^2}{\|\bm{W}_t\|} \cancelto{0}{\bm{W}_t^T \nabla L (\bm{W}_t)} - \frac{\eta_{\eff}\|\bm{W}_t\|^2}{\|\bm{W}_t\|} \lambda \bm{W}_t^T \bm{W}_t + \frac{\eta_{\eff}^2\|\bm{W}_t\|}{2} \Tr \bm{\Sigma}_{\overline{\bm{W}}_t}\biggr) \dd t \; + \nonumber \\
    + \frac{\eta_{\eff} \|\bm{W}_t\|^2}{\|\bm{W}_t\|} \Big(\; \cancelto{0}{\big(\bm{\Sigma}_{\bm{W}_t}\big)^{\frac{1}{2}} \bm{W}_t} \quad\; \Big)^T \dd \bm{B}_t = \Bigl(-\eta_{\eff}\lambda r^3_t + \frac{\eta_{\eff}^2r_t}{2} \Tr \bm{\Sigma}_{\overline{\bm{W}}_t}\Bigr)\dd t
\end{gather}
Now, the derivatives of $\overline{\bm{x}}$ (here $\delta$ denotes the Kronecker delta)
\begin{gather}
    \frac{\partial \overline{\bm{x}}}{\partial t} = 0, \qquad \frac{\partial \overline{x}_k}{\partial x_i} = \frac{1}{\|\bm{x}\|} (\bm{P}_{\bm{x}})_{ik} = \frac{\delta_{ik}}{\|\bm{x}\|} - \frac{x_i x_k}{\|\bm{x}\|^3}, \qquad
    \frac{\partial}{\partial x_j} \Bigl(\frac{\delta_{ik}}{\|\bm{x}\|}\Bigr) = -\frac{\delta_{ik} x_j}{\|\bm{x}\|^3} \\
    \frac{\partial}{\partial x_j} \Bigl(\frac{x_i x_k}{\|\bm{x}\|^3}\Bigr) = \frac{\|\bm{x}\|^3}{\|\bm{x}\|^6} \Bigl(\frac{\partial x_i}{\partial x_j} x_k + \frac{\partial x_k}{\partial x_j} x_i\Bigr) - \frac{x_i x_k \cdot \frac{3}{2} \|\bm{x}\| \cdot 2x_j}{\|\bm{x}\|^6} = \frac{\delta_{ij} x_k + \delta_{kj} x_i}{\|\bm{x}\|^3} - \frac{3x_ix_jx_k}{\|\bm{x}\|^5} \\
    \frac{\partial^2 \overline{x}_k}{\partial x_i \partial x_j} = \frac{3x_ix_jx_k}{\|\bm{x}\|^5} - \frac{\delta_{ij} x_k + \delta_{ki} x_j \delta_{kj} x_i}{\|\bm{x}\|^3}
\end{gather}
The Ito's correction term for $\overline{x}_k$ is
\begin{equation}
    \frac12 \sum_{ij} \frac{\partial^2 \overline{x}_k}{\partial x_i \partial x_j} (\eta_{\eff}^2 \|\bm{x}\|^4 \bm{\Sigma}_{\bm{x}})_{ij}
\end{equation}
Expanding each part separately
\begin{align}
    \sum_{ij} \delta_{ij} x_k (\bm{\Sigma}_{\bm{x}})_{ij} &= x_k \sum_{ij} \delta_{ij} (\bm{\Sigma}_{\bm{x}})_{ij} = x_k \Tr \bm{\Sigma}_{\bm{x}} \\
    \sum_{ij} \delta_{ik} x_j (\bm{\Sigma}_{\bm{x}})_{ij} &= \sum_{j} x_j (\bm{\Sigma}_{\bm{x}})_{kj} = (\Sigma_{\bm{x}} \bm{x})_k = 0 \\
    \sum_{ij} \delta_{kj} x_i (\bm{\Sigma}_{\bm{x}})_{ij} &= \sum_{i} x_i (\bm{\Sigma}_{\bm{x}})_{ik} = (\Sigma_{\bm{x}} \bm{x})_k = 0 \\
    3\sum_{ij} x_ix_jx_k (\bm{\Sigma}_{\bm{x}})_{ij} &= 3x_k\sum_{ij} x_ix_j (\bm{\Sigma}_{\bm{x}})_{ij} = 3x_k (\bm{x}^T \bm{\Sigma}_{\bm{x}} \bm{x}) = 0
\end{align}
For the overall direction vector $\overline{\bm{x}}$, the Ito's correction is
\begin{equation}
    -\frac{\eta_{\eff}^2 \|\bm{x}\|}{2} \bm{x} \Tr \bm{\Sigma}_{\bm{x}} = -\frac{\eta_{\eff}^2 \|\bm{x}\|^2}{2} \overline{\bm{x}} \Tr \bm{\Sigma}_{\bm{x}} = -\frac{\eta_{\eff}^2}{2} \overline{\bm{x}} \Tr \bm{\Sigma}_{\overline{\bm{x}}}
\end{equation}
The resulting SDE for the direction vector is
\begin{gather}
    \dd \overline{\bm{W}}_t = \Bigl(\cancelto{0}{\frac{\partial \overline{x}}{\partial t}} - \frac{\eta_{\eff}\|\bm{W}_t\|^2}{\|\bm{W}_t\|} \bm{P}_{\bm{W}_t} \nabla L(\bm{W}_t) - \frac{\eta_{\eff}\|\bm{W}_t\|^2}{\|\bm{W}_t\|} \lambda \cancelto{0}{\bm{P}_{\bm{W}_t} \bm{W}_t} \quad - \frac{\eta_{\eff}^2}{2} \Tr \big(\bm{\Sigma}_{\overline{\overline{\bm{W}}_t}}\big) \overline{\bm{W}}_t \Bigr) \dd t \; + \nonumber \\
    \label{eq:54}
    \frac{\eta_{\eff} \|\bm{W}_t\|^2}{\|\bm{W}_t\|} \bm{P}_{\bm{W}_t} \big(\bm{\Sigma}_{\bm{W}_t}\big)^{\frac{1}{2}} \dd \bm{B}_t = -\Bigl(\eta_{\eff}\nabla L(\overline{\bm{W}_t}) + \frac{\eta_{\eff}^2}{2} \Tr \big(\bm{\Sigma}_{\overline{\bm{W}}_t}\big) \overline{\bm{W}}_t\Bigr)\dd t + \eta_{\eff} \big(\bm{\Sigma}_{\bm{\overline{W}}_t}\big)^{\frac{1}{2}}\dd\bm{B}_t
\end{gather}
One may notice that the Ito's correction term gives a component, which is counter to the direction vector (i.e., it is of the form $-\alpha \overline{\bm{W}}_t$, where $\alpha$ is a positive scalar). 
This component is required to preserve the norm $\|\overline{\bm{W}}_t\|^2 = 1$ in the Ito formulation.
Indeed, if we write down the squared norm after the update:
\begin{gather}
    \|\overline{\bm{W}}_t + \dd \overline{\bm{W}}_t\|^2 = \|\overline{\bm{W}}_t\|^2 + 2\langle \overline{\bm{W}}_t, \dd \overline{\bm{W}}_t\rangle + \|\dd \overline{\bm{W}}_t\|^2 = \\
    = \|\overline{\bm{W}}_t\|^2 \underbrace{- \cancelto{0}{2\eta_{\eff} \overline{\bm{W}}_t^T \nabla L(\overline{\bm{W}}_t)\dd t} - \eta_{\eff}^2 \Tr \big(\bm{\Sigma}_{\overline{\bm{W}}_t}\big) \|\overline{\bm{W}}_t\|^2 \dd t + \cancelto{0}{2\eta_{\eff} \overline{\bm{W}}_t^T \big(\bm{\Sigma}_{\bm{\overline{W}}_t}\big)^{\frac{1}{2}}\dd\bm{B}_t}}_{2\langle \overline{\bm{W}}_t, \dd \overline{\bm{W}}_t\rangle} + \underbrace{\eta_{\eff}^2 \|\big(\bm{\Sigma}_{\bm{\overline{W}}_t}\big)^{\frac{1}{2}} \dd \bm{B}_t\|^2}_{\|\dd \overline{\bm{W}}_t\|^2} = \\
    = 1 - \eta_{\eff}^2 \Tr \big(\bm{\Sigma}_{\overline{\bm{W}}_t}\big) \dd t + \eta_{\eff}^2 \Tr \big(\dd \bm{B}_t^T \bm{\Sigma}_{\bm{\overline{W}}_t} \dd \bm{B}_t\big) = 1 - \eta_{\eff}^2 \Tr \big(\bm{\Sigma}_{\overline{\bm{W}}_t}\big) \dd t + \eta_{\eff}^2 \Tr \Bigl(\bm{\Sigma}_{\bm{\overline{W}}_t} \dd \bm{B}_t \dd \bm{B}_t^T \Bigr) = \\
    = 1 - \eta_{\eff}^2 \Tr \big(\bm{\Sigma}_{\overline{\bm{W}}_t}\big) \dd t + \eta_{\eff}^2 \Tr \Bigl(\bm{\Sigma}_{\bm{\overline{W}}_t} \bm{I}_d \dd t \Bigr) = 1
\end{gather}
Here we substitute $(\dd t)^2 = 0$, $\dd t \cdot \dd (\bm{B}_t)_i = 0$, $(\dd \bm{B}_t)_i^2 = \dd t$, and $(\dd \bm{B}_t)_i\cdot(\dd \bm{B}_t)_j=0$ for $i\ne j$.
Eq.~\ref{eq:54} also describes the dynamics of the direction vector for the fixed sphere case.

Now, we analyze the solutions to these equations. If we have $\Tr \Sigma_{\overline{\bm{w}}}=\text{const}$ for all $\overline{\bm{w}}$, then the stationary radius $r^*$ is given by
\begin{equation}
    r^* = \sqrt{\frac{\eta_{\eff}}{2\lambda} \Tr \Sigma_{\overline{\bm{w}}}} 
\end{equation}
In the \textbf{isotropic noise model}, the covariance matrix trace reduces to
\begin{equation}
    \Tr \Sigma_{\overline{\bm{w}}} = \Tr \Bigl(\bm{P}_{\overline{\bm{w}}} (\sigma^2 \bm{I}_d) \bm{P}_{\overline{\bm{w}}}\Bigr) = \sigma^2 \Tr \Bigl(\bm{P}_{\overline{\bm{w}}} \bm{P}_{\overline{\bm{w}}}\Bigr) = \sigma^2 \Tr \Bigl(\bm{P}_{\overline{\bm{w}}} \Bigr) = \sigma^2 \Tr \Bigl(\bm{I}_d - \overline{\bm{w}}\, \overline{\bm{w}}^T \Bigr) = \sigma^2 (d-1)
\end{equation}
Thus we have
\begin{equation}
    r^* = \sqrt{\frac{\eta_{\eff}\sigma^2(d-1)}{2\lambda}} 
\end{equation}
Eq.~\ref{eq:54} in the isotropic setting becomes
\begin{equation}
    \dd \overline{\bm{W}}_t = -\Bigl(\eta_{\eff}\nabla L(\overline{\bm{W}_t}) + \frac{\eta_{\eff}^2\sigma^2(d-1)}{2} \overline{\bm{W}}_t\Bigr)\dd t + \eta_{\eff} \sigma \bm{P}_{\bm{\overline{W}}_t}\dd\bm{B}_t
\end{equation}
According to~\cite{wang22three}, the stationary distribution for this SDE is given by
\begin{equation}
    \rho_{\overline{\bm{w}}} (\overline{\bm{w}}) \propto \exp \Bigl(-\frac{L(\overline{\bm{w}})}{\tau_{\eff}}\Bigr), \quad \text{ where } \tau_{\eff} = \frac{\eta_{\eff}\sigma^2}{2}
\end{equation}

\subsubsection{SGD with fixed LR}
The SGD iterates in the fixed LR case are
\begin{equation}
    \bm{w}_{k+1} = \bm{w}_k - \eta \Bigl(\nabla L_{\mathcal{B}_k} (\bm{w}_k) + \lambda \bm{w}_k \Bigr) = \bm{w}_k - \eta \Bigl(\nabla L (\bm{w}_k) + \lambda \bm{w}_k \Bigr) - \eta \big(\bm{\Sigma} _{\bm{w}}\big)^{1/2} \bm{\varepsilon},
\end{equation}
where $\bm{\varepsilon} \sim \mathcal{N}(0, \bm{I}_d)$.
This discrete dynamics leads to the following SDE
\begin{equation}
    \dd\bm{W}_t = -\eta \Big(\nabla L(\bm{W}_t) + \lambda \bm{W}_t\Big)\dd t  + \eta \big(\bm{\Sigma}_{\bm{W}_t}\big)^{\frac{1}{2}}\dd\bm{B}_t
\end{equation}
The derivation of the SDE for $r_t$ and $\overline{\bm{W}}_t$ is analogous to the fixed ELR case.
The only difference is that we need to divide the Ito's correction by $r_t^4$ and the rest of terms by $r_t^2$.
This gives the following equations
\begin{align}
    \dd r_t &= \Bigl(-\eta\lambda r_t + \frac{\eta^2}{2r_t^3} \Tr \bm{\Sigma}_{\overline{\bm{W}}_t}\Bigr)\dd t \\
    \label{eq:67}
    \dd \overline{\bm{W}}_t &= -\Bigl(\frac{\eta}{r_t^2}\nabla L(\overline{\bm{W}_t}) + \frac{\eta^2}{2r_t^4} \Tr \big(\bm{\Sigma}_{\overline{\bm{W}}_t}\big) \overline{\bm{W}}_t\Bigr)\dd t + \frac{\eta}{r_t^2} \big(\bm{\Sigma}_{\bm{\overline{W}}_t}\big)^{\frac{1}{2}}\dd\bm{B}_t
\end{align}
The stationary radius for the \textbf{constant covariance trace} and in the \textbf{isotropic noise model}, respectively, is
\begin{equation}
    r^* = \sqrt[4]{\frac{\eta_{\eff}}{2\lambda} \Tr \Sigma_{\overline{\bm{w}}}} = \sqrt[4]{\frac{\eta_{\eff}\sigma^2(d-1)}{2\lambda}} 
\end{equation}
Eq.~\ref{eq:67} in the isotropic case becomes
\begin{equation}
    \dd \overline{\bm{W}}_t = -\Bigl(\frac{\eta}{r_t^2}\nabla L(\overline{\bm{W}_t}) + \frac{\eta^2\sigma^2(d-1)}{2r_t^4} \overline{\bm{W}}_t\Bigr)\dd t + \frac{\eta\sigma}{r_t^2} \bm{P}_{\bm{\overline{W}}_t}\dd\bm{B}_t
\end{equation}
At stationary, we can replace $r_t$ with $r^*$, so the resulting distribution of $\overline{\bm{w}}$ is
\begin{equation}
    \rho_{\overline{\bm{w}}} (\overline{\bm{w}}) \propto \exp \Bigl(-\frac{L(\overline{\bm{w}})}{\tau/(r^*)^2}\Bigr), \quad \text{ where } \tau = \frac{\eta\sigma^2}{2}
\end{equation}

\subsection{Minimization of thermodynamic potentials}
\label{app:proofs:potentials}

In this section, we prove Theorems~\ref{thm:1} and~\ref{thm:2}. 
Theorem~\ref{thm:1} is trivial: it is well-established that the Gibbs distribution $\rho_{\overline{\bm{w}}}(\overline{\bm{w}}) \propto \exp(-L(\overline{\bm{w}})/T)$ minimizes the Helmholtz energy $F=U-TS$ with $U=\mathbb{E}_{\rho_{\overline{\bm{w}}}} [L(\overline{\bm{w}})]$~\cite{le2008introduction}.
Thus, setting temperature equal to the denominator in the stationary distribution, $T=\tau_{\eff} = \frac{\eta_{\eff}\sigma^2}{2}$, guarantees that the $\rho^*_{\overline{\bm{w}}}$ minimizes $F$. 

As for Theorem~\ref{thm:2}, we derive the expressions for $p$ and $V(r)$ than lead to minimization of the Gibbs energy $G=U-TS+pV$.
First, we can decompose $G$ into two terms, depending separately on $\rho_{\overline{\bm{w}}}$ and $r$:
\begin{gather}
    G = U(\rho_{\bm{w}}) - TS(\rho_{\bm{w}}) + pV(r) = U(\rho_{\overline{\bm{w}}}) - T \Big(S(\rho_{\overline{\bm{w}}}) + (d-1) \log r\Big) + pV(r) = \nonumber \\
    = \underbrace{\Big( U(\rho_{\overline{\bm{w}}}) - T S(\rho_{\overline{\bm{w}}})\Big)}_{\text{depends on } \rho_{\overline{\bm{w}}}} + \underbrace{\Big(pV(r) - T (d-1) \log r\Big)}_{\text{depends on } r}
\end{gather}
Similarly to the Helmholtz energy case, setting $T$ equal to the denominator of the Gibbs distribution (which is $\frac{\eta_{\eff}\sigma^2}{2}$ in the fixed-ELR case and $\sqrt{\frac{\eta\lambda\sigma^2}{2(d-1)}}$ in the fixed-LR case) guarantees minimization of the first bracket.
The second bracket depends only on $r$, and we choose $V(r)$ so that it is minimized at the stationary radius $r^*$. Differentiating and substituting $r=r^*$ gives
\begin{equation}
    pV'(r^*) - \frac{T(d-1)}{r^*} = 0 \Rightarrow V'(r^*) = \frac{T(d-1)}{pr^*}
\end{equation}
Now, using $T=\frac{\lambda (r^*)^2}{(d-1)}$, valid for both fixed-ELR and fixed-LR cases (Eq.~\ref{eqt:77}),
\begin{equation}
    \label{eqt:77} T=\tau_{\eff}=\frac{\lambda(r^*)^2}{d-1} \qquad\qquad\qquad T=\frac{\tau}{(r^*)^2} = \frac{\lambda(r^*)^4}{(d-1)(r^*)^2}=\frac{\lambda(r^*)^2}{d-1}
\end{equation}
we obtain $V'(r^*) = \frac{\lambda}{p} r^*$. 
This naturally leads to the choice $p=\lambda$ and $V=r^2/2$ (with $V_0=0$ for zero volume at zero radius).
Notably, this choice works consistently for both fixed-ELR and fixed-LR protocols despite their different temperatures $T$, further supporting that these definitions are natural.
Interestingly, even though $r^{d-1}$ is a more direct \emph{geometric} notion of volume of the sphere $\mathbb{S}^{d-1}(r)$ (i.e., the surface area), the \emph{thermodynamic} volume scales as $r^2$ and does not depend on the dimensionality of the weight space. 

\subsection{First Law of Thermodynamics}
\label{app:proofs:td_laws}

In this section, we show that the First Law of Thermodynamics for reversible processes (i.e., where $\delta Q = TdS$ is substituted) holds for stationary distributions of SGD, $\dd U = T\dd S - p \dd V$.
Recall that
\begin{equation}
    U = \mathbb{E}_{\rho_{\overline{\bm{w}}}} \bigl[L(\overline{\bm{w}})\bigr], \qquad S=\mathbb{E}_{\rho_{\overline{\bm{w}}}} \bigl[-\log \rho_{\overline{\bm{w}}} (\overline{\bm{w}})\bigr] + \frac{d-1}{2}\log (2V), \qquad \rho_{\overline{\bm{w}}} (\overline{\bm{w}}) = \frac{1}{Z(T)} \exp \Bigl(-\frac{L(\overline{\bm{w}})}{T}\Bigr),
\end{equation}
where $Z(T)$ denotes the normalization constant. We can represent $S$ as
\begin{equation}
    S = \mathbb{E}_{\rho_{\overline{\bm{w}}}} \Bigl[\log Z(T) + \frac{L(\overline{\bm{w}})}{T} \Bigr] + \frac{d-1}{2}\log (2V) = \frac{U}{T} + \log Z(T) + \frac{d-1}{2}\log (2V)
\end{equation}
Taking the differential, we obtain
\begin{equation}
    \dd S = \frac{1}{T} \dd U - \frac{U}{T^2} \dd T + \frac{Z'(T)}{Z(T)} \dd T + \frac{d-1}{2V}\dd V
\end{equation}
Now, we show that $\frac{U}{T^2} = \frac{Z'(T)}{Z(T)}$
\begin{gather}
    \frac{Z'(T)}{Z(T)} = \frac{1}{Z(T)}\cdot\frac{\dd}{\dd T} \int_{\mathbb{S}^{d-1}} \exp\Bigl(-\frac{L(\overline{\bm{w}})}{T}\Bigr) d\overline{\bm{w}} = \frac{1}{Z(T)}\cdot\int_{\mathbb{S}^{d-1}} \frac{\dd}{\dd T} \exp\Bigl(-\frac{L(\overline{\bm{w}})}{T}\Bigr) d\overline{\bm{w}} = \nonumber \\
    = \frac{1}{Z(T)}\cdot \int_{\mathbb{S}^{d-1}} \exp\Bigl(-\frac{L(\overline{\bm{w}})}{T}\Bigr) \frac{L(\overline{\bm{w}})}{T^2}d\overline{\bm{w}} = \frac{1}{T^2}\cdot \int_{\mathbb{S}^{d-1}} \frac{L(\overline{\bm{w}})}{Z(T)} \exp\Bigl(-\frac{L(\overline{\bm{w}})}{T}\Bigr) d\overline{\bm{w}} = \frac{U}{T^2}
\end{gather}
Thus, $\dd S = \frac{1}{T} \dd U + \frac{d-1}{2V}\dd V$. Hence
\begin{equation}
    T\dd S - p\dd V = \dd U + \underbrace{\frac{T(d-1)}{2V}}_{p} \dd V - p\dd V= \dd U
\end{equation}
Note that we used the stationary distribution on the unit sphere and the ideal gas law $p = \frac{T(d-1)}{2V}$ (i.e., the stationary radius expression) to derive this relation.

Once the First Law is established, we can also derive the expression for Helmholtz and Gibbs energies.
The key idea is that these potentials should be total differentials in the corresponding natural variables: $T$, $V$ for $F$ and $T$, $p$ for $G$.
This is done via so-called Legendre transformation
\begin{align}
    F &= U - \frac{\partial U}{\partial S} S = U - TS, \qquad \dd F = \dd U - \dd (TS) = T\dd S - p\dd V - T\dd S - S \dd T= -S\dd T - p \dd V \\
    \label{eq:46}
    G &= F - \frac{\partial F}{\partial V} V = F + pV , \qquad \dd G = \dd F + \dd (pV) = -S\dd T - p\dd V + p\dd V + V \dd p = -S \dd T + V\dd p
\end{align}

\subsection{Maxwell relations}
\label{app:proofs:maxwell}

In this section, we derive specific Maxwell relations for three considered training protocols. 

\subsubsection{SGD on a fixed sphere}
For the fixed sphere case, verifying the Maxwell relation is equivalent to checking that the pressure follows the ideal gas law $p=\frac{T(d-1)}{2V}$
\begin{equation}
\label{eq:47}
\Big(\frac{\partial p}{\partial T}\Big)_V = \Big(\frac{\partial S}{\partial V}\Big)_T = \frac{d-1}{2V}
\end{equation}
Although there is no explicit application of weight decay in this training protocol, we can estimate the \emph{effective weight decay} $\lambda_{\eff}$, which is associated with the projection back to the sphere $\mathbb{S}^{d-1}(r)$ after each training step.
In other words, it is the value of the weight decay coefficient required to maintain the dynamics on the same sphere, which we estimate empirically as
\begin{equation}
    \lambda_{\eff} = \mathbb{E}\bigg[\frac{\|\bm{w}_k - \eta \nabla L_{\mathcal{B}_k} (\bm{w}_k)\| - \|\bm{w}_k\|}{\eta \|\bm{w}_k\|}\bigg],
\end{equation}
where $\eta = \eta_{\eff}\|\bm{w}_k\|^2$ and $\|\bm{w}_k\| = r$.
Similarly to the fixed ELR/LR cases, we interpret the effective weight decay $\lambda_{\text{eff}}$ as pressure $p$.
Thus, verifying both the Maxwell relation and the ideal gas law reduces to checking whether 
\begin{equation}
    \lambda_{\text{eff}} = \frac{T (d - 1)}{2V} = \frac{\eta_{\eff} \sigma^2 (d-1)}{2r^2}
\end{equation}  

\subsubsection{SGD with fixed ELR}
Training protocol for SGD with fixed ELR corresponds to fixing $T$ and $p$ with the Maxwell relation $-\big(\frac{\partial S}{\partial p}\big)_T = \big(\frac{\partial V}{\partial T}\big)_p$. 
Considering $V = \frac{T(d-1)}{2p}$, we get $\big(\frac{\partial V}{\partial T}\big)_p = \frac{d-1}{2p}$.
Finally, given $\big(\frac{\partial S}{\partial p}\big)_T = \frac{1}{p} \big(\frac{\partial S}{\partial \log p}\big)_T$ and substituting $p = \lambda$, we get (note that fixing $T$ corresponds to fixing $\eta_{\eff}$)
\begin{equation}
    \Bigl(\frac{\partial S}{\partial \log \lambda}\Bigr)_{\eta_{\eff}} = -\frac{d-1}{2}
\end{equation}

\subsubsection{SGD with fixed LR}
In the fixed LR case, $T$ is dependent on both $\eta$ and $\lambda$, so we introduce this pair as new natural variables (instead of $T$ and $p$).
We also consider $\sigma=\sigma(\eta\lambda)$ to account for variable variance in the experiments, as shown in Subfigure~\ref{fig:resnet_cifar10_lr}a.
Despite $\sigma$ being variable, we obtain the same Eq.~\ref{eq:25}. 
To start with, we express the differential of $T = \sqrt{\frac{\eta\lambda\sigma^2(\eta\lambda)}{2(d-1)}}$ (for brevity, we omit the brackets in $\sigma(\eta\lambda)$ and $\sigma'(\eta\lambda)$)
\begin{gather}
    \Bigl(\frac{\partial T}{\partial \eta}\Bigr)_{\lambda} = \sqrt{\frac{\lambda\sigma^2}{2(d-1)}} \cdot \frac{1}{2\sqrt{\eta}} + \sqrt{\frac{\eta\lambda}{2(d-1)}} \cdot \sigma' \lambda = \sqrt{\frac{\eta\lambda}{2(d-1)}} \Bigl(\frac{\sigma}{2\eta} + \sigma'\lambda\Bigr) \\
    \Bigl(\frac{\partial T}{\partial \lambda}\Bigr)_{\eta} = \sqrt{\frac{\eta\sigma^2}{2(d-1)}} \cdot \frac{1}{2\sqrt{\lambda}} + \sqrt{\frac{\eta\lambda}{2(d-1)}} \cdot \sigma' \eta = \sqrt{\frac{\eta\lambda}{2(d-1)}} \Bigl(\frac{\sigma}{2\lambda} + \sigma'\eta\Bigr) \\
    \dd T = \Bigl(\frac{\partial T}{\partial \eta}\Bigr)_{\lambda} \dd \eta + \Bigl(\frac{\partial T}{\partial \lambda}\Bigr)_{\eta} \dd \lambda
\end{gather}
Thus, the differential of Gibbs energy (Eq.~\ref{eq:46}) is
\begin{equation}
    \dd G = -S\dd T + V\dd p = -S \Bigl(\frac{\partial T}{\partial \eta}\Bigr)_{\lambda} \dd \eta - S\Bigl(\frac{\partial T}{\partial \lambda}\Bigr)_{\eta} \dd \lambda + V\dd \lambda = -S \Bigl(\frac{\partial T}{\partial \eta}\Bigr)_{\lambda} \dd \eta + \biggl[V - S\Bigl(\frac{\partial T}{\partial \lambda}\Bigr)_{\eta} \biggr] \dd \lambda 
\end{equation}
To derive the Maxwell relation, we need to equate the second cross-derivatives
\begin{gather}
    -\frac{\partial}{\partial \lambda}\biggl[S \Bigl(\frac{\partial T}{\partial \eta}\Bigr)_{\lambda}\biggr] = \frac{\partial}{\partial \eta}\biggl[V - S\Bigl(\frac{\partial T}{\partial \lambda}\Bigr)_{\eta} \biggr] \\
    -\Bigl(\frac{\partial S}{\partial \lambda}\Bigr)_{\eta} \Bigl(\frac{\partial T}{\partial \eta}\Bigr)_{\lambda} - \cancel{S\Bigl(\frac{\partial^2 T}{\partial \eta \partial \lambda}\Bigr)} = \Bigl(\frac{\partial V}{\partial \eta}\Bigr)_{\lambda} - \Bigl(\frac{\partial S}{\partial \eta}\Bigr)_{\lambda} \Bigl(\frac{\partial T}{\partial \lambda}\Bigr)_{\eta} - \cancel{S\Bigl(\frac{\partial^2 T}{\partial \eta \partial \lambda}\Bigr)} \\
    -\Bigl(\frac{\partial S}{\partial \lambda}\Bigr)_{\eta} \Bigl(\frac{\partial T}{\partial \eta}\Bigr)_{\lambda} = \Bigl(\frac{\partial V}{\partial T}\Bigr)_{\lambda} \Bigl(\frac{\partial T}{\partial \eta}\Bigr)_{\lambda} - \Bigl(\frac{\partial S}{\partial \eta}\Bigr)_{\lambda} \Bigl(\frac{\partial T}{\partial \lambda}\Bigr)_{\eta} \\
    -\Bigl(\frac{\partial S}{\partial \lambda}\Bigr)_{\eta} \cancel{\sqrt{\frac{\eta\lambda}{2(d-1)}}} \Bigl(\frac{\sigma}{2\eta} + \sigma'\lambda\Bigr) = \frac{d-1}{2\lambda} \cancel{\sqrt{\frac{\eta\lambda}{2(d-1)}}} \Bigl(\frac{\sigma}{2\eta} + \sigma'\lambda\Bigr) - \Bigl(\frac{\partial S}{\partial \eta}\Bigr)_{\lambda} \cancel{\sqrt{\frac{\eta\lambda}{2(d-1)}}} \Bigl(\frac{\sigma}{2\lambda} + \sigma'\eta\Bigr) \\ 
    -\frac{1}{\lambda} \Bigl(\frac{\partial S}{\partial \log \lambda}\Bigr)_{\eta}  \Bigl(\frac{\sigma}{2\eta} + \sigma'\lambda\Bigr) = \frac{d-1}{2\lambda} \Bigl(\frac{\sigma}{2\eta} + \sigma'\lambda\Bigr) - \frac{1}{\eta} \Bigl(\frac{\partial S}{\partial \log \eta}\Bigr)_{\lambda} \Bigl(\frac{\sigma}{2\lambda} + \sigma'\eta\Bigr) \\ 
    - \Bigl(\frac{\partial S}{\partial \log \lambda}\Bigr)_{\eta} \; \cancel{\Bigl(\frac{\sigma}{2\eta\lambda} + \sigma'\Bigr)} = \frac{d-1}{2} \cancel{\Bigl(\frac{\sigma}{2\eta\lambda} + \sigma'\Bigr)} - \Bigl(\frac{\partial S}{\partial \log \eta}\Bigr)_{\lambda} \; \cancel{\Bigl(\frac{\sigma}{2\eta\lambda} + \sigma'\Bigr)} \\ 
    \Bigl(\frac{\partial S}{\partial \log \eta}\Bigr)_{\lambda} - \Bigl(\frac{\partial S}{\partial \log \lambda}\Bigr)_{\eta} = \frac{d-1}{2}
\end{gather}
All terms including $\sigma$ and $\sigma'$ are eliminated, so the formula holds for both constant and variable $\sigma$.

\subsection{Heat capacity and adiabatic constant}
\label{app:proofs:heat_cap}

In this section, we derive the relation for heat capacities $C_p - C_V = R$ in our analogy and show that the adiabatic process is given by $pV^{\gamma} = \text{const}$ with $\gamma = C_p / C_V$. We use the expressions $\delta Q = \dd U + p \dd V$, $V = \frac{T(d-1)}{2p}$, and the fact that $U$ is a function of temperature $T$
\begin{equation}
    C_V = \Bigl(\frac{\delta Q}{\dd T}\Bigr)_V = \Bigl(\frac{\partial U}{\partial T}\Bigr)_V = \frac{\dd U}{\dd T}, \qquad C_p = \Bigl(\frac{\delta Q}{\dd T}\Bigr)_p = \Bigl(\frac{\partial U}{\partial T}\Bigr)_p + p \Bigl(\frac{\partial V}{\partial T}\Bigr)_p = \frac{\dd U}{\dd T} + p \frac{d-1}{2p} = C_V + R
\end{equation}
Now, we set $\delta Q = 0$ and relate $\dd p$ and $\dd V$
\begin{gather}
    0 = \delta Q = \dd U + p\dd V = C_V \, \dd T + p\dd V = C_V \, \dd \Bigl(\frac{PV}{R}\Bigr) + p\dd V = \nonumber \\
    = \frac{C_V}{R} \big(p\dd V + V \dd p\big) + p\dd V = \frac{C_p}{R} p\dd V + \frac{C_V}{R} V\dd p \\
    C_V V \dd p = - C_p p \dd V \Leftrightarrow \frac{dp}{p} = -\gamma \frac{dV}{V} \Leftrightarrow \log p = -\gamma \log V + \text{const} \Leftrightarrow \nonumber \\
    \Leftrightarrow \log (pV^\gamma) = \text{const} \Leftrightarrow pV^\gamma = \text{const}
\end{gather}
 
\subsection{Spherical entropy estimator}
\label{app:proofs:entropy}

In this section, we describe the entropy estimation protocol. 
We use the nearest neighbor entropy estimator~\cite{KozLeo87}.
Given a sample $\{\bm{x}_1, \dots, \bm{x}_N\}$ with $\bm{x}_i \in \mathbb{R}^d$, the estimator $\hat{S}_{\mathbb{R}^d}$ is
\begin{equation}
    \label{eq:28}
    \hat{S}_{\mathbb{R}^d}(\bm{x}) = \frac{d}{N}\sum_{i=1}^{N} \log \zeta_i + C(N, d),
\end{equation}
where $\zeta_i$ is the L2 distance from $\bm{x}_i$ to its nearest neighbor and $C(N, d)$ is the function of sample size and dimensionality,
\begin{equation}
    C(N, d) = \log (N - 1) - \log \Gamma \left(\frac{d}{2} + 1\right)+ \frac{d}{2}\log \pi + \gamma,
\end{equation}
where $\Gamma$ denotes the gamma function, and $\gamma \approx 0.577$ is the Euler constant.

However, this estimator operates in the Euclidian space, while we are interested in the entropy $S_{\mathbb{S}^{d-1}}(\overline{\bm{w}})$ on the unit sphere.
To mitigate this mismatch, we convert $\overline{\bm{w}}$ to spherical coordinates $\bm{\theta} = \{\theta_1, \dots, \theta_{d-2}, \phi\}$, estimate entropy of this angular vector with Eq.~\ref{eq:28}, and correct it by the expected Jacobian of the spherical coordinate transformation.
Under the change of variables, the densities are related as $\rho_{\bm{\theta}} (\bm{\theta}) = \rho_{\overline{\bm{w}}} (\overline{\bm{w}}) J(\bm{\theta})$, where $J (\bm{\theta}) = \prod_{j=1}^{d-2} \sin (\theta_j)^{d-1-j}$.
Hence
\begin{gather}
    S_{\mathbb{S}^{d-1}}(\overline{\bm{w}}) = \mathbb{E}_{\rho_{\overline{\bm{w}}}} \bigl[-\log \rho_{\overline{\bm{w}}} (\overline{\bm{w}})\bigr] = \mathbb{E}_{\rho_{\overline{\bm{w}}}} \bigl[-\log \rho_{\bm{\theta}} (\bm{\theta}) + \log J (\bm{\theta})\bigr] = S_{\mathbb{R}^{d-1}}({\bm{\theta}}) + \mathbb{E}_{\rho_{\overline{\bm{w}}}} \bigl[\log J (\bm{\theta})\bigr], \\
    \text{where } \log J (\bm{\theta}) = \sum_{j=1}^{d-2} (d-1-j) \log \sin (\theta_j)
\end{gather}

The total entropy is then
\begin{equation}
    \hat{S}(\bm{w}) = \hat{S}_{\mathbb{S}^{d-1}}(\overline{\bm{w}}) + (d-1)\log \mathbb{E} \|\bm{w}\|
\end{equation}

\section{Additional results for isotropic noise model}
\label{app:toy}

\subsubsection{Statistics of VMF distribution}
For consistency with the existing sources, we derive the statistics of the VMF distribution for inverse temperature $\kappa=1/T$ and then we rewrite them in terms of the original temperature $T$.
The density is given by\footnote{\href{https://en.wikipedia.org/wiki/Von_Mises–Fisher_distribution}{https://en.wikipedia.org/wiki/Von\_Mises–Fisher\_distribution}}
\begin{equation}
    \rho_{\overline{\bm{w}}}({\overline{\bm{w}}}) = C_d(\kappa) \exp \bigl(-\kappa \bm{\mu}^T\overline{\bm{w}}\bigr), \qquad \text{where } C_d(\kappa) = \frac{\kappa^{d/2-1}}{(2\pi)^{d/2} I_{d/2-1} (\kappa)},
\end{equation}
where $I_{d/2-1} (\kappa)$ is the modified Bessel function of first kind.
The expected loss $U$ and entropy $S$ for this distribution are
\begin{equation}
    U = \mathbb{E}_{\rho_{\overline{\bm{w}}}} \bigl[1 + \bm{\mu}^T \overline{\bm{w}}\bigr] = 1 - \bm{\mu}^T A_d  (\kappa) \bm{\mu} = 1 - A_d (\kappa) \quad S = -\log C_d (\kappa) - \kappa A_d(\kappa), \quad \text{with } A_d(\kappa) = \frac{I_{d/2} (\kappa)}{I_{d/2-1} (\kappa)}
\end{equation}
We are interested in the asymptotics of these functions in the limit $\kappa \to \infty$ (i.e., $T\to 0$, so that we can apply this approximation for sufficiently small values of $\eta$ and $\lambda$).
\cite{NIST:DLMF} give the following approximation for $I_{\nu}(\kappa)$ in \href{https://dlmf.nist.gov/10.40}{10.40.1}
\begin{equation}
    I_{\nu} (\kappa) = \frac{e^{\kappa}}{\sqrt{2\pi\kappa}} \Bigl(1 - \frac{4\nu^2 - 1}{8\kappa} + \overline{o} (\kappa^{-1})\Bigr)
\end{equation}
Let $\nu = d/2$. The asymptotics for $A_d(\kappa)$ is
\begin{gather}
    A_d (\kappa) = \frac{I_{\nu} (\kappa)}{I_{\nu - 1} (\kappa)} = \frac{1 - \frac{4\nu^2 - 1}{8\kappa} + \overline{o}(\kappa^{-1})}{1 - \frac{4 (\nu - 1)^2 - 1}{8\kappa} + \overline{o}(\kappa^{-1})} = 1 - \frac{(4\nu^2 - 1) - (4(\nu - 1)^2 - 1)}{8\kappa} + \overline{o}(\kappa^{-1}) = \nonumber \\
    = 1 - \frac{8\nu - 4}{8\kappa} + \overline{o}(\kappa^{-1}) = 1 - \frac{2\nu - 1}{2\kappa} + \overline{o}(\kappa^{-1}) = 1 - \frac{d - 1}{2\kappa} + \overline{o}(\kappa^{-1})
\end{gather}
Thus, the expected loss $U$ is:
\begin{equation}
    U = 1 - A_d(\kappa) = \frac{d-1}{2\kappa} + \overline{o}(\kappa^{-1}) = \frac{d-1}{2}T + \overline{o}(T)
\end{equation}
The entropy $S$ is
\begin{gather}
    S = -\log C_d (\kappa) - \kappa A_d(\kappa) = -\Bigl(\frac{d}{2} - 1\Bigr) \log \kappa + \frac{d}{2}\log(2\pi) + \log \biggl(\frac{e^{\kappa}}{\sqrt{2\pi\kappa}} \Bigl(1 - \frac{4(d/2-1)^2 - 1}{8\kappa} + \overline{o} (\kappa^{-1})\Bigr)\biggr) - \nonumber \\
    -\kappa \Bigl(1 - \frac{d - 1}{2\kappa} + \overline{o}(\kappa^{-1})\Bigr) = -\Bigl(\frac{d}{2} - 1\Bigr) \log \kappa + \frac{d}{2}\log(2\pi) + \cancel{\kappa} - \frac{1}{2}\log(2\pi) - \frac{1}{2} \log \kappa - \cancel{\kappa} + \frac{d-1}{2} + \overline{o} (1) = \nonumber \\
    = \frac{d-1}{2} \log \Bigl(\frac{2\pi e}{\kappa}\Bigr) + \overline{o} (1) = \frac{d-1}{2} \log \Bigl(2\pi e T\Bigr) + \overline{o} (1)
\end{gather}

\begin{figure*}
    \centering
    \includegraphics[width=0.98\textwidth]{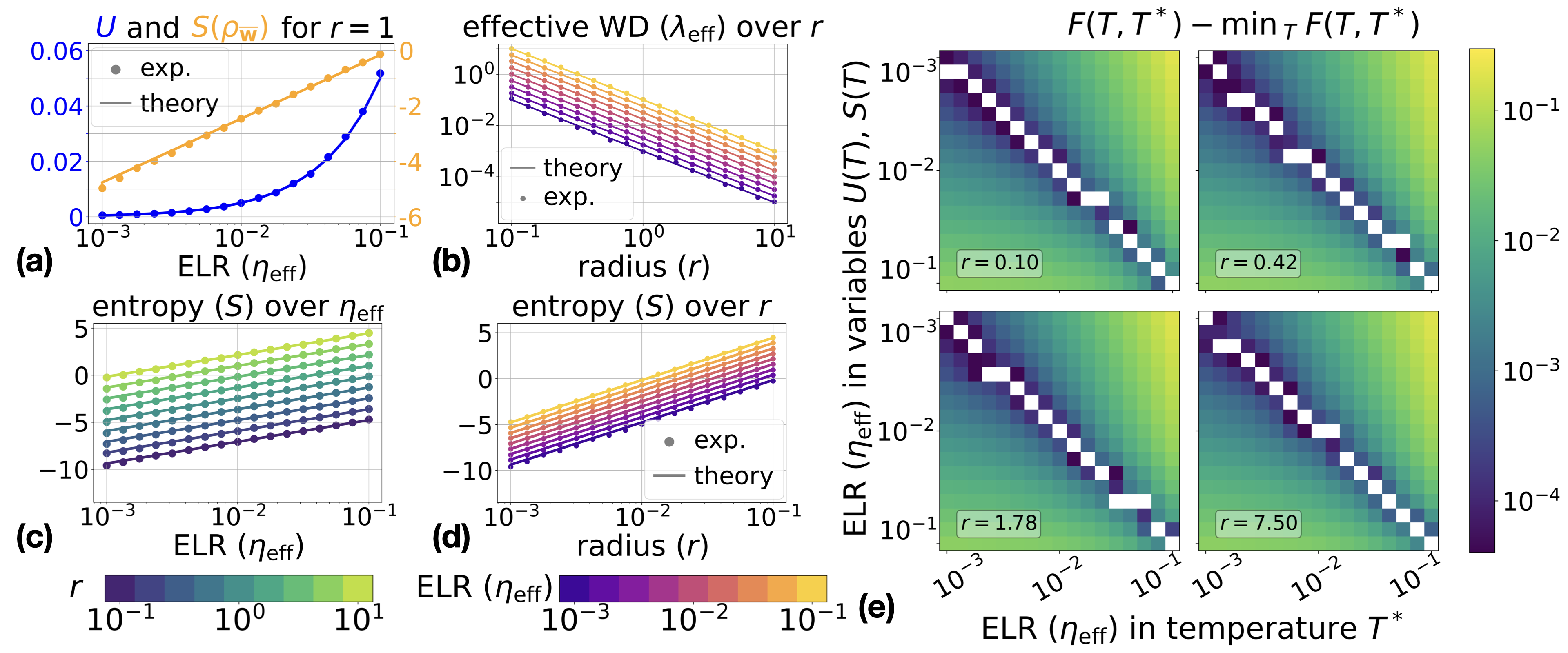}
    \caption{
    Results for the VMF isotropic noise model on a fixed sphere with radius~$r$ and ELR~$\eta_{\eff}$. Subfigures a--d: points are numerical measurements, solid lines are theoretical predictions: $U=\frac{d-1}{2}T$, $S(\rho_{\overline{\bm{w}}})=\frac{d-1}{2}\log(2\pi eT)$, $S = S(\rho_{\overline{\bm{w}}}) + (d-1)\log r$, and $\lambda_{\eff}=\frac{T(d-1)}{2V}$, with $T=\frac{\eta_{\eff}\sigma^2}{2}$ and $V=\frac{r^2}{2}$. Subfigure e: Helmholtz energy minimization (\hyperref[V2]{\bf{V2}}). Each subplot corresponds to a radius value $r$. On the horizontal axis, we vary $\eta_{\eff}$ in temperature $T^*$ of Helmholtz energy $F$; on the vertical axis, we consider stationary distributions induced by different $\eta_{\eff}$. The colormap shows the difference between $F$ and its minimum across different stationary distributions (i.e., across each column), with the minimizer marked by a white square. Ideally, white squares coincide with the diagonal; in practice, they either match or lie very close.}
    \label{fig:isotropic_vmf_sphere}
\end{figure*}

\begin{figure*}
    \centering
    \includegraphics[width=0.98\textwidth]{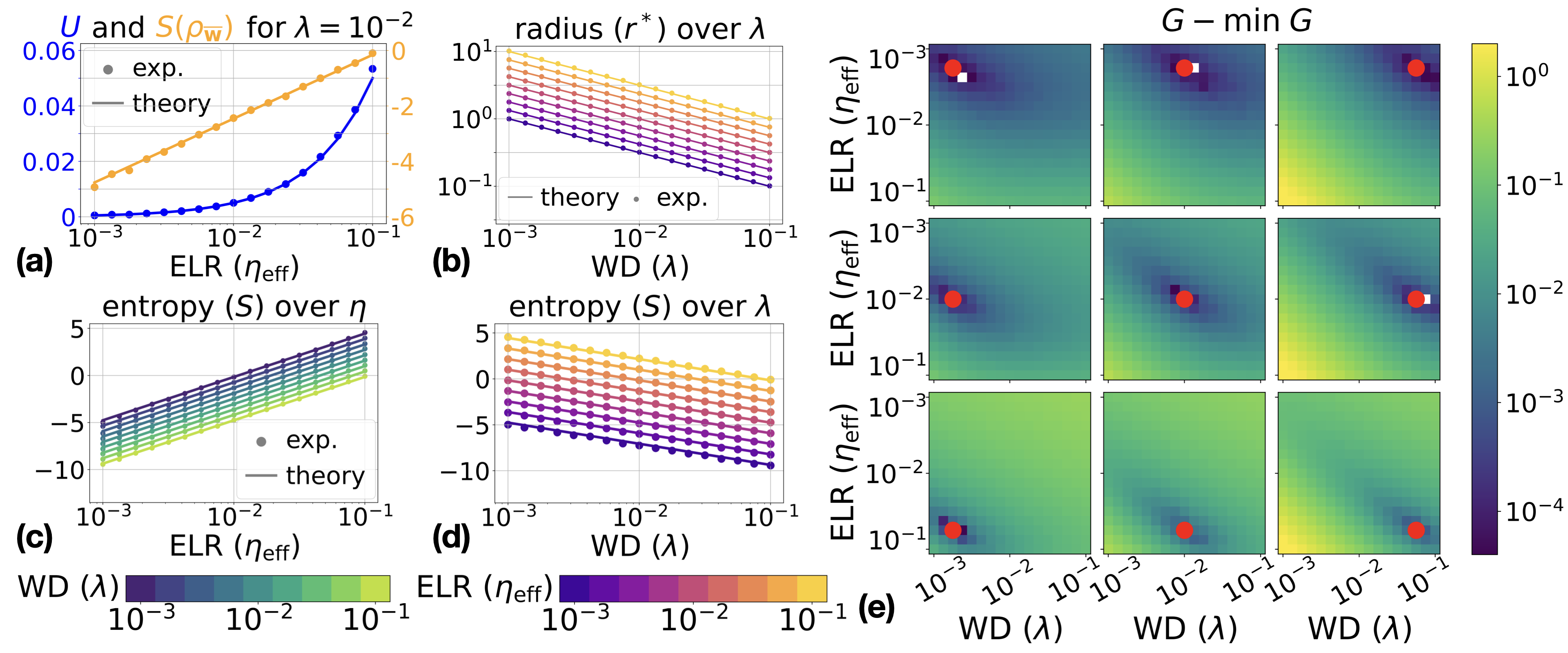}
    \caption{
    Results for the VMF isotropic noise model with fixed ELR~$\eta_{\eff}$ and WD~$\lambda$. Subfigures a--d: points are numerical measurements, solid lines are theoretical predictions: $U=\frac{d-1}{2}T$, $S(\rho_{\overline{\bm{w}}})=\frac{d-1}{2}\log(2\pi eT)$, $S = S(\rho_{\overline{\bm{w}}}) + (d-1)\log r^*$, and $r^*=\sqrt{\frac{T(d-1)}{p}}$, with $T=\frac{\eta_{\eff}\sigma^2}{2}$ and $p=\lambda$. Subfigure e: Gibbs energy minimization (\hyperref[V2]{\bf{V2}}). Each subplot corresponds to a fixed pair $(\eta_{\eff}^*, \lambda^*)$, denoted with red circle. The colormap shows the difference between $G$ and its minimum across stationary distributions, with the minimizer marked by a white square. Ideally, red circles coincide with white squares; in practice, they either match or lie very close.}
    \label{fig:isotropic_vmf_elr}
\end{figure*}

\paragraph{Detailed experimental setup.}
We launch noisy gradient descent given by Eq.~\ref{eq:24} for $2 \cdot 10^6$ iterations. 
The gradient variance is set to $\sigma=1$.
We consider $\eta, \eta_{\eff}, \lambda \in [10^{-3}, 10^{-1}]$.
In the case of the fixed sphere, we sweep over $r \in [10^{-1}, 10^{1}]$. 
We sample $\mu$ from the uniform distribution on $\mathbb{S}^{d-1}$ and keep it the same for all the considered hyperparameter values.
For all of these ranges, we take $17$ values equally spaced in the logarithmic domain.
We log training loss and radius every $50$ iterations.
To calculate the mean values for stationary distributions, we average over last $5000$ logs.
Entropy is logged every $40000$ iterations.
The stationary entropy is the average of the last $10$ logs.
The isotropic noise model is implemented on CPU; running one training protocol for a square grid of hyperparameters requires $<10$ minutes.

\paragraph{Results for VMF model.}
In Figure~\ref{fig:isotropic_vmf_sphere}, we present the results of numerical simulation for training on a fixed sphere.
Similarly to the fixed LR case, empirical $U$ and $S(\rho_{\overline{\bm{w}}})$ closely follow their theoretic predictions in Subfigure~\ref{fig:isotropic_vmf_sphere}a, confirming that the stationary distribution is of the VMF form.
The minimization of Helmholtz energy in Subfigure~\ref{fig:isotropic_vmf_sphere}e proves \hyperref[V2]{\bf{V2}}. 
Subfigure~\ref{fig:isotropic_vmf_sphere}b verifies \hyperref[V3]{\bf{V3}}, which is equivalent to checking the stationary pressure (i.e., the effective weight decay, Eq.~\ref{eq:47}).
Similarly, Figure~\ref{fig:isotropic_vmf_elr} demonstrates the case of training with the fixed ELR.
All empirical quantities meet the corresponding theoretic predictions, and the Gibbs energy $G$ is minimized among the considered stationary distributions.

\section{Additional results for neural networks}
\label{app:networks}

\begin{table}[t]
    \renewcommand{\arraystretch}{1.1}
    \centering
    \begin{tabular}{ccccc}
    \toprule
    \textbf{Training setup} & \makecell{\bf ResNet-18 \\ \bf CIFAR-10} & \makecell{\bf ResNet-18 \\ \bf CIFAR-100} & \makecell{\bf ConvNet \\ \bf CIFAR-10} & \makecell{\bf ConvNet \\ \bf CIFAR-100} \\
    \toprule
    width multipliet $k$ & $4$ & $4$ & $8$ & $16$ \\
    number of trainable parameters $d$ & $43692$ & $43692$ & $24408$ & $97200$ \\
    training iterations $t$ & $10^6$ & $2\cdot 10^6$ & $2\cdot 10^6$ & $2\cdot 10^6$ \\
    entropy queue size $N$ & $1000$ & $1000$ & $4000$ & $4000$ \\
    \midrule 
    ELR grid $\eta_{\eff}$ & \multicolumn{2}{c}{$\{10^{-4 + n/6}|n=0,\dots, 13\}$} & \multicolumn{2}{c}{$\{10^{-4 + n/6}|n=0,\dots, 12\}$} \\
    LR grid $\eta$ & \multicolumn{2}{c}{$\{10^{-3 + n/6}|n=0,\dots, 18\}$} & \multicolumn{2}{c}{$\{10^{-3 + n/6}|n=0,\dots, 12\}$} \\
    WD grid $\lambda$ & \multicolumn{2}{c}{$\{10^{-3 + n/2}|n=0,\dots, 4\}$} & \multicolumn{2}{c}{$\{10^{-3 + n/4} \text{ and } 10^{-2 + n/4} |n=0,\dots, 8\}$} \\
    radius grid $r$ & \multicolumn{4}{c}{$\{2^{-2 + n} | n =0, \dots, 4\}$} \\
    \end{tabular}
    \caption{Differences across configurations among the training setups. In the WD grid for ConvNet, $10^{-3 + n/4}$ and $10^{-2 + n/4}$ correspond to CIFAR-10 and CIFAR-100, respectively. We consider higher LR values compared to ELR values to maintain similar temperature ranges. For the same reason we adopt higher WD values for ConvNet CIFAR-100 experiment.}
    \label{tab:exp_setup}
\end{table}

\paragraph{Detailed experimental setup.}
We train two architectures, ResNet-18~\cite{deep_resnet} and a ConvNet with four convolutional layers (adapted from~\cite{kodryan2022training}), on the CIFAR-10~\cite{cifar10} and CIFAR-100~\cite{cifar100} datasets.
Both models are made fully scale-invariant by inserting a BatchNorm layer without affine parameters after each convolutional layer.
The final linear layer is kept fixed with its weight norm set to $10$.
In both the fixed ELR and fixed LR training protocols, the initial norm of trainable parameters is set to $\|\bm{w}_0\| = 1$.
We use a batch size of $B = 128$ across all experiments, sampling batches independently at each iteration, thus, there is no notion of epochs.

We apply no data augmentations other than channel-wise normalization.
For CIFAR-10, we use mean $(0.4914, 0.4822, 0.4465)$ and standard deviation $(0.2023, 0.1994, 0.2010)$; for CIFAR-100, we use mean $(0.5071, 0.4867, 0.4408)$ and standard deviation $(0.2675, 0.2565, 0.2761)$.
New weights are added to the entropy estimation queue every $25$ iterations.
All other hyperparameters that differ across training setups, including grids for $\eta_{\mathrm{eff}}$, $\eta$, $\lambda$, and $r$, are listed in Table~\ref{tab:exp_setup}.
We record metrics every $200$ iterations during the first $10{,}000$ iterations of training (corresponding to 50 logs).
The remaining $150$ logs are sampled at logarithmically spaced intervals until the end of training.
For stationary metrics, we report averages computed over the last $30$ logs.

\paragraph{Computational resources.}
All experiments are conducted on NVIDIA A100 and H100 GPUs.
Depending on the architecture and GPU type, individual training runs range from approximately 2–3 hours (for the ConvNet on CIFAR-10) to 6–9 hours (for the overparameterized ResNet-18, described in Appendix~\ref{app:op_models}).
The total computational cost amounts to roughly $6{,}000$ GPU hours.

\begin{figure*}[h]
    \centering
    \begin{tabular}{c}
        \textbf{ResNet-18 CIFAR-10} \\ \includegraphics[width=0.98\textwidth]{figures/resnet-cifar10-LR.png} \\
        \textbf{ResNet-18 CIFAR-100} \\ \includegraphics[width=0.98\textwidth]{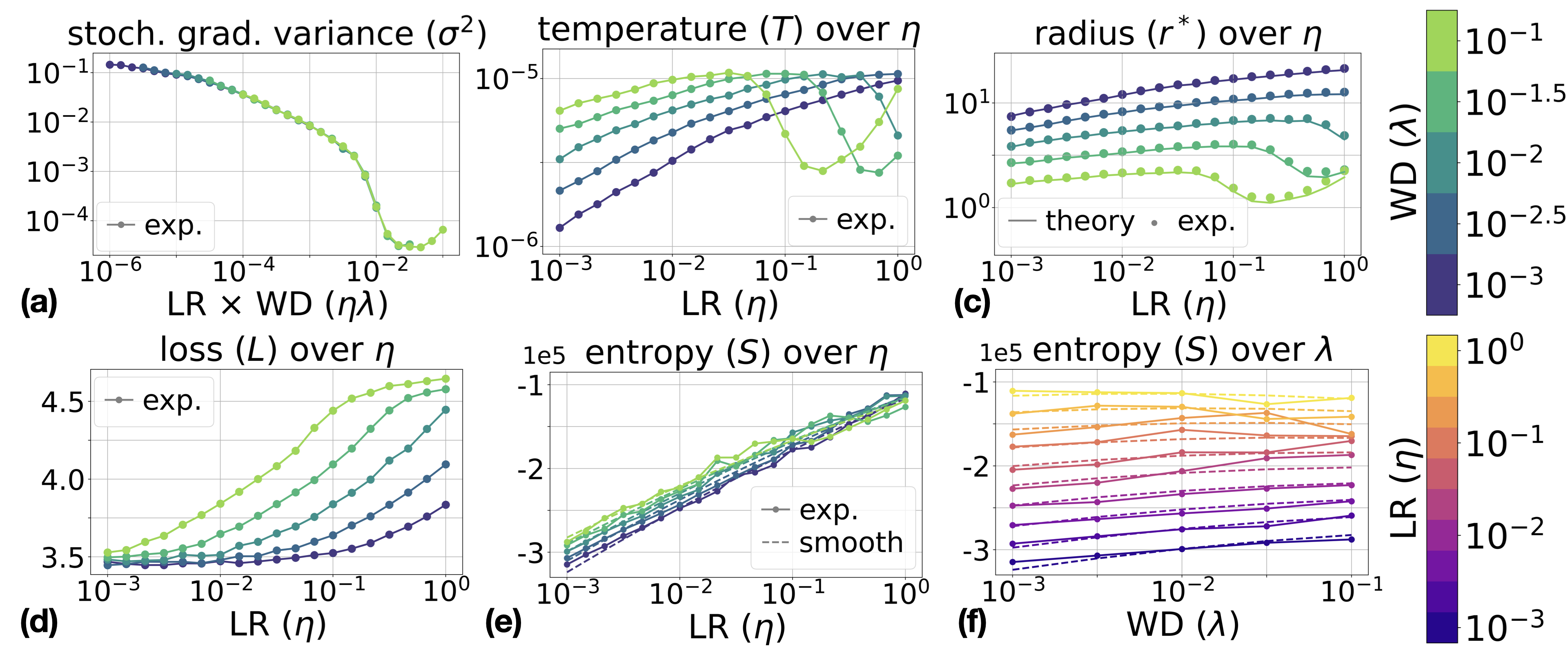} 
    \end{tabular}
    \caption{
    Results for ResNet-18 on CIFAR-10 and CIFAR-100 with fixed LR~$\eta$ and WD~$\lambda$. Subfigures a, b, d: empirically measured $\sigma^2$, mean loss $L$, and temperature $T$ given by $T=\sqrt{\frac{\eta\lambda\sigma^2}{2(d-1)}}$, respectively. Subfigure c: stationary radius $r^*=\sqrt{\frac{T(d-1)}{p}}$ (solid lines, theory) vs. experimental values (points). Subfigures e and f: entropy $S$ as a function of $\eta$ and $\lambda$; solid lines with markers show experimental estimates, dashed lines their smoothed versions.}
    \label{fig:app_resnet_lr}
\end{figure*}

\begin{figure*}
    \centering
    \begin{tabular}{c}
        \textbf{ConvNet CIFAR-10} \\ \includegraphics[width=0.98\textwidth]{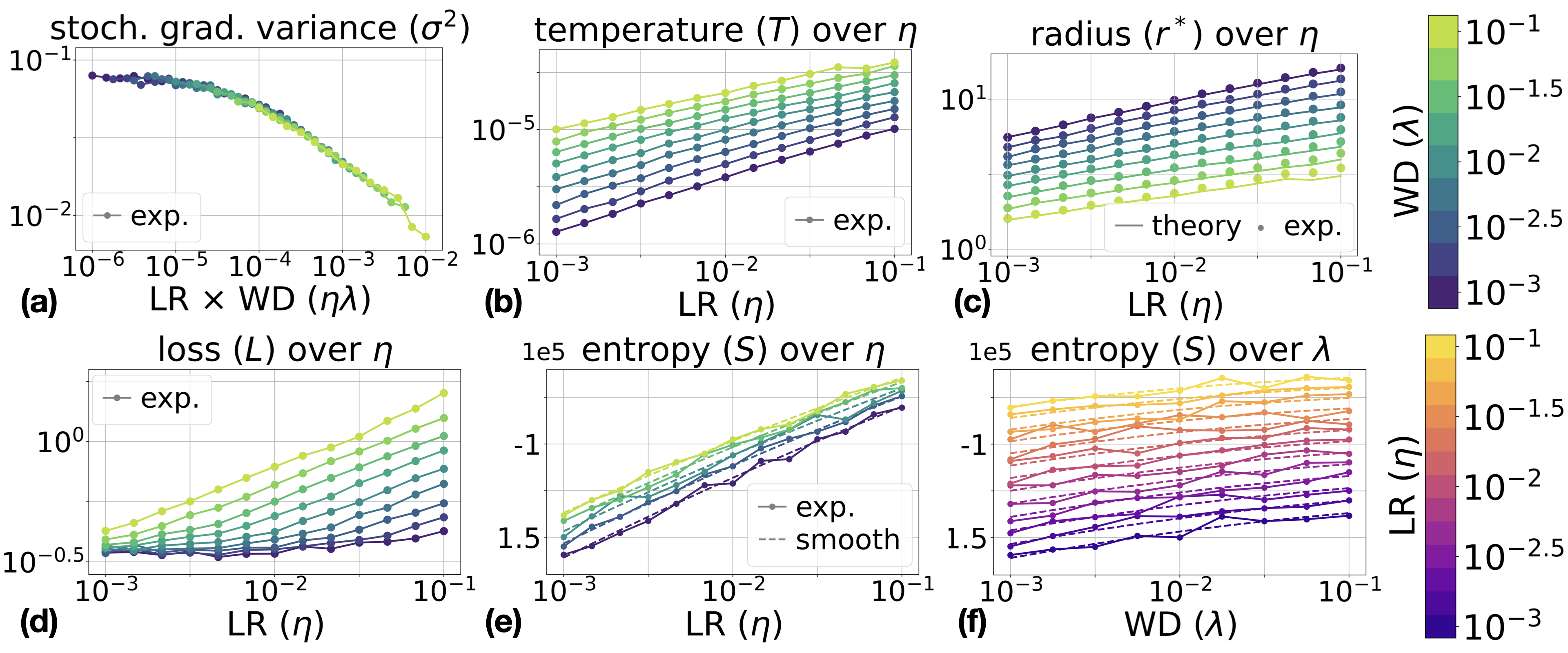} \\
        \textbf{ConvNet CIFAR-100} \\ \includegraphics[width=0.98\textwidth]{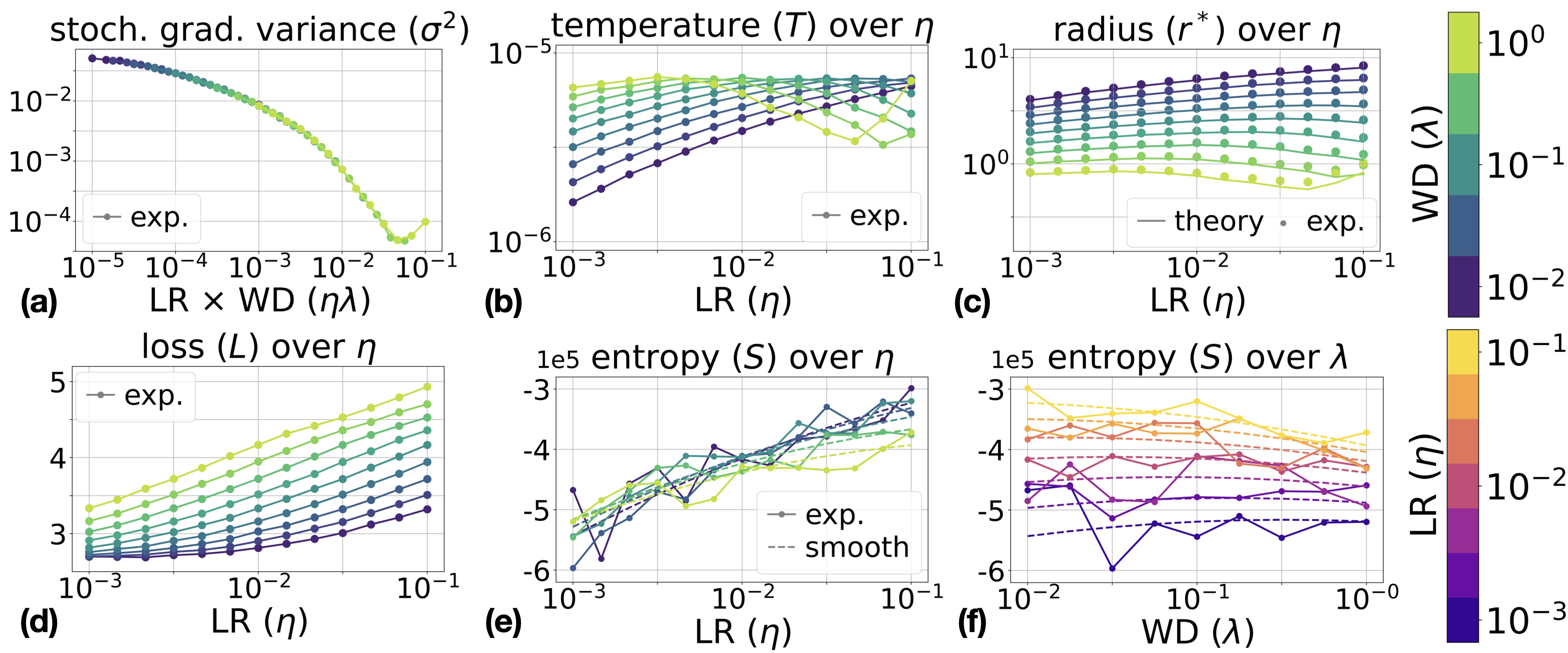}
    \end{tabular}
    \caption{
    Results for ConvNet on CIFAR-10 and CIFAR-100 with fixed LR~$\eta$ and WD~$\lambda$. Subfigures a, b, d: empirically measured $\sigma^2$, mean loss $L$, and temperature $T$ given by $T=\sqrt{\frac{\eta\lambda\sigma^2}{2(d-1)}}$, respectively. Subfigure c: stationary radius $r^*=\sqrt{\frac{T(d-1)}{p}}$ (solid lines, theory) vs. experimental values (points). Subfigures e and f: entropy $S$ as a function of $\eta$ and $\lambda$; solid lines with markers show experimental estimates, dashed lines their smoothed versions.}
    \label{fig:app_convnet_lr}
\end{figure*}

\begin{figure*}
    \centering
    \begin{tabular}{c}
        \textbf{ResNet-18 CIFAR-10} \\ \includegraphics[width=0.98\textwidth]{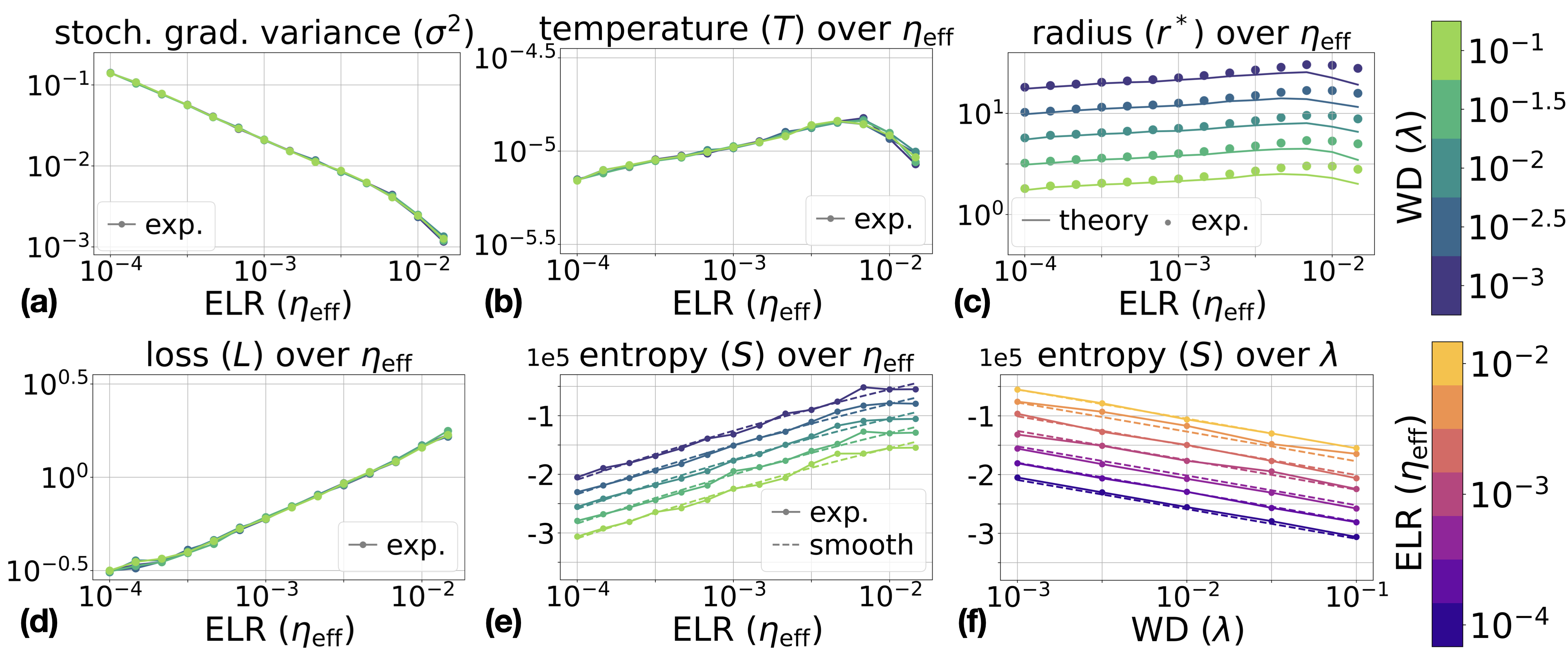} \\
        \textbf{ResNet-18 CIFAR-100} \\ \includegraphics[width=0.98\textwidth]{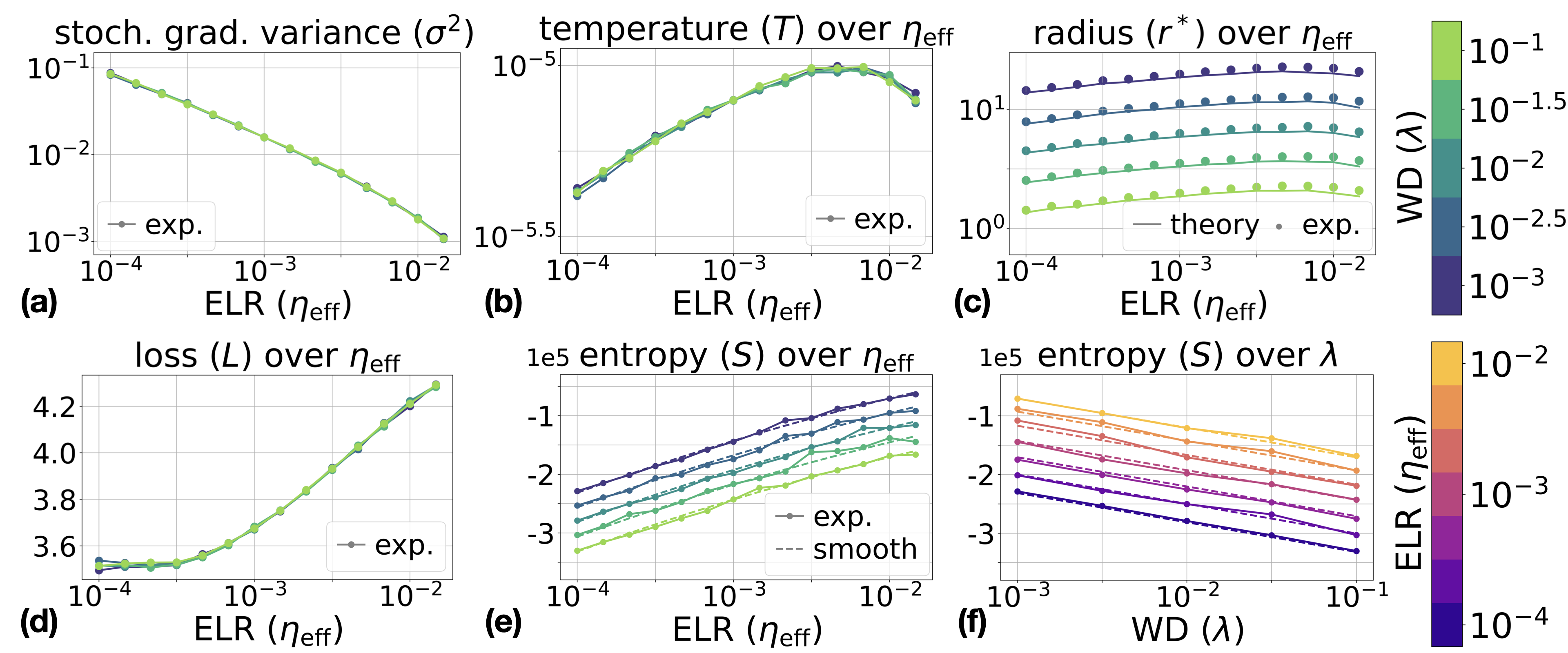} 
    \end{tabular}
    \caption{
    Results for ResNet-18 on CIFAR-10 and CIFAR-100 with fixed ELR~$\eta_{\eff}$ and WD~$\lambda$. Subfigures a, b, d: empirically measured $\sigma^2$, mean loss $L$, and temperature $T$ given by $T=\frac{\eta_{\eff}\sigma^2}{2}$, respectively. Subfigure c: stationary radius $r^*=\sqrt{\frac{T(d-1)}{p}}$ (solid lines, theory) vs. experimental values (points). Subfigures d and e: entropy $S$ as a function of $\eta_{\eff}$ and $\lambda$; solid lines with markers show experimental estimates, dashed lines their smoothed versions.}
    \label{fig:app_resnet_elr}
\end{figure*}

\begin{figure*}
    \centering
    \begin{tabular}{c}
        \textbf{ConvNet CIFAR-10} \\ \includegraphics[width=0.98\textwidth]{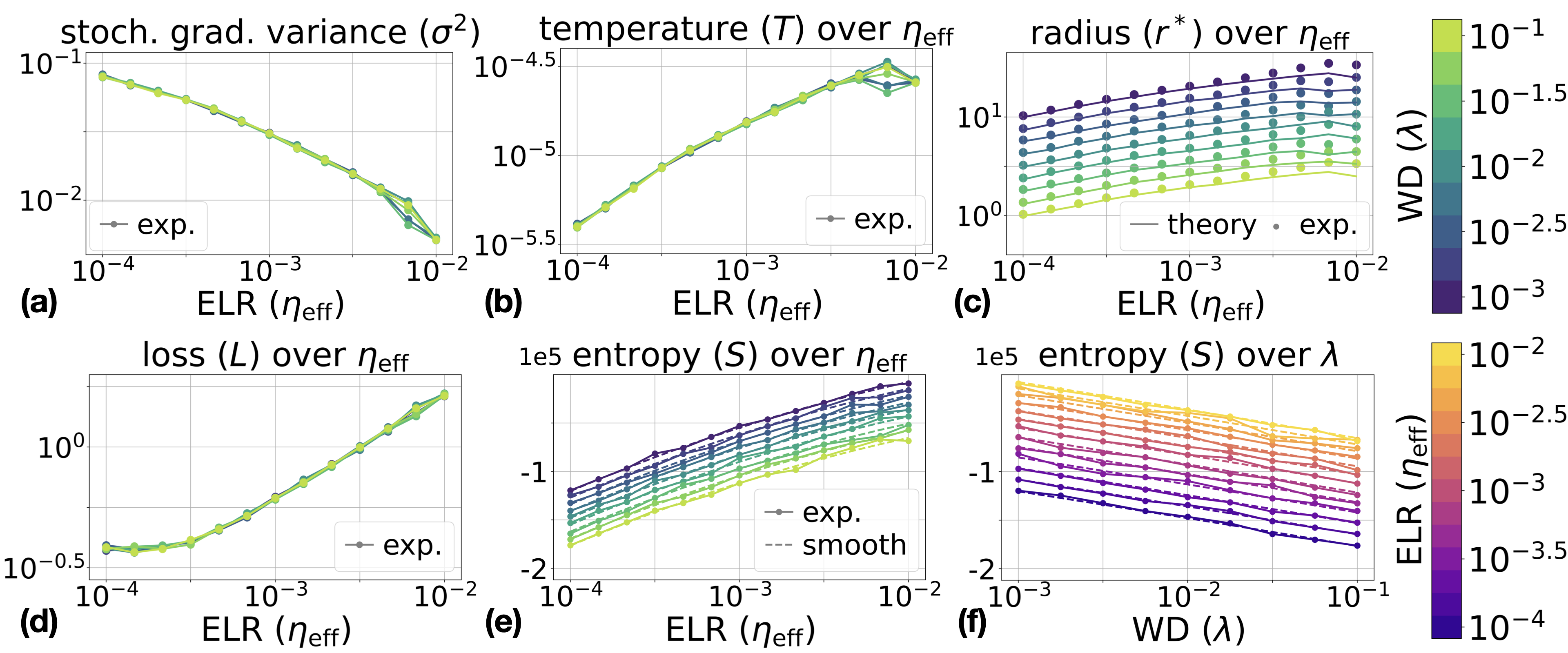} \\
        \textbf{ConvNet CIFAR-100} \\ \includegraphics[width=0.98\textwidth]{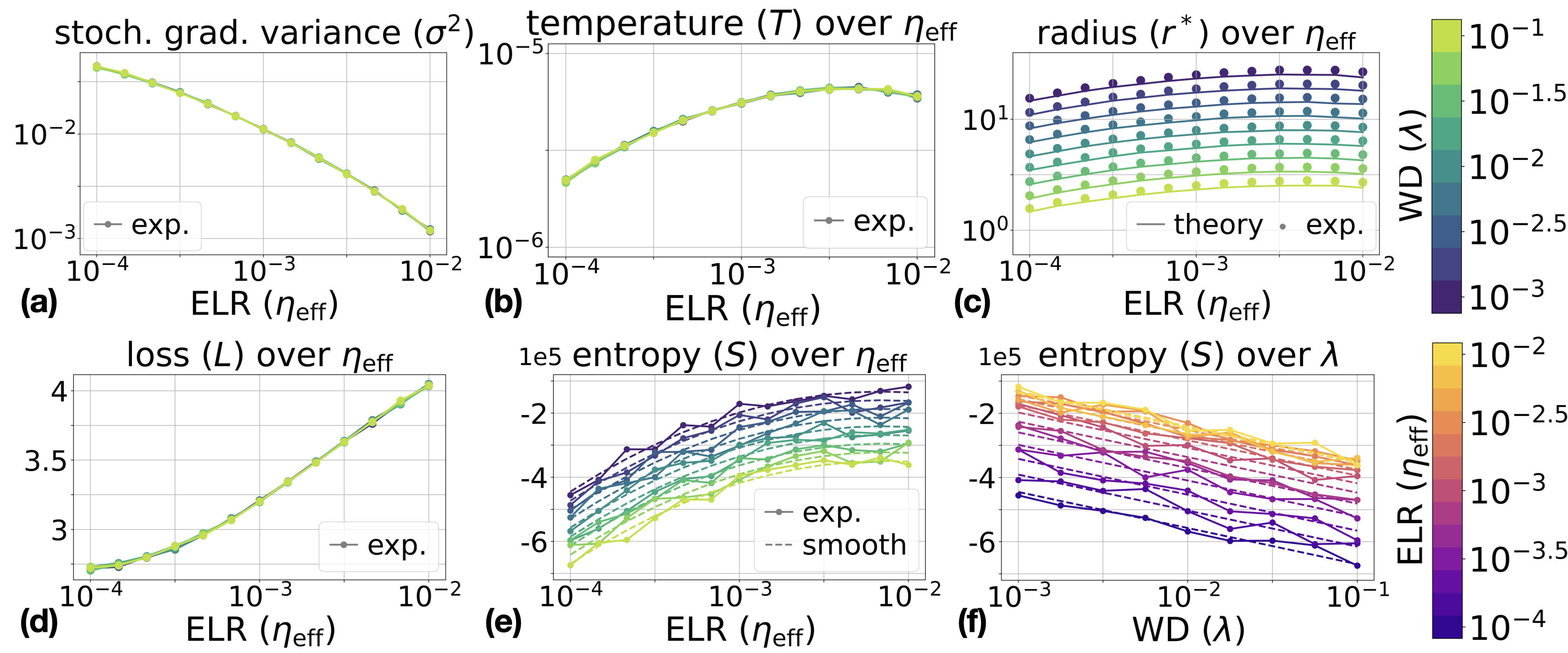}
    \end{tabular}
    \caption{
    Results for ConvNet on CIFAR-10 and CIFAR-100 with fixed ELR~$\eta_{\eff}$ and WD~$\lambda$. Subfigures a, b, d: empirically measured $\sigma^2$, mean loss $L$, and temperature $T$ given by $T=\frac{\eta_{\eff}\sigma^2}{2}$, respectively. Subfigure c: stationary radius $r^*=\sqrt{\frac{T(d-1)}{p}}$ (solid lines, theory) vs. experimental values (points). Subfigures e and f: entropy $S$ as a function of $\eta_{\eff}$ and $\lambda$; solid lines with markers show experimental estimates, dashed lines their smoothed versions.}
    \label{fig:app_convnet_elr}
\end{figure*}

\begin{figure*}
    \centering
    \begin{tabular}{c}
        \textbf{ResNet-18 CIFAR-10} \\ \includegraphics[width=0.98\textwidth]{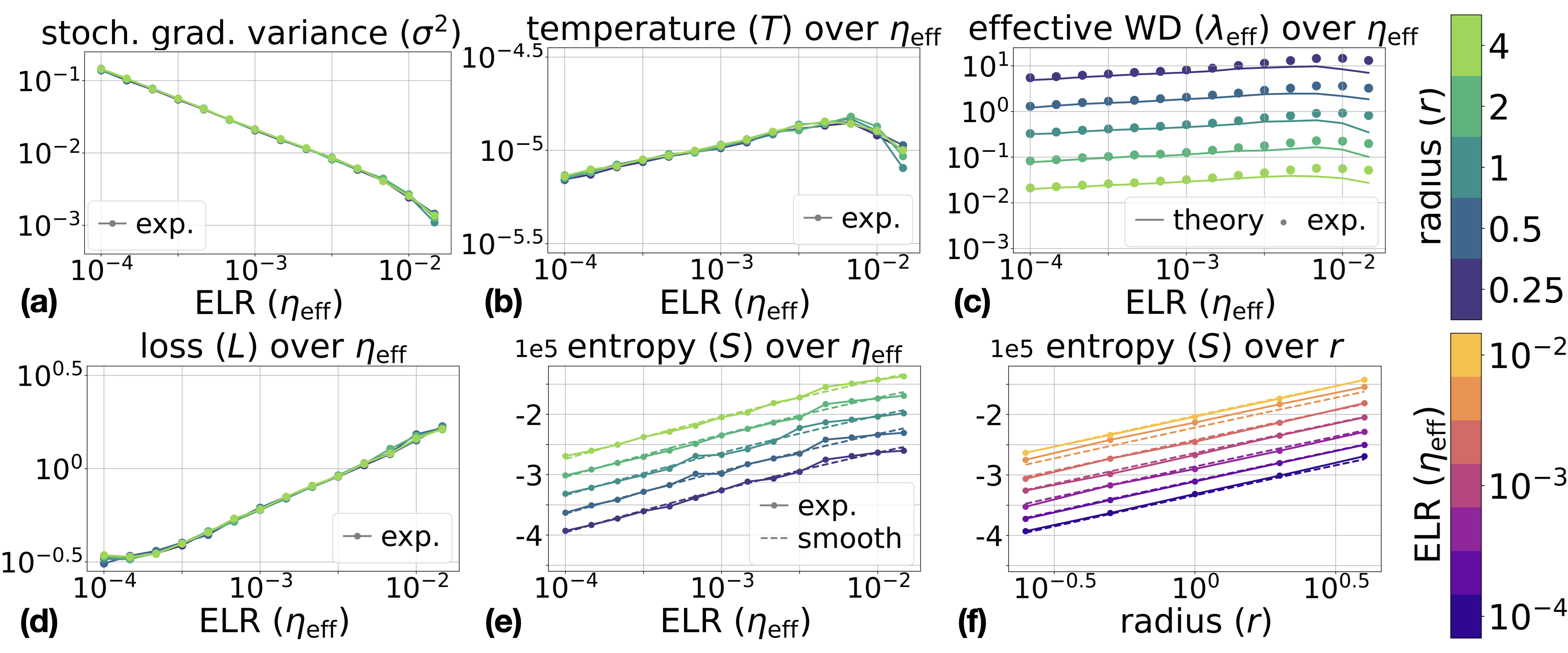} \\
        \textbf{ResNet-18 CIFAR-100} \\ \includegraphics[width=0.98\textwidth]{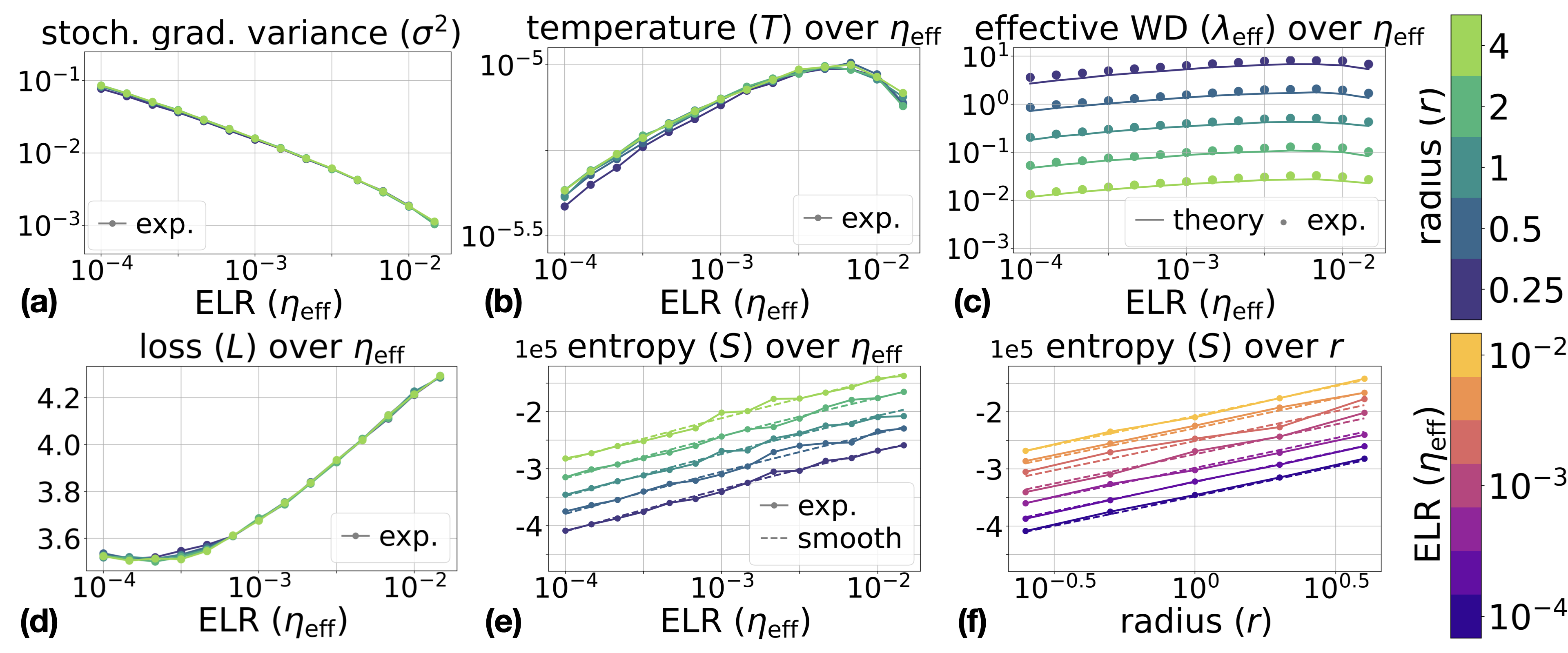} 
    \end{tabular}
    \caption{
    Results for ResNet-18 on CIFAR-10 and CIFAR-100 on a fixed sphere with radius $r$ and fixed ELR~$\eta_{\eff}$. Subfigures a, b, d: empirically measured $\sigma^2$, mean loss $L$, and temperature $T$ given by $T=\frac{\eta_{\eff}\sigma^2}{2}$, respectively. Subfigure c: effective weight decay coefficient $\lambda_{\eff}=\frac{T(d-1)}{2V}$ for $V=\frac{r^2}{2}$ (solid lines, theory) vs. experimental values (points). Subfigures e and f: entropy $S$ as a function of $\eta_{\eff}$ and $r$; solid lines with markers show experimental estimates, dashed lines their smoothed versions.}
    \label{fig:app_resnet_sphere}
\end{figure*}

\begin{figure*}
    \centering
    \begin{tabular}{c}
        \textbf{ConvNet CIFAR-10} \\ \includegraphics[width=0.98\textwidth]{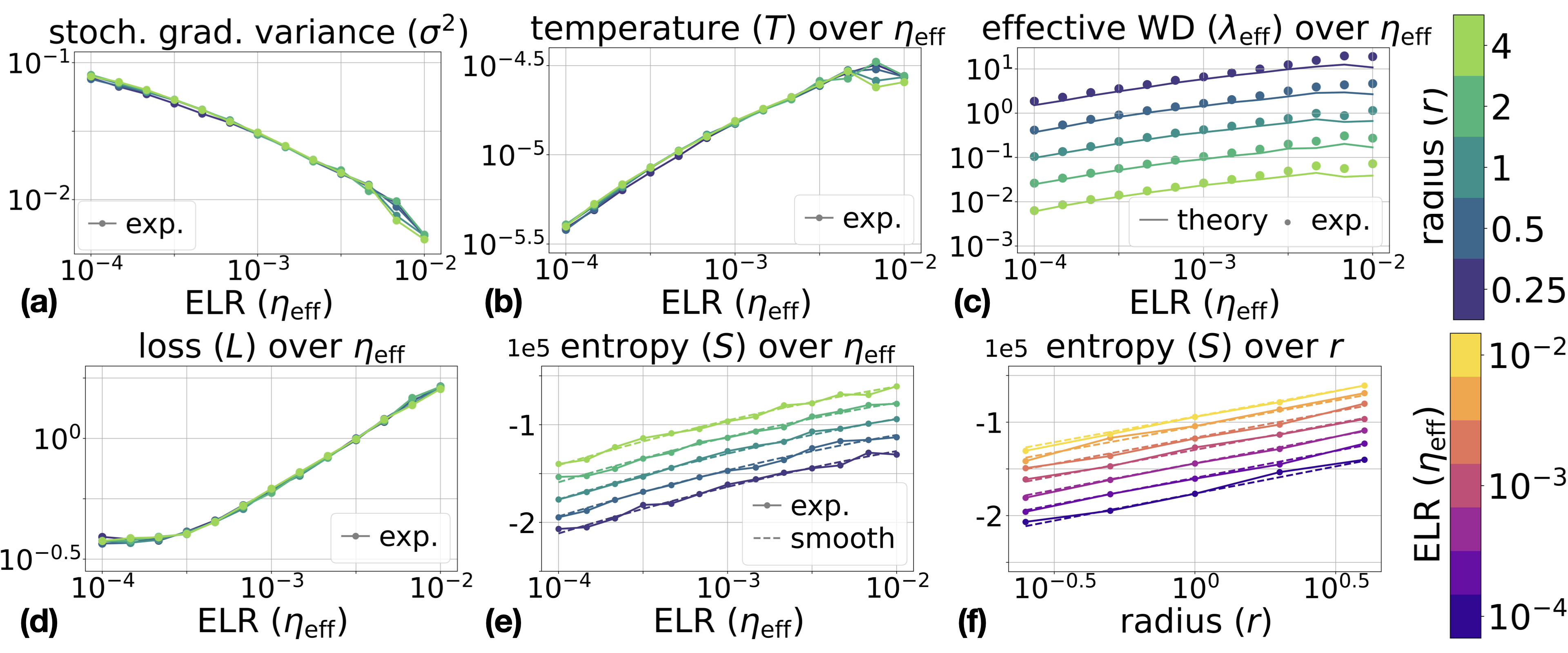} \\
        \textbf{ConvNet CIFAR-100} \\ \includegraphics[width=0.98\textwidth]{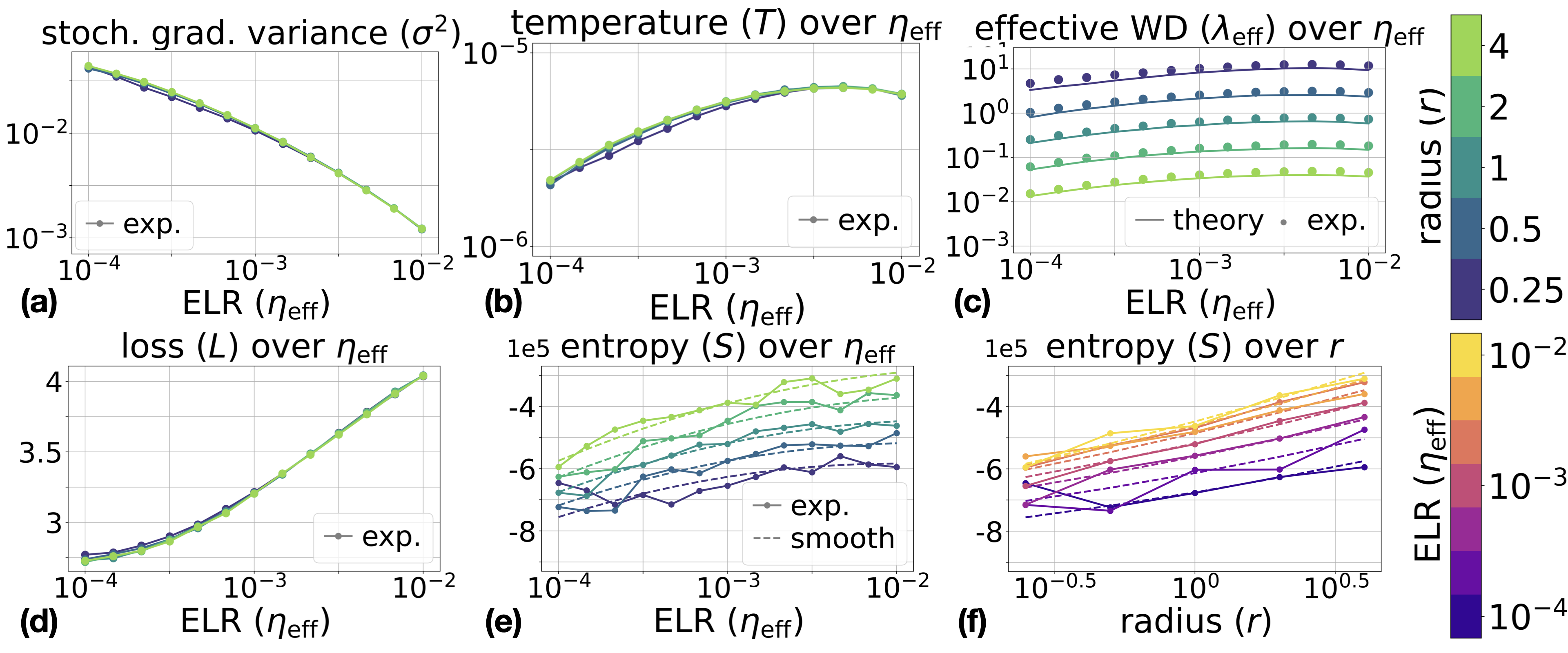}
    \end{tabular}
    \caption{
    Results for ConvNet on CIFAR-10 and CIFAR-100 on a fixed sphere with radius $r$ and fixed ELR~$\eta_{\eff}$. Subfigures a, b, d: empirically measured $\sigma^2$, mean loss $L$, and temperature $T$ given by $T=\frac{\eta_{\eff}\sigma^2}{2}$, respectively. Subfigure c: effective weight decay coefficient $\lambda_{\eff}=\frac{T(d-1)}{2V}$ for $V=\frac{r^2}{2}$ (solid lines, theory) vs. experimental values (points). Subfigures e and f: entropy $S$ as a function of $\eta_{\eff}$ and $r$; solid lines with markers show experimental estimates, dashed lines their smoothed versions.}
    \label{fig:app_convnet_sphere}
\end{figure*}

\paragraph{Results.}
The results, complementing Figure~\ref{fig:resnet_cifar10_lr} from the main text, are presented in Figures~\ref{fig:app_resnet_lr}, \ref{fig:app_convnet_lr}, \ref{fig:app_resnet_elr}, \ref{fig:app_convnet_elr}, \ref{fig:app_resnet_sphere}, \ref{fig:app_convnet_sphere}.
These cover all four architecture-dataset pairs and three training protocols.
Overall, these experiments confirm the results presented in the main text.  
First, the variance of stochastic gradients $\sigma$ depends solely on $\eta_{\text{eff}}$ (in the fixed ELR and fixed sphere settings) and on the product $\eta \lambda$ (in the fixed LR setting).  
Second, the temperature $T$ generally increases with $\eta$ or $\eta_{\text{eff}}$, except at the largest values, where non-monotonic behavior is observed.  
Specifically, $T$ first decreases, due to deviations of $\sigma^2$ from its general trend, and then increases again (Figures~\ref{fig:app_resnet_lr}b, \ref{fig:app_convnet_lr}b) in setups that explore higher ranges of hyperparameters.
\cite{sadrtdinov2024where} demonstrate that for relatively large LRs (so-called regime 2B), the properties of the loss landscape in the region where the network settles change substantially, affecting both generalization and sharpness metrics.
This phenomenon may explain the non-trivial dependencies observed in $\sigma^2$ and $T$.  
Finally, the experimental values of the stationary radius $ r^* $ and effective weight decay $ \lambda_{\mathrm{eff}} $ closely match their theoretical predictions.  

To verify the Maxwell relations~\hyperref[V3]{\bf V3}, we smooth the entropy values using polynomial regression up to quadratic terms\footnote{In the fixed ELR case, we similarly approximate the entropy as a function of $\log \eta_{\eff}$ and $\log \lambda$.}.
\begin{equation}
    S(\log \eta, \log \lambda) = a_0 + a_1\log\eta + a_2\log\lambda + a_3 \log^2\eta + a_4\log^2\lambda + a_5 \log\eta \log\lambda,
\end{equation}
We adopt a quadratic model since the lines of $S$ vary in slope for different values of $\lambda$ (see Subfigure~\ref{fig:resnet_cifar10_lr}e), making a simpler linear regression insufficient, as it enforces identical slopes across $\lambda$.
The resulting partial derivatives are
\begin{equation}
\Big(\frac{\partial S}{\partial \log \eta}\Big)_{\lambda} = a_1 + 2 a_3 \log \eta + a_5 \log \lambda \qquad
\Big(\frac{\partial S}{\partial \log \lambda}\Big)_{\eta} = a_2 + 2 a_4 \log \lambda + a_5 \log \eta
\end{equation}
In the fixed ELR case, we need to check that $\big(\frac{\partial S}{\partial \log \lambda}\big)_{\eta_{\eff}} = -\frac{d-1}{2}$, which is equivalent to $a_2 = -\frac{d - 1}{2}$, $2a_4 = 0$, and $a_5 = 0$.
We report the coefficients divided by $\frac{d-1}{2}$ to evaluate their relative contribution to the derivative, i.e., we should get $\frac{a_2}{(d-1)/2} \approx -1$, $\frac{2a_4}{(d-1)/2} \approx 0$, and $\frac{a_5}{(d-1)/2} \approx 0$.

\begin{table}[t]
    \renewcommand{\arraystretch}{1.3}
    \centering
    \begin{tabular}{cccccc}
    \toprule
    \multicolumn{2}{c}{\textbf{Training setup}} & \makecell{\bf ResNet-18 \\ \bf CIFAR-10} & \makecell{\bf ResNet-18 \\ \bf CIFAR-100} & \makecell{\bf ConvNet \\ \bf CIFAR-10} & \makecell{\bf ConvNet \\ \bf CIFAR-100} \\
    \toprule
    \multirow{5}{*}{Fixed ELR} & approximation $R^2$ & $0.9926$ & $0.9949$ & $0.9975$ & $0.9791$ \\
    & $a_2/(\frac{d - 1}{2})$ & $-1.029$ & $-1.022$ & $-1.128$ & $-0.930$ \\
    & $2a_4/(\frac{d - 1}{2})$ & $-0.008$ & $-0.006$ & $-0.016$ & $-0.0015$ \\
    & $a_5/(\frac{d - 1}{2})$ & $0.0002$ & $0.0005$ & $-0.006$ & $0.009$ \\
    & max. relative error & $2.5\%$ & $1.9\%$ & $6.5\%$ & $4.0\%$ \\
    \midrule 
    \multirow{5}{*}{Fixed LR} & approximation $R^2$ & $0.9939$ & $0.9894$ & $0.9926$ & $0.8412$ \\
    & $(a_1-a_2)/(\frac{d - 1}{2})$ & $0.993$ & $1.075$ & $0.989$ & $0.659$ \\
    & $(2a_3-a_5)/(\frac{d - 1}{2})$ & $0.005$ & $0.00008$ & $-0.038$ & $-0.047$ \\
    & $(2a_4-a_5)/(\frac{d - 1}{2})$ & $0.005$ & $-0.008$ & $-0.033$ & $0.0007$ \\
    & max. relative error & $3.0\%$ & $5.6\%$ & $17.6\%$ & $23.3\%$
    \end{tabular}
    \caption{Verification of Maxwell relation for quadratic approximation of entropy.}
    \label{tab:maxwell}
\end{table}

In the fixed LR case, the Maxwell relation is
\begin{equation}
    \Big(\frac{\partial S}{\partial \log \eta}\Big)_{\lambda} - \Big(\frac{\partial S}{\partial \log \lambda}\Big)_{\eta} = \frac{d-1}{2},
\end{equation}
which is equivalent to $a_1 - a_2 = \frac{d-1}{2}$, $2a_3 - a_5 = 0$, and $2a_4 - a_5 = 0$ (we check that $\frac{a_1-a_2}{(d-1)/2} \approx 1$, $\frac{2a_3 - a_5}{(d-1)/2} \approx 0$, and $\frac{2a_4 - a_5}{(d-1)/2} \approx 0$).
The results of the numerical verification are summarized in Table~\ref{tab:maxwell}.
For most training setups, the maximum relative error remains low (below $10\%$).  
Two notable exceptions occur in the ConvNet experiments on CIFAR-10 and CIFAR-100 with a fixed LR, where the discrepancies increase to $17.6\%$ and $23.3\%$, respectively. 
These higher errors appear near the boundaries of the hyperparameter ranges; when considering the average rather than the maximum error, the values drop to $6.1\%$ and $12.3\%$, indicating that the approximation is substantially more accurate in the interior of the ranges, where entropy is estimated more reliably. 
The larger error in the CIFAR-100 setting is also explained by the highly noisy entropy values observed in Figure~\ref{fig:app_convnet_lr}, which result in a lower coefficient of determination ($R^2 \approx 0.84$) and a less reliable entropy approximation.
Another factor contributing to the error is the non-stationarity of the stochastic gradient noise.
Since Assumption~\ref{asm:5} does not hold in practice, the Maxwell relations are only approximately satisfied.

\clearpage
\section{Results for practical ResNet-18}
\label{app:prac_resnet}

\begin{figure*}[t]
    \centering
    \begin{tabular}{c}
        \textbf{Practical ResNet CIFAR-10} \\ \includegraphics[width=0.98\textwidth]{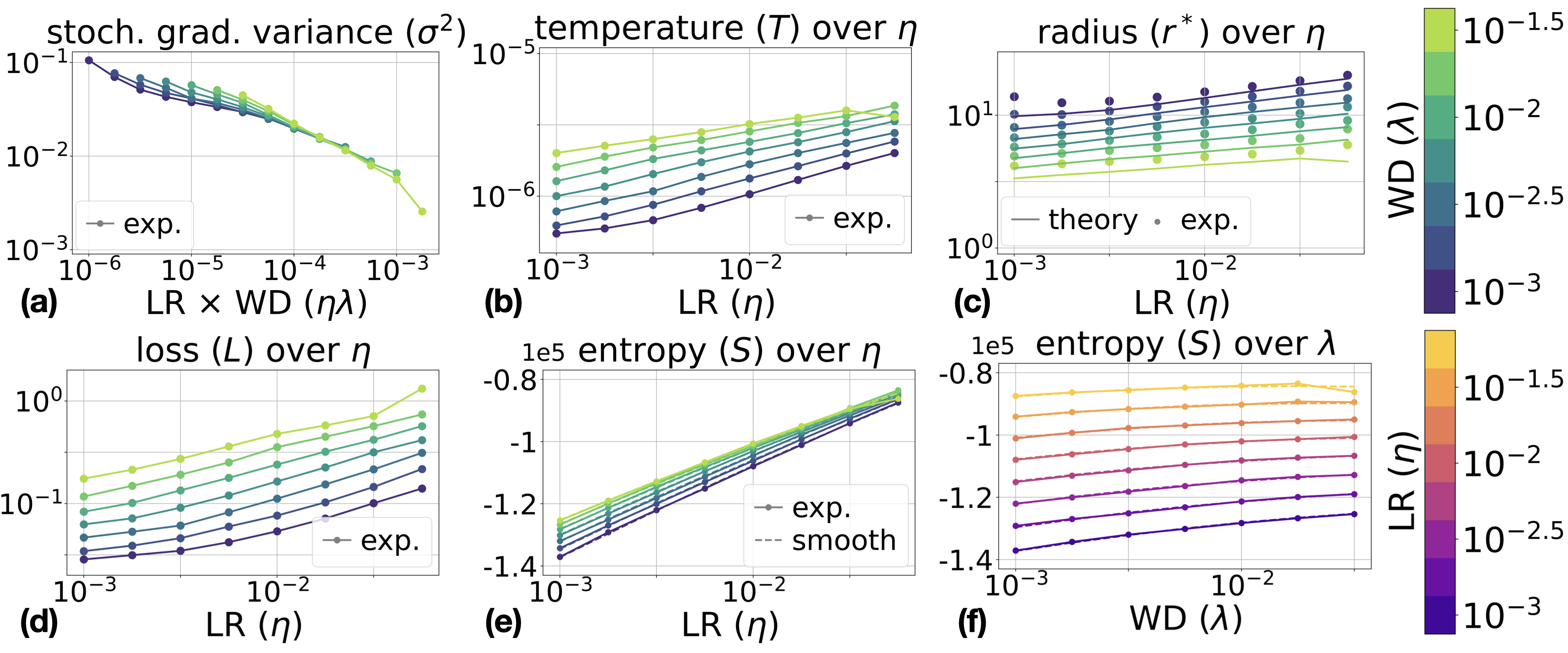}
    \end{tabular}
    \caption{
    Results for practical ResNet on CIFAR-10 with fixed LR~$\eta$ and WD~$\lambda$. Subfigures a, b, d: empirically measured $\sigma^2$, mean loss $L$, and temperature $T$ given by $T=\sqrt{\frac{\eta\lambda\sigma^2}{2(d-1)}}$, respectively. Subfigure c: stationary radius $r^*=\sqrt{\frac{T(d-1)}{p}}$ (solid lines, theory) vs. experimental values (points). Subfigures e and f: entropy $S$ as a function of $\eta$ and $\lambda$; solid lines with markers show experimental estimates, dashed lines their smoothed versions.}
    \label{fig:app_prac_resnet_lr}
\end{figure*}

\paragraph{Experimental setup.}
To demonstrate that our findings hold, at least to some extent, in more general settings, we conducted an experiment using a practical configuration in which some weights are not scale-invariant and data augmentations are applied.
Specifically, we trained a pre-activation ResNet-18 with width factor $k=8$, containing $176152$ parameters, $1600$ of which are non-scale-invariant (including affine parameters in BatchNorm layers and the linear classification head).
The share of scale-invariant parameters is dominant, so we expect the overall dynamics to be similar to fully scale-invariant networks. 
The network was trained on CIFAR-10 with fixed LR and WD in the ranges $\eta \in \{10^{-3 + n/4}|n=0,\dots, 7\}$ and $\lambda \in \{10^{-3 + n/4}|n=0,\dots, 6\}$, respectively.
Smaller hyperparameter values did not lead to stabilization of the relevant metrics even after $t=2\cdot 10^6$ iterations, whereas larger values caused training instabilities.
This mirrors the logic used when selecting hyperparameters in our scale-invariant experiments.
Standard data augmentations (random crops) were applied.
We measured both the radius and the entropy of the full parameter set, combining scale-variant and scale-invariant parameters.
The rest of the training configuration is consistent with experimental setup described in Appendix~\ref{app:networks}. 

\paragraph{Results.}
The results are shown in Figure~\ref{fig:app_prac_resnet_lr}.
Subfigure~\ref{fig:app_prac_resnet_lr}a demonstrates that the gradient variance $\sigma^2$ is no longer a simple function of $\eta\lambda$; instead, it depends on both hyperparameters in a more intricate manner, reflecting the effects of scale-variant parameters.
We also find that the theoretical prediction of the stationary radius $r^*$ systematically underestimates its empirical value (Subfigure~\ref{fig:app_prac_resnet_lr}c), consistent with Appendix~G of~\cite{kosson2024rotational}. This discrepancy arises because stochastic gradients are not perfectly orthogonal to the radial direction, requiring a stronger centripetal force (i.e., larger weight decay) to stabilize the dynamics and resulting in a larger radius.
In contrast, the stationary entropy behaves similarly to the fully scale-invariant case: it is well approximated by the quadratic model (Subfigures~\ref{fig:app_prac_resnet_lr}e,f), with $R^2 \approx 0.9996$. Moreover, the Maxwell relations hold approximately, with a maximum absolute error of $8.6\%$, as indicated by $\frac{a_1-a_2}{(d-1)/2} \approx 1.195$, $\frac{2a_3-a_5}{(d-1)/2} \approx 0.011$, and $\frac{2a_4-a_5}{(d-1)/2} \approx -0.023$.

\section{Discretization error of SGD}
\label{app:discr_error}

In this section, we discuss the discrepancy between the theoretical and experimental values of the stationary radius $r^*$, illustrated in Subfigure~\ref{fig:resnet_cifar10_lr}c.

\subsubsection{Correcting SDE predictions}
Since the SDE framework models continuous-time dynamics, the full-batch gradient $\nabla L(\bm{w})$, which is orthogonal to $\bm{w}$, does not contribute to the centrifugal force and therefore does not influence the radius dynamics.
Although stochastic gradients $\nabla L_{\mathcal{B}_k}(\bm{w})$ are also orthogonal to $\bm{w}$, the Ito's correction term, arising from the quadratic variation of Brownian motion ($(\dd \bm{B}_t)_i^2 = \dd t$), introduces an additional deterministic centrifugal force.
Because the SDE formulation only approximates discrete-time training dynamics, discrepancies between the continuous and discrete descriptions grow with larger $\eta_{\eff}$ (fixed sphere/fixed ELR cases) or larger $\eta \lambda$ (fixed LR case).
\begin{wrapfigure}{r}{0.27\textwidth}
  \begin{center}
    \includegraphics[width=0.27\textwidth]{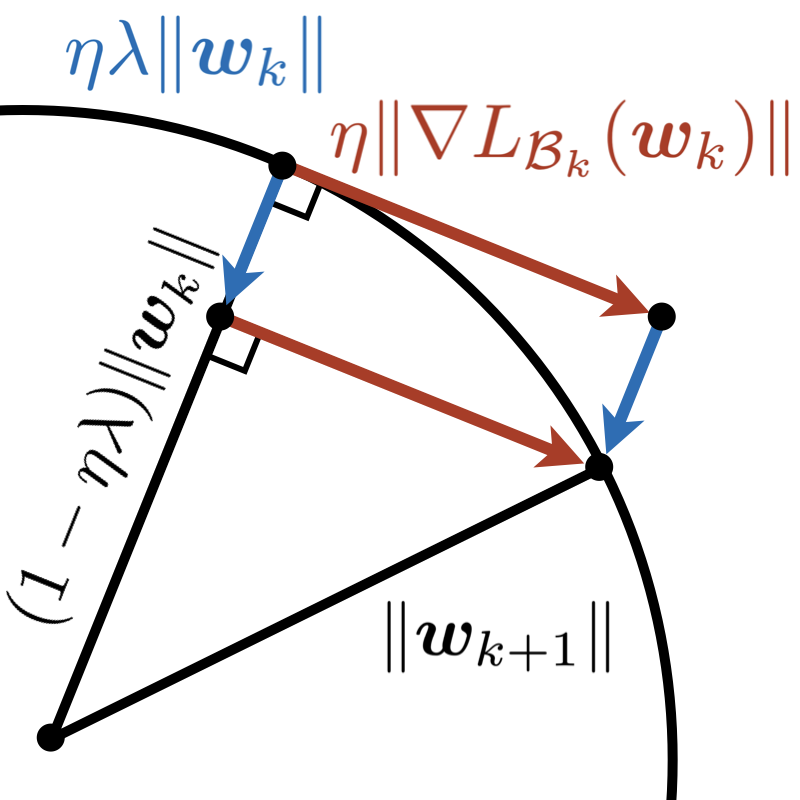}
  \end{center}
  \caption{The balance between {\color{BrickRed}{centrifugal}} and {\color{NavyBlue}{centripetal}} forces, which preserves the weight norm $\|\bm{w}_k\|$ after the SGD step. Stochastic gradient $\nabla L_{\mathcal{B}_k}(\bm{w}_k)$ is orthogonal to the weight vector $\bm{w}_k$.}
  \label{fig:pythagoras}
\end{wrapfigure}

To account for this mismatch, we introduce a geometric correction that explicitly considers both the deterministic and stochastic components of the gradient.
A similar geometric reasoning has been discussed in~\cite{kosson2024rotational}.
Specifically, we focus on the stationary regime, where the centrifugal force (due to stochastic gradients) and the centripetal force (induced by weight decay) are in equilibrium, as illustrated in Figure~\ref{fig:pythagoras}.
Since the stochastic gradient $\nabla L_{\mathcal{B}_k}(\bm{w}_k)$ is orthogonal to the weight vector $\bm{w}_k$ (Eq.~\ref{eqt:2}, \ref{eqt:3}), we can apply the Pythagorean theorem to describe their relationship
\begin{equation}
    \|w_{k+1}\|^2 = (1 - \eta\lambda)^2 \|w_{k}\|^2 + \eta^2 \|\nabla L_{\mathcal{B}_k} (\bm{w}_k)\|^2
\end{equation}
Taking the expectation of both sides over the stationary distribution, we get $\mathbb{E}\|\bm{w}_{k}\|^2 = \mathbb{E}\|\bm{w}_{k + 1}\|^2 = \mathbb{E}\|\bm{w}\|^2$ and $\mathbb{E} \|\nabla L_{\mathcal{B}_k} (\bm{w}_k)\|^2 = \|\nabla L_{\mathcal{B}} (\bm{w})\|^2$.
Hence
\begin{gather}
    \mathbb{E} \|\bm{w}\|^2 = \eta^2 \mathbb{E} \|\nabla L_{\mathcal{B}} (\bm{w})\|^2 + (1 - \eta\lambda)^2 \mathbb{E} \|\bm{w}\|^2 \\
    \mathbb{E} \|\bm{w}\|^2 = \eta^2 \mathbb{E} \|\nabla L_{\mathcal{B}} (\bm{w})\|^2 + (1 - 2\eta\lambda + \eta^2\lambda^2) \mathbb{E} \|\bm{w}\|^2 \\
    \eta \mathbb{E} \|\nabla L_{\mathcal{B}} (\bm{w})\|^2 = (2\lambda - \eta\lambda^2) \mathbb{E} \|\bm{w}\|^2
\end{gather}
We further assume that both $\eta$ and $\lambda$ are small ($\eta, \lambda \ll 1$), allowing us to neglect the term $\eta\lambda^2 \mathbb{E}\|\bm{w}\|^2$ as a higher-order approximation.
Additionally, we assume that the variance of $\|\bm{w}\|^2$ is negligible, which enables us to write (using Eq.~\ref{eqt:2}, \ref{eqt:3})
\begin{equation}
    \mathbb{E} \|\nabla L_{\mathcal{B}} (\bm{w})\|^2 = \mathbb{E} \biggl[\frac{\|\nabla L_{\mathcal{B}} (\overline{\bm{w}})\|^2}{\|\bm{w}\|^2}\biggr] \approx \frac{\mathbb{E}\|\nabla L_{\mathcal{B}} (\overline{\bm{w}})\|^2}{\mathbb{E}\|\bm{w}\|^2}
\end{equation}
Therefore, we get the following equation, which ties $\mathbb{E}\|\bm{w}\|^2$ and $\mathbb{E} \|\nabla L_{\mathcal{B}} (\bm{w})\|^2$
\begin{equation}
    \eta \frac{\mathbb{E} \|\nabla L_{\mathcal{B}} (\overline{\bm{w}})\|^2}{\mathbb{E}\|\bm{w}\|^2} = 2\lambda \, \mathbb{E} \|\bm{w}\|^2
\end{equation}
Finally, we express the expectation of stochastic gradient norm via the covariance matrix trace and the expectation of the full-batch gradient norm (Eq.~\ref{eqt:1}):
\begin{gather}
    \mathbb{E} \|\nabla L_{\mathcal{B}} (\overline{\bm{w}})\|^2 = \mathbb{E}\Big\|\nabla L (\overline{\bm{w}}) + (\bm{\Sigma}_{\overline{\bm{w}}})^{1/2} \bm{\varepsilon}\Big\|^2 = \mathbb{E} \|\nabla L (\overline{\bm{w}})\|^2 + 2 \, \mathbb{E} \Bigl[\nabla L (\overline{\bm{w}})^T (\bm{\Sigma}_{\overline{\bm{w}}})^{1/2} \bm{\varepsilon} \Bigr] \; + \\
    + \mathbb{E} \|(\bm{\Sigma}_{\overline{\bm{w}}})^{1/2} \bm{\varepsilon}\|^2 = \mathbb{E} \|\nabla L (\overline{\bm{w}})\|^2 + \Tr \bm{\Sigma}_{\overline{\bm{w}}} \nonumber
\end{gather}
Thus, the resulting relation is
\begin{equation}
    \eta \frac{\Tr \bm{\Sigma}_{\overline{\bm{w}}} + \mathbb{E} \|\nabla L (\overline{\bm{w}})\|^2}{\mathbb{E}\|\bm{w}\|^2} = 2\lambda \, \mathbb{E} \|\bm{w}\|^2
\end{equation}
By denoting $r^*_{\text{discr}} = \mathbb{E} \|\bm{w}\| \approx \sqrt{\mathbb{E}\|\bm{w}\|^2}$, we can compare this discrete prediction of radius to the continuous-time prediction $r^*_{\text{SDE}}$.
In the fixed LR case, we have
\begin{equation}
    r^*_{\text{discr}} = \sqrt[4]{\frac{\eta}{2\lambda} (\Tr \bm{\Sigma}_{\overline{\bm{w}}} + \mathbb{E} \|\nabla L \bigl(\overline{\bm{w}})\|^2\bigr)}, \qquad r^*_{\text{SDE}} = \sqrt[4]{\frac{\eta\sigma^2(d-1)}{2\lambda}} = \sqrt[4]{\frac{\eta}{2\lambda} \Tr \bm{\Sigma}_{\overline{\bm{w}}}}
\end{equation}
In the fixed ELR case, $\eta = \eta_{\eff} \mathbb{E} \|\bm{w}\|^2$, so the values of $r^*$ and $r^*_{\text{SDE}}$ are
\begin{equation}
    r^*_{\text{discr}} = \sqrt{\frac{\eta_{\eff}}{2\lambda} \bigl(\Tr \bm{\Sigma}_{\overline{\bm{w}}} + \mathbb{E} \|\nabla L \bigl(\overline{\bm{w}})\|^2\bigr)}, \qquad r^*_{\text{SDE}} = \sqrt{\frac{\eta_{\eff}\sigma^2(d-1)}{2\lambda}} = \sqrt{\frac{\eta_{\eff}}{2\lambda} \Tr \bm{\Sigma}_{\overline{\bm{w}}}}
\end{equation}
In the fixed sphere case, the values of effective WD $\lambda_{\eff, \text{discr}}$ and $\lambda_{\eff,  \text{SDE}}$ are given by
\begin{equation}
    \lambda_{\eff, \text{discr}} = \frac{\eta_{\eff}}{2r^2} \bigl(\Tr \bm{\Sigma}_{\overline{\bm{w}}} + \mathbb{E} \|\nabla L \bigl(\overline{\bm{w}})\|^2\bigr), \qquad \lambda_{\eff,  \text{SDE}} = \frac{\eta_{\eff}}{2r^2} \Tr \bm{\Sigma}_{\overline{\bm{w}}}
\end{equation}
We observe that, in all three cases, the expected squared norm of the full-batch gradient is added to the trace of the covariance matrix, thereby correcting the SDE prediction.

\begin{figure*}
    \centering
    \begin{tabular}{c}
        \textsc{Fixed LR \qquad\qquad\qquad\qquad\qquad Fixed ELR \qquad\qquad\qquad\qquad\qquad Fixed Sphere} \\
        \textbf{ResNet-18 CIFAR-10} \\
        \includegraphics[width=0.98\textwidth]{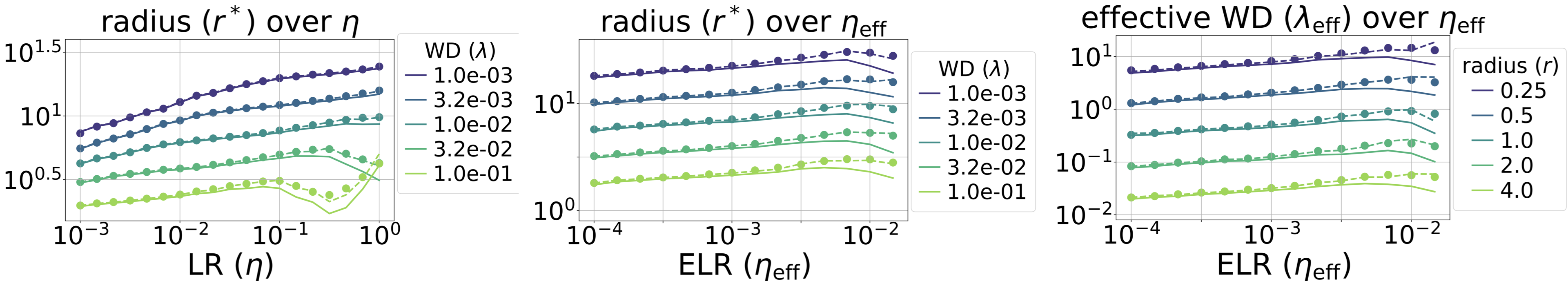} \\
        \textbf{ResNet-18 CIFAR-100} \\
        \includegraphics[width=0.98\textwidth]{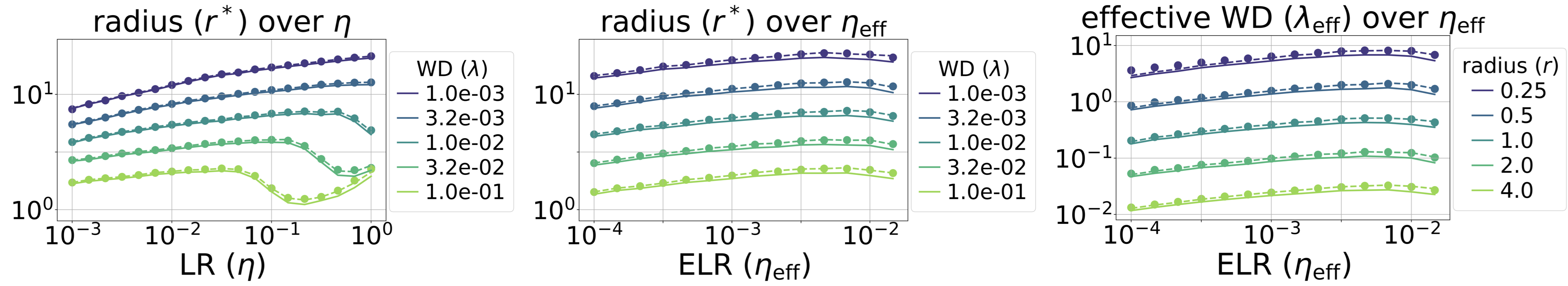} \\
        \textbf{ConvNet CIFAR-10} \\
        \includegraphics[width=0.98\textwidth]{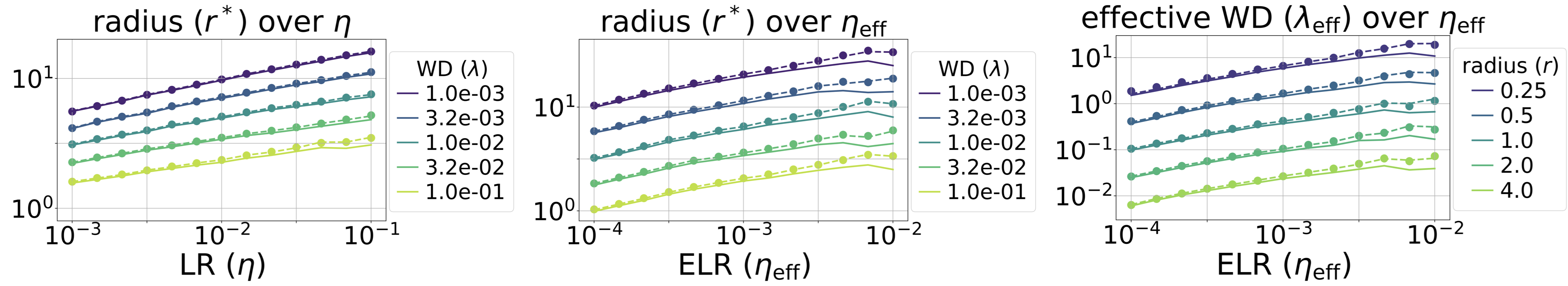} \\
        \textbf{ConvNet CIFAR-100} \\
        \includegraphics[width=0.98\textwidth]{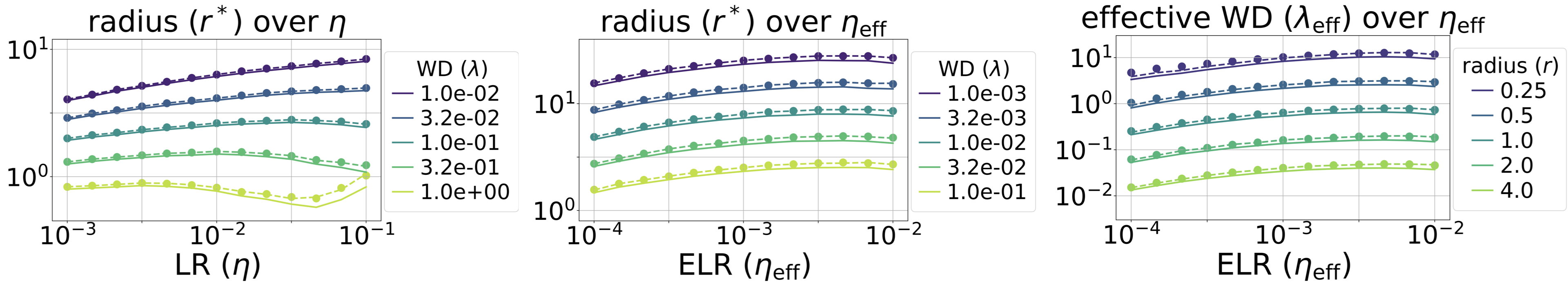} \\
        \includegraphics[width=0.5\textwidth]{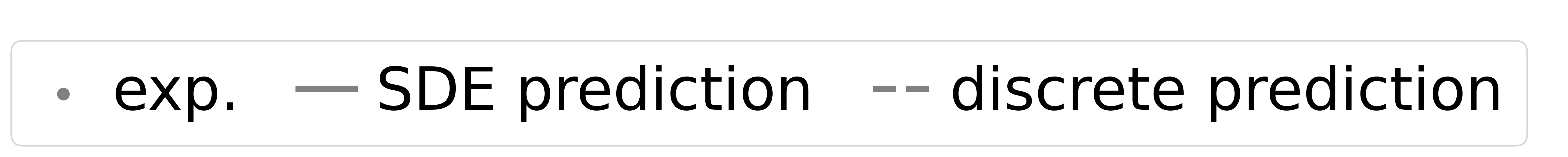}
    \end{tabular}
    \caption{
    Comparison between the discrete-time and SDE predictions of the stationary radius $r^*$ for the fixed LR (left column) and fixed ELR (center column) cases, and of the effective weight decay $\lambda_{\eff}$ for the fixed sphere case (right column).
    }
    \label{fig:discr_error}
\end{figure*}

\paragraph{Results.}
Figure~\ref{fig:discr_error} compares the discrete-time and SDE predictions across all four architecture–dataset pairs.
We observe that the discrete-time predictions more closely match the experimental results, particularly for larger values of $\eta$ and $\eta_{\eff}$.

\section{Overparameterized models}
\label{app:op_models}

\paragraph{Experimental setup.}
We train a scale-invariant ResNet-18  on the CIFAR-10 dataset.
To ensure the model is overparameterized, i.e., capable of fitting the training set with $100\%$ accuracy, we increase the width multiplier to $k=32$.
The entropy queue size is reduced to $500$.
Training is performed using several fixed ELR values and a weight decay coefficient of $\lambda = 10^{-3}$.
All other hyperparameters are identical to those used for the thinner ResNet-18 model trained on CIFAR-10 with $k=4$.

\begin{figure*}[t]
    \centering
    \includegraphics[width=0.85\textwidth]{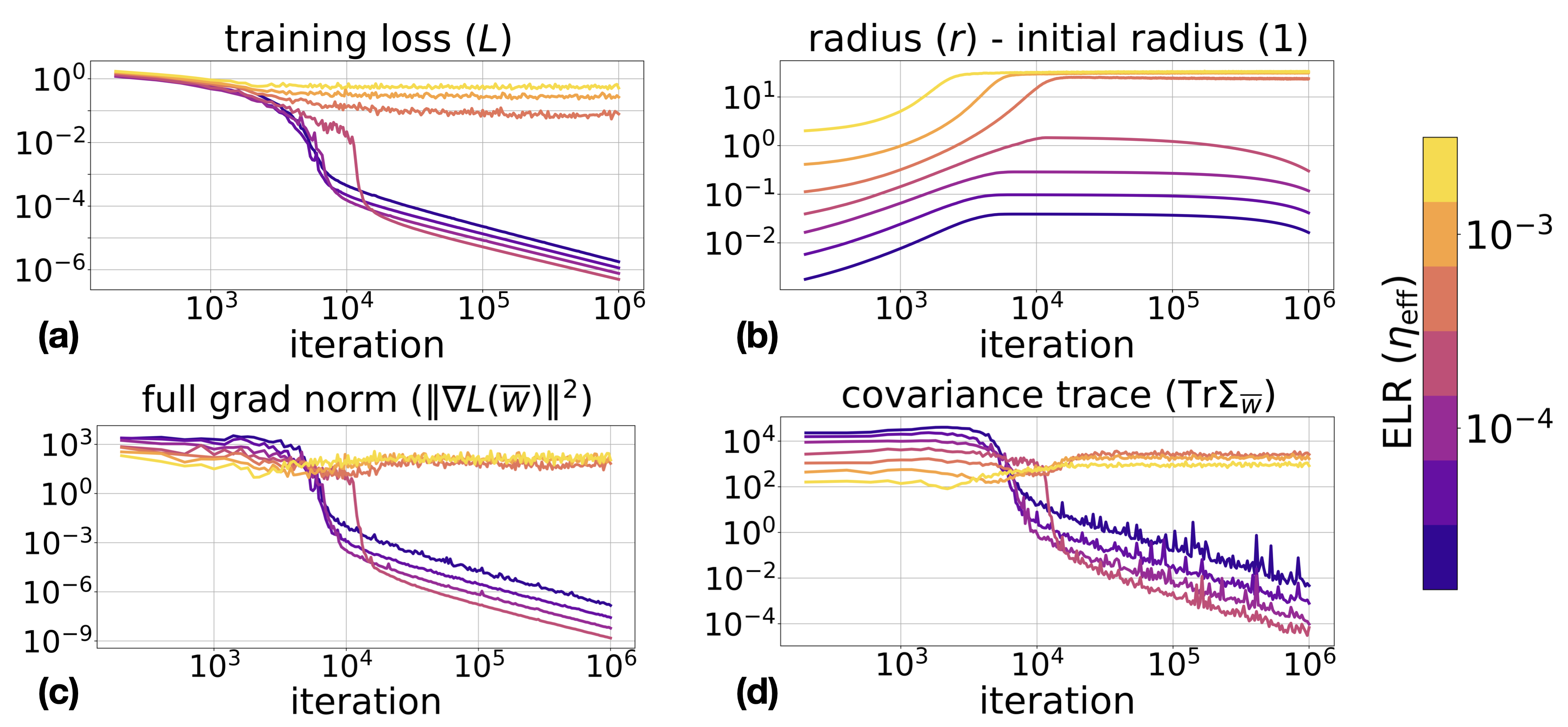}
    \caption{
    Results for overparameterized ResNet-18 on CIFAR-10 with fixed ELR $\eta_{\eff}$ and weight decay $\lambda = 10^{-3}$.
    }
    \label{fig:op_resnet_cifar10}
\end{figure*}

\paragraph{Results.}
Figure~\ref{fig:op_resnet_cifar10} shows the learning curves for training loss, parameter radius, and gradient-related metrics (the squared norm of the full-batch gradient, $\|\nabla L(\overline{\bm{w}})\|^2$, and the trace of the covariance matrix, $\Tr \Sigma_{\overline{\bm{w}}}$).
We observe that all four metrics stabilize for the three largest ELRs, indicating the onset of stationary behavior.
In contrast, for smaller ELRs, the training loss and gradient metrics continue to decrease steadily after approximately $10^4$ iterations.
This behavior closely matches the convergence regime previously reported by~\cite{kodryan2022training} or so-called interpolation mode of~\cite{interp_sgd} and is specific to overparameterized networks.

When overparameterized models enter the interpolation mode~\cite{interp_sgd}, SGD with a fixed LR behaves similarly to full-batch gradient descent: it converges to a minimum rather than stabilizing at a stationary distribution~\cite{du2018gradient,zou2018stochastic,pmlr-v89-nacson19a}.
This occurs because stochastic fluctuations diminish during training, specifically, $\Tr \Sigma_{\overline{\bm{w}}}$ decreases, as shown in Subfigure~\ref{fig:op_resnet_cifar10}d, so the noise in the stochastic gradients no longer prevents convergence.
From a thermodynamic perspective, this corresponds to the limit $T \to 0$, where the Gibbs distribution~\ref{T3} becomes degenerate, concentrating on the microstate $i$ with the lowest energy $E_i$.

As optimization approaches the minimum, the centrifugal force $\eta \nabla L_{\mathcal{B}_k} (\bm{w}_k)$ weakens due to the decreasing norm of the stochastic gradient.
However, the centripetal force, $-\eta\lambda\bm{w}_k$, remains large, gradually driving the radius to shrink.
Eventually (after an infinite amount of training iterations) the centripetal force would vanish entirely, and the radius would collapse, leading to fluctuations around the origin, analogous to full-batch gradient descent in the scale-invariant setting~\cite{li2020reconciling}.
Although the radius for small ELRs remains larger than its initial value in Subfigure~\ref{fig:op_resnet_cifar10}b, it shows a consistent decline after approximately $10^5$ iterations.
In practice, convergence to the loss minimum on the sphere can be slower than convergence to the origin, leading to instabilities and periodic behavior~\cite{lobacheva2021periodic} caused by an increase in the gradient norm when moving towards the origin.  

\end{document}